\documentclass[12pt,draft,letterpaper]{report}

\newcommand{\thesistitle}{Image Segmentation Using Subspace Representation and Sparse Decomposition}
\newcommand{\thesisauthor}{Shervin Minaee}
\newcommand{\thesisadvisor}{Yao Wang}
\newcommand{\graddate}{May 2018} 

\newcommand{\thesisdedication}{\textit{Dedicated to my family.}}

\usepackage{setspace}
\usepackage{xspace}
\usepackage{algorithmic}
\usepackage{algorithm}
\usepackage{microtype}
\usepackage{subfigure}
\usepackage{color}
\usepackage{url}
\usepackage{lipsum}
\usepackage{fancyhdr}

\fancypagestyle{plain}{%
\fancyhf{}
\rhead{\thepage}
}

\usepackage[final]{graphicx}

\usepackage{indentfirst}

\usepackage{amsthm} 
\usepackage[tbtags]{amsmath} 

\usepackage{amsfonts}
\usepackage{amssymb}

\usepackage{xspace}
\usepackage{algorithmic}
\usepackage{algorithm}
\usepackage{microtype}
\usepackage{subfigure}
\usepackage{color}
\usepackage{todonotes}
\usepackage{url}
\usepackage{dcolumn}

\usepackage{epsfig}
\usepackage{blkarray}
\usepackage{array}
\usepackage{epstopdf}
\usepackage{calrsfs}

\usepackage{multirow}

\usepackage{listings}

\usepackage[numbers,sort]{natbib}

\usepackage[all,cmtip]{xy}

\usepackage{hyperref}
\hypersetup{
    colorlinks=true,
    linkcolor=blue,
    filecolor=blue,
    urlcolor=blue,
    citecolor=blue
}

\newcommand{\argmin}{\arg\!\min}

\begin{document}

\pagenumbering{roman}
\thispagestyle{empty}
\begin{center}
{\bfseries 
  {\large\thesistitle}
  \vspace{1in}
  
 {\large {\bf DISSERTATION}}\\
  \vspace{.5in}
  
  \begin{doublespace}
  {\large  
  Submitted in Partial Fulfillment of\\
  the Requirements for\\
  the Degree of\\}
  \end{doublespace}
  \vspace{.5in}
  
  {\large DOCTOR OF PHILOSOPHY \\ Electrical and Computer Engineering Department}\\
  \vspace{.5in}
  
  at the \\
  \vspace{.2in}
  
  {\large
  NEW YORK UNIVERSITY\\
  \vspace{-0.05in}
  TANDON SCHOOL OF ENGINEERING\\
  }
  \vspace{.2in}
  
  by
  \vspace{.5in}

  {\large\thesisauthor}
  \vspace{.5in}

  {\large\graddate}
}

\end{center}

\newpage

%
\setcounter{page}{1}
\thispagestyle{empty}
\begin{center}
{\bfseries 
  {\large\thesistitle}
  \vspace{.25in}
  
  DISSERTATION\\
  \vspace{.25in}
  
  \begin{doublespace}
  Submitted in Partial Fulfillment of\\
  the Requirements for\\
  the Degree of\\
  \end{doublespace}
  \vspace{.25in}
  
  DOCTOR OF PHILOSOPHY (Electrical and Computer Engineering)\\
  \vspace{.25in}
  
  at the \\
  \vspace{.1in}
  
  {\large
  NEW YORK UNIVERSITY\\
  \vspace{-0.05in}
  TANDON SCHOOL OF ENGINEERING\\
  }
  \vspace{.2in}
  
  by
  \vspace{.3in}

  \thesisauthor
  \vspace{.3in}

  \graddate
}

\end{center}

\vspace{0.2in}

\noindent
\makebox[\textwidth]{\hfill\makebox[2.5in]{Approved: \hfill}}
\vspace{0.1in}

\noindent
\makebox[\textwidth]{\hfill\makebox[2.5in]{\hrulefill}}\\
\makebox[\textwidth]{\hfill\makebox[2.5in]{\hfill Department Chair Signature\hfill}}
\vspace{0.05in}

\noindent
\makebox[\textwidth]{\hfill\makebox[2.5in]{\hrulefill}}\\
\makebox[\textwidth]{\hfill\makebox[2.5in]{\hfill Date \hfill}}

\noindent
University ID:      N18100633\\
Net ID:  sm4841\\



Approved by the Guidance Committee:
\singlespacing

\underline{Major}: Electrical and Computer Engineering
\vspace{0.7in}

\noindent
\makebox[\textwidth]{\hfill\makebox[3in]{\hrulefill}}\\
\makebox[\textwidth]{\hfill\makebox[3in]{\hfill \textbf{Yao Wang}\hfill}}
\makebox[\textwidth]{\hfill\makebox[3in]{\hfill Professor of \hfill}}
\makebox[\textwidth]{\hfill\makebox[3in]{\hfill Electrical and Computer Engineering \hfill}}
\makebox[\textwidth]{\hfill\makebox[3in]{\rule[-4pt]{2in}{1pt}}}\\
\makebox[\textwidth]{\hfill\makebox[3in]{\hfill Date \hfill}}\\
\vspace{0.5in}

\noindent
\makebox[\textwidth]{\hfill\makebox[3in]{\hrulefill}}\\
\makebox[\textwidth]{\hfill\makebox[3in]{\hfill \textbf{Ivan Selesnick}\hfill}}
\makebox[\textwidth]{\hfill\makebox[3in]{\hfill Professor of \hfill}}
\makebox[\textwidth]{\hfill\makebox[3in]{\hfill Electrical and Computer Engineering \hfill}}
\makebox[\textwidth]{\hfill\makebox[3in]{\rule[-4pt]{2in}{1pt}}}\\
\makebox[\textwidth]{\hfill\makebox[3in]{\hfill Date \hfill}}\\
\vspace{0.5in}

\noindent
\makebox[\textwidth]{\hfill\makebox[3in]{\hrulefill}}\\
\makebox[\textwidth]{\hfill\makebox[3in]{\hfill \textbf{Carlos Fernandez-Granda}\hfill}}
\makebox[\textwidth]{\hfill\makebox[3in]{\hfill Assistant Professor of \hfill}}
\makebox[\textwidth]{\hfill\makebox[3in]{\hfill Mathematics and Data Science \hfill}}
\makebox[\textwidth]{\hfill\makebox[3in]{\rule[-4pt]{2in}{1pt}}}\\
\makebox[\textwidth]{\hfill\makebox[3in]{\hfill Date \hfill}}\\
\vspace{3.5in}

\doublespacing

\begin{center}
Microfilm or other copies of this dissertation are obtainable from
\vspace{4in}

UMI Dissertation Publishing\\
ProQuest CSA\\
789 E. Eisenhower Parkway\\
P.O. Box 1346\\
Ann Arbor, MI 48106-1346

\end{center}
\newpage

\section*{Vita}
\addcontentsline{toc}{section}{Vita}
Shervin Minaee was born in Tonekabon, Iran on Feb 15th,  1989.
He received his B.S. in Electrical Engineering (2012) from Sharif University of Technology, Tehran, Iran, and his M.S. in Electrical Engineering (2014) from New York University, New York, USA.
Since January 2015, he has been working toward his Ph.D. degree at the Electrical and Computer Engineering Department of New York University, under the supervision of Prof. Yao Wang.
During Summers 2014 and 2015, he worked as a research intern at Huawei Research Labs, at Santa Clara, CA, on computer vision applications in video compression.
In Summer 2016 he worked at AT\&T labs, at Middletown, NJ, on Deep learning application in costumer care automation.
During Summer 2017 he did an internship at Samsung Research America, working on personalized video commercials using a machine learning framework.
His research interests are machine learning applications in computer vision, medical image analysis and video processing.
\newpage

\section*{Acknowledgements}
\addcontentsline{toc}{section}{Acknowledgements}
I would like to thank my advisor, Prof. Yao Wang, for all her help, supports, and encouragement during my Ph.D. years. 
This thesis would not have been possible without her supports. 
Her knowledge and hardworking attitude inspired not only my vision toward research, but towards other aspects of life as well.
I would also like to express my sincerest gratitude to my Ph.D. guidance committee, Prof. Ivan Selesnick and Prof. Carlos Fernandez-Granda.
Their knowledge and insightful suggestions have been very helpful for my research.

During my Ph.D. study, I was lucky to work with several colleagues and friends in the video lab and NYU-Wireless including Amirali Abdolrashidi, Fanyi Duanmu, Meng Xu, Yilin Song, Xin Feng, Shirin Jalali, and Farideh Rezagah.
I am also grateful to other labmates: Xuan, Jen-wei, Yuanyi, Zhili, Eymen, Yuan, Chenge, Ran, Siyun, and Amir.

Besides my academic mentors and colleagues, I was very fortunate to work with Dr. Haoping Yu and Dr. Zhan Ma from Huawei Labs, Dr. Zhu Liu and Dr. Behzad Shahraray from AT\&T labs, and Dr. Imed Bouazizi and Dr. Budagavi from Samsung Research America, during my research internships.

I was also very lucky to find extraordinary friends during these years in New York. 
I owe them all my thanks, especially to Amirali, Milad, Bayan, Sara, Hossein, Mahdieh, Sima, Mohammad, Soheil, Ashkan, Hamed, Hadi, Mahla, Parisa, Zahra, Amir, Shayan, Negar, Khadijeh, Aria, Mehrnoosh, Nima, Alireza, Beril, Anil, Mathew, and Chris. 
I would also like to thank my old friends, Navid, Reza, Faramarz, Sina, Milad, Nima, Saeed, and Morteza.

Last but not least, I thank and dedicate this thesis to my family for their endless love and support.
\newpage

\vspace*{\fill}
\begin{center}
  \thesisdedication
\end{center}
\vfill
\newpage

\section*{}
\begin{center}
{\bfseries 
  {\bf ABSTRACT}\\
  \vspace{.25in}
  {\bf \thesistitle}\\  
  \vspace{.25in}
  {\bf by}\\  
  \vspace{.25in}
  {\bf \thesisauthor}\\
  \vspace{.25in}
  {\bf Advisor: Prof. \thesisadvisor, Ph.D.}\\
  \vspace{.25in}
  {\bf Submitted in Partial Fulfillment of the Requirements for}\\
  {\bf the Degree of Doctor of Philosophy (Computer Science)}\\
  \vspace{.25in}
  {\bf \graddate}  
  \vspace{.25in}
}
\end{center}
\addcontentsline{toc}{section}{Abstract}
Image foreground extraction is a classical problem in image processing and vision, with a large range of applications. 
In this dissertation, we focus on the extraction of text and graphics in mixed-content images, and design novel approaches for various aspects of this problem.

We first propose a sparse decomposition framework, which models the background by a subspace containing smooth basis vectors, and foreground as a sparse and connected component. We then formulate an optimization framework to solve this problem, by adding suitable regularizations to the cost function to promote the desired characteristics of each component. We present two  techniques to solve the proposed optimization problem, one based on alternating direction method of multipliers (ADMM), and the other one based on robust regression. Promising results are obtained for screen content image segmentation using the proposed algorithm.

We then propose a robust subspace learning algorithm for the representation of the background component using training images that could contain both background and foreground components, as well as noise. 
With the learnt subspace for the background, we can further improve the segmentation results, compared to using a fixed subspace.

Lastly, we investigate a different class of signal/image decomposition problem,  where only one signal component is active at each signal element. 
In this case, besides estimating each component, we need to find their supports, which can be specified by a binary mask.
We propose a mixed-integer programming problem, that jointly estimates the two components and their supports through an alternating optimization scheme.
We show the application of this algorithm on various problems, including image segmentation, video motion segmentation, and also separation of text from textured images.


\newpage

\tableofcontents

\listoffigures\addcontentsline{toc}{section}{\listfigurename}
\newpage

\listoftables\addcontentsline{toc}{section}{\listtablename}
\newpage

\pagenumbering{arabic} 


\chapter{Introduction}
Image segmentation is a classical problem in image processing and computer vision, which deals with partitioning the image into multiple similar regions.
Despite its long history, it is still not a fully-solved problem, due to the variation of images and segmentation objective. 
There are a wide range sub-categories of image segmentation, including semantic segmentation, instance-aware semantic segmentation, foreground segmentation in videos, depth segmentation, and foreground segmentation in still images.
This work develops various algorithms for foreground segmentation in screen content and mixed-content images, where the foreground usually refers to the text and graphics of the image, and multiple aspects of this problem are studied.
We start from sparsity based image segmentation algorithms, and then present a robust subspace learning algorithm to model the background component, and finally present an algorithm for masked signal decomposition.

\section{Motivation}
With the new categories of images such as screen content images, new techniques and modifications to previous algorithms are needed to process them. 
Screen content images refer to images appearing on the display  screens  of  electronic  devices  such  as  computers and  smart  phones.  
These  images  have  similar  characteristics  as  mixed  content  documents  (such  as  a  magazine page).  
They  often  contain  two  layers,  a  pictorial  smooth background  and  a  foreground  consisting  of  text  and  line graphics.
They show different characteristics from photographic images, such as sharp edges, and having less distinct colors in each region.
For example coding these images with traditional transform based coding algorithm, such as JPEG \cite{jpeg} and HEVC intra frame coding \cite{hevc}, may not be the best way for compressing these images, mainly because of the sharp   discontinuities in the foreground. 
In these cases, segmenting the image into two layers and coding them separately may be more efficient. 
Also because of different characteristics in the content of these images from photographic images, traditional image segmentation techniques may not work very well.

There have been some previous works for segmentation of mixed-content images, such  as  hierarchical  k-means  clustering  in  DjVu \cite{djvu}, and  shape  primitive extraction and coding (SPEC) \cite{spec}, but these works are mainly designed for images where the background is very simple and do not have a lot of variations and they usually do not work well when the background has a large color dynamic range, or there are regions in background with similar colors to foreground.
We propose different algorithms for segmentation of screen content and mixed-content images, by carefully addressing the problems and limitation of previous works.

\section{Contribution}
This thesis focuses on developing segmentation methods that can overcome the challenges of screen content and mixed content image segmentation and a suitable subspace learning scheme to improve the results.
More specifically, the following aims are pursued:

\begin{itemize}
    \item Developing a sparse decomposition algorithm for segmenting the foreground in screen content images, by modeling the background as a smooth component and foreground as a sparse and connected component.
Suitable regularization terms are added to  the optimization framework to impose the desired properties on each component.
    
    \item Developing a probabilistic algorithm for foreground segmentation using robust estimation.  
    RANSAC algorithm \cite{ransac}  is proposed for sampling pixels from image and building a model representation of the background, and treating the outliers of background model as foreground region.
    This algorithm is guaranteed to find the outliers (foreground pixels in image segmentation) with an arbitrary high probability.
    
    \item Developing a subspace learning algorithm for modeling the underlying signal and image in the presence of  structured  outliers  and  noise, using an  alternating  optimization  algorithm  for  solving this  problem,  which  iterates  between  learning  the  subspace  and finding  the  outliers.
    This algorithm is very effective for many of the real-world situations where acquiring clean signal/image is not possible, as it automatically detects the outliers and performs the subspace learning on the clean part of the signal.
    
    \item Proposing a novel signal decomposition algorithm for the case where different components are overlaid on top of each other, i.e. the value of each signal element is coming from one and only one of its components.   In this case, to separate signal components, we need to find a binary mask which shows the support of the corresponding component.
    We propose a mixed integer programming problem which jointly estimates both components and finds their supports. 
    We also propose masked robust principal component analysis (Masked-RPCA) algorithm that performs sparse and low-rank decomposition under overlaid model.    
    This is inspired by our masked signal decomposition framework, and can be thought as the extension of that framework for 1D signals, to 2D signals.
    

\end{itemize}

\section{Outline}
This thesis is organized as follows:

In Chapter 2, we discuss about two novel foreground segmentation algorithms, which we developed for screen content images, one using sparse decomposition and the other one using robust regression. 
The core idea of these two approaches is that the background component of screen content images can be modeled with a smooth subspace representation.
We also provide the details of the optimization approach for solving the sparse decomposition problem.
To demonstrate the performance of these algorithms, we prepared and manually labeled a dataset of over three hundred screen content image blocks, and evaluated the performance of these models on that dataset and compared with previous works.

In Chapter 3, we present the robust subspace learning approach, which is able to learn a smooth subspace representation for modeling the background layer in the presence of structured outliers and noise. This algorithm can not only be used for image segmentation, but it can also be used for subspace learning for any signal, which is heavily corrupted with outliers and noise.
We also study the application of this algorithm for text extraction in images with complicated background and provide the experimental results.

In Chapter 4, a new signal decomposition algorithm is proposed for the case where the signal components are overlaid on top of each other, rather than simple addition. In this case, beside estimating each signal component, we also need to estimate its support. 
We propose an optimization framework, which can jointly estimate both signal components and their supports.
We show that this scheme could significantly improve the segmentation results for text over textures.
We also show the application of this algorithm for motion segmentation in videos, and also 1D signal decomposition.
We then discuss about the extension of "Robust Principal Component Analysis (RPCA)" \cite{rpca}, to masked-RPCA, for doing sparse and low-rank decomposition under overlaid model.


Finally we conclude this thesis in Chapter 5, and discuss future research directions along the above topics.

\chapter{The Proposed  Foreground Segmentation Algorithms}
Image segmentation is the process of assigning a label to each image pixel, in a way that pixels with the same label have a similar property, such as similar color, or depth, or belonging to the same object.
One specific case of image segmentation is the foreground-background separation, which is to segment an image into 2 layers.
Given an image of size $N \times M$, there are $2^{NM}$ possible foreground segmentation results.
Foreground segmentation could deal with images or videos as input. 
We mainly focus on foreground segmentation in still images in this work, which could refer to segmenting an object of interest (such as the case in medical image segmentation), or segmenting the texts and graphics from mixed content images.
It is worth mentioning that foreground segmentation from video usually has a slight different objective from the image counterpart, which is to segment the moving objects from the background.

Foreground segmentation from still images has many applications in image compression \cite{coding1}-\cite{coding3}, text extraction \cite{text1}-\cite{text2}, biometrics recognition \cite{biometrics1}-\cite{biometrics3}, and medical image segmentation \cite{medseg1}-\cite{medseg2}.

In this chapter,  we first give an overview of some of the popular algorithms for foreground segmentation, such as algorithms based on k-means clustering \cite{djvu}, sparse and low-rank decomposition \cite{lowrank}, and shape primitive extraction and coding (SPEC) \cite{spec}, and discuss some of their difficulties in dealing with foreground segmentation in complicated images.
We then study the background modeling in the mixed-content images, and  propose two algorithms to perform foreground segmentation.

The proposed methods make use of the fact that the background in each block is usually smoothly varying and can be modeled well by a linear combination of a few smoothly varying basis functions, while the foreground text and graphics create sharp discontinuity. 
The proposed algorithms separate the background and foreground pixels by trying to fit background pixel values in the block into a smooth function using two different schemes. 
One is based on robust regression \cite{robreg}, where the inlier pixels will be considered as background, while remaining outlier pixels will be considered foreground. 
The second approach uses a sparse decomposition framework where the background and foreground layers are modeled with smooth and sparse components respectively.

The proposed methods can be used in different applications such as text extraction, separate coding of background and foreground for compression of screen content, and medical image segmentation.

\section{Related Works}
Different algorithms have been proposed in the past for foreground-background segmentation in mixed content document images and screen-content video frames such as hierarchical k-means clustering in DjVu \cite{djvu} and shape primitive extraction and coding (SPEC) \cite{spec}, sparse and low-rank decomposition \cite{lowrank}.
Also in \cite{text1}, and algorithm is proposed for text extraction in screen content images called scale and orientation invariant text segmentation.

The hierarchical k-means clustering method proposed in DjVu applies the k-means clustering algorithm with $k$=2 on blocks in multi-resolution. 
It first applies the k-means clustering algorithm on large blocks to obtain foreground and background colors and then uses them as the initial foreground and background colors for the smaller blocks in the next stage. It also applies some post-processing at the end to refine the results. This algorithm has difficulty in segmenting regions where background and foreground color intensities overlap and it is hard to determine whether a pixel belongs to the background or foreground just based on its intensity value.

In the shape primitive extraction and coding (SPEC) method, which was developed for segmentation of screen content, a two-step segmentation algorithm is proposed. In the first step the algorithm classifies each block of size $16 \times 16$ into either pictorial block or text/graphics based on the number of colors. If the number of colors is more than a threshold, 32, the block will be classified into pictorial block, otherwise to text/graphics.
In the second step, the algorithm refines the segmentation result of pictorial blocks, by extracting shape primitives (horizontal line, vertical line or a rectangle with the same color) and then comparing the size and color of the shape primitives with some threshold.
Because blocks containing smoothly varying background over a narrow range can also have a small color number,  it is hard to find a fixed color number threshold that can robustly separate pictorial blocks and text/graphics blocks. Furthermore, text and graphics  in screen content images typically have some variation in their colors, even in the absence of sub-pixel rendering. 
These challenges limit the effectiveness of SPEC.

In sparse and low-rank decomposition the image is assumed to consist of a low rank component and a sparse component, and low-rank decomposition is used to separate the low rank component from the sparse component. 
Because the smooth backgrounds in screen content images may not always have low rank and the foreground may happen to have low rank patterns (e.g. horizontal and vertical lines), applying such decomposition and assuming the low rank component is the background and the sparse component is the foreground may not always yield satisfactory results.

The above problems with prior approaches motivate us to design a segmentation algorithm that does not rely solely on the pixel intensity but rather exploits the smoothness of the background region, and the sparsity and connectivity of foreground. 
In other words, instead of looking at the intensities of individual pixels and deciding whether each pixel should belong to background or foreground, we first look at the smoothness of a group of pixels and then decide whether each pixel should belong to background or foreground.

\section{Background Modeling for Foreground  \\Separation}
One core idea of this work lies in the fact that if an image block only contains background pixels, it should be well represented with a few smooth basis functions. 
By well representation we mean that the approximated value at a pixel with the smooth functions should have an error less than a desired threshold at every pixel. Whereas if the image has some foreground pixels overlaid on top of a smooth background, those foreground pixels cannot be well represented using the smooth representation.
Since the foreground pixels cannot be modeled with this smooth representation they would usually have a large distortion by using this model.
Therefore the background segmentation task simplifies into finding the set of inlier pixels, which can be approximated well using this smooth model.
Now some questions arise here:
\begin{enumerate}
\item What is  a good class of smooth models that can represent the background layer accurately and compactly?
\item How can we derive the background model parameters such that they are not affected by foreground pixels, especially if we have many foreground pixels?
\end{enumerate}

For the first question, we divide each image into non-overlapping blocks of size $N \times N$, and represent each image block, denoted by $F(x,y)$, with a smooth model as a linear combination of a set of two dimensional smooth functions as $\sum_{k=1}^K \alpha_k P_k(x,y)$. Here low frequency two-dimensional DCT basis functions are used as $P_k(x,y)$, and the reason why DCT basis are used and how the number $K$ is chosen is explained at the end of this section. 
The 2-D DCT function is defined as Eq. \eqref{eq:ch2_1}:
\begin{equation}
P_{u,v}(x,y)= \beta_u \beta_v cos((2x+1)\pi u/2N) cos((2y+1)\pi v/2N) \label{eq:ch2_1}
\end{equation}
where $u$ and $v$ denote the frequency indices of the basis and $\beta_u$ and $\beta_v$ are normalization factors and $x$ and $y$ denote spatial coordinate of the image pixel.
We order all the possible basis functions in the conventional zig-zag order in the (u,v) plane, and choose the first $K$ basis functions.

The second question is kind of a chicken and egg problem: To find the model parameters we need to know which pixel belongs to the background and to know which pixel belongs to background we need to know what the model parameters are. 
One simple way is to define some cost function, which measures the goodness of fit between the original pixel intensities and the ones predicted by the smooth model, and then minimize the cost function. If we use the $\ell_p$-norm  of the fitting error ($p$ can be 0, 1, or 2), the problem can be written as:
\begin{equation}
\{\alpha_1^*,...,\alpha_K^*\}= \argmin_{\alpha_1,...,\alpha_K}  \sum_{x,y} |F(x,y)- \sum_{k=1}^K \alpha_k P_k(x,y)|^p \label{eq:ch2_2}
\end{equation}
We can also look at the 1D version of the above optimization problem by converting the 2D blocks of size $N \times N$ into a vector of length $N^2$, denoted by $\boldsymbol{f}$, by concatenating the columns and denoting $\sum_{k=1}^K \alpha_k P_k(x,y)$ as $ \textbf{P}\boldsymbol{\alpha}$ where $\textbf{P}$ is a matrix of size $N^2\times K$ in which the k-th column corresponds to the vectorized version of $P_k(x,y)$ and, $\boldsymbol{\alpha}=[\alpha_1,...,\alpha_K]^\text{T}$. 
Then the problem can be formulated as:
\begin{equation}
\alpha^*= \argmin_{\alpha} \| f -P\alpha \|_p \label{eq:ch2_3}
\end{equation}

If we use the $\ell_2$-norm (i.e. $p=2$) for the cost function, the problem is simply the least squares fitting problem and is very easy to solve. In fact it has a closed-form solution as below:
\begin{gather}
\alpha^*= \argmin_{\alpha} \| f-P\alpha \|_2 \Rightarrow \alpha= (P^T P)^{-1}P^T f \label{eq:ch2_4}
\end{gather}
But this formulation has a problem that the model parameters, $\boldsymbol{\alpha}$, can be adversely affected by foreground pixels. Especially in least-square fitting (LSF), by squaring the residuals, the larger residues will get larger weights in determining the model parameters. 
We propose two approaches for deriving the model parameters, which are more robust than LSF, in the following sections.

Now we explain why DCT basis functions are used to model the background layer.
To find a good set of bases for background, we first applied Karhunen-Loeve transform \cite{klt} to a training set of smooth background images, and the derived bases turn out to be very similar to 2D DCT and 2D orthonormal polynomials.
Therefore we compared these two sets of basis functions, the DCT basis and the orthonormal polynomials, which are known to be efficient for smooth image representation. 
The two dimensional DCT basis are outer-products of 1D DCT basis, and are well known to be very efficient for representing natural images.

To derive 2D orthonormal polynomials over an image block of size $N \times N$, we start with the $N$ 1D vectors obtained by evaluating the simple polynomials $f_n(x)=x^n$, at $x=\{1,2,...,N \}$, for $n=0, 1,.., N-1$ and orthonormalize them using Gram-Schmidt process to get $N$ orthonormal bases.
After deriving the 1D polynomial bases, we construct 2D orthonormal polynomial bases using the outer-product of 1D bases.

\begin{figure}[h]
\begin{center}
    \includegraphics [scale=0.5] {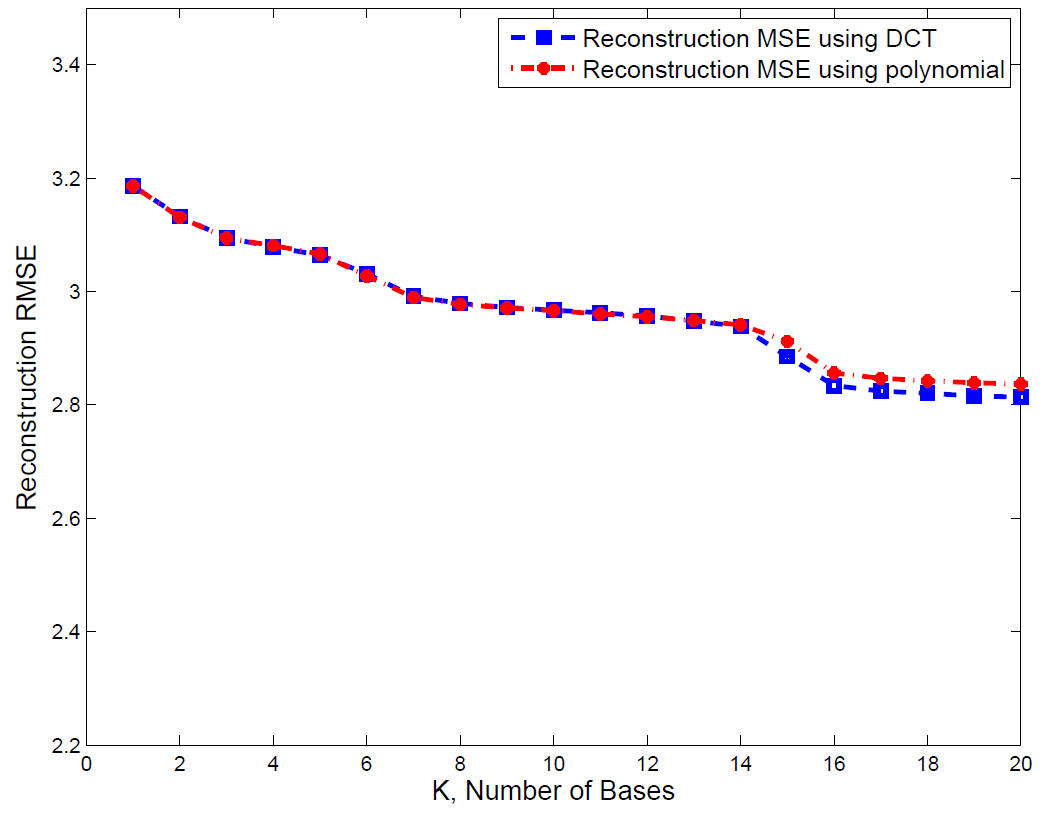}
\end{center}
  \caption{Background reconstruction RMSE vs. the number of bases.}  \label{fig:fig2_1}
\end{figure}

To compare DCT and orthonormal polynomial bases, we collected many smooth background blocks of size $64 \times 64$ from several images and tried to represent those blocks with the first $K$ polynomials and DCT basis functions in zigzag order. 
Because each block contains only smooth background pixels, we can simply apply least squares fitting to derive the model coefficients using Eq (2).
Then we use the resulting model to predict pixels' intensities and find the mean squared error (MSE) for each block. The reconstruction RMSEs (root MSE) as a function of the number of used bases, $K$, for both DCT and polynomials are shown in Figure~\ref{fig:fig2_1}. As we can see DCT has slightly smaller RMSE, so it is preferred over orthonormal polynomials.

It is worth to note that for more complicated background patterns, one could use the hybrid linear models \cite{hybrid1}, \cite{hybrid2} to represent the background using a union of subspaces.
But for screen content images, the background can usually be well-represented by a few low frequency DCT bases.

\section{First Approach: Robust Regression Based Segmentation}
Robust regression is a form of regression analysis, which is developed to overcome some limitations of traditional algorithms \cite{robreg}. The performance of most of the traditional regression algorithms can be significantly affected if the assumptions about underlying data-generation process are violated and they are highly sensitive to the presence of outliers. The outlier can be thought as any data-point or observation which does not follow the same pattern as the rest of observations. The robust regression algorithms are designed to find the right model for a dataset even in the presence of outliers. They basically try to remove the outliers from dataset and use the inliers for model prediction.

RANSAC \cite{ransac} is a popular robust regression algorithm. It is an iterative approach that performs the parameter estimation by minimizing the number of outliers (which can be thought as minimizing the $\ell_0$-norm).
RANSAC repeats two iterative procedures to find a model for a set of data. In the first step, it takes a subset of the data and derives the parameters of the model only using that subset. The cardinality of this subset is the smallest sufficient number to determine the model parameters. In the second step, it tests the model derived from the first step against the entire dataset to see how many samples can be modeled consistently. A sample will be considered as an outlier if it has a fitting error larger than a threshold that defines the maximum allowed deviation. RANSAC repeats the procedure a fixed number of times and at the end, it chooses the model with the largest consensus set (the set of inliers) as the optimum model. There is an analogy between our segmentation framework and model fitting in RANSAC. We can think of foreground pixels as outliers for the smooth model representing the background. Therefore RANSAC can be used to perform foreground segmentation task.

The proposed RANSAC algorithm for foreground/background segmentation of a block of size $N \times N$ is as follows:
\begin{enumerate}
\item Select a subset of $K$ randomly chosen pixels. Let us denote this subset by $S=\{(x_l,y_l), \ l=1,2,\ldots,K\}$.
\item Fit the model $\sum_{k=1}^K \alpha_k P_k(x,y) $ to the pixels $(x_l,y_l) \in S$ and find the $\alpha_k$'s. This is done by solving the set of $K$ linear equations $\sum_k \alpha_k P_k(x_l,y_l) = F(x_l,y_l), \ l=1,2,\ldots,K$.
\item Test all $N^2$ pixels $F(x,y)$ in the block against the fitted model. Those pixels that can be predicted with an error less than $\epsilon_{in}$ will be considered as the inliers.
\item Save the consensus set of the current iteration if it has a larger size than the largest consensus set identified so far.
\item If the inlier ratio, which is the ratio of inlier pixels to the total number of pixels, is more than 95\%, stop the algorithm.
\item Repeat this procedure up to $M_{\rm iter}$ times.
\end{enumerate}
After this procedure is finished, the pixels in the largest consensus set will be considered as inliers or equivalently background. 
The final result of RANSAC can be refined by refitting over all inliers once more  and finding all pixels with error less than $\epsilon_{in}$.
To boost the speed of the RANSAC algorithm, we stop once we found a consensus set which has an inlier ratio more than $0.95$. 


\section{Second Approach: Sparse Decomposition}
Sparse representation has been used for various applications in recent years, including face recognition, super-resolution, morphological component analysis, denosing, image restoration and sparse coding \cite{sr1}-\cite{sr8}.
In this work, we explored the application of sparse decomposition for image segmentation.
As we mentioned earlier, the smooth background regions can be well represented with a few smooth basis functions, whereas the high-frequency component of the image belonging to the foreground, cannot be represented with this smooth model. 
But using the fact that foreground pixels occupy a relatively small percentage of the images we can model the foreground with a sparse component overlaid on background. 
Therefore it is fairly natural to think of mixed content image as a superposition of two components, one smooth and the other one sparse, as shown below:
\begin{equation}
F(x,y)= \sum_{k=1}^K \alpha_k P_k(x,y) + S(x,y) \label{eq:ch2_5}
\end{equation}
where $\sum_{i=1}^K \alpha_i P_i(x,y)$ and $S(x,y)$ correspond to the smooth background region and foreground pixels respectively. 
Therefore we can use sparse decomposition techniques to separate these two components.
After decomposition, those pixels with large value in the $S$ component will be considered as foreground.
We will denote this algorithm as "SD", for notation brevity.

To have a more compact notation, we will look at the 1D version of this problem. Denoting the 1D version of $S(x,y)$ by $s$, 
Eq. \eqref{eq:ch2_5} can be written as:
\begin{equation}
f= {P\alpha}+s \label{eq:ch2_6}
\end{equation}

Now to perform image segmentation, we need to impose some prior knowledge about background and foreground to our optimization problem. 
Since we do not know in advance how many basis functions to include for the background part, we allow the model to choose from a large set of bases that we think are sufficient to represent the most "complex" background, while minimizing coefficient $\ell_0$ norm to avoid overfitting of the smooth model on the foreground pixels. Because if we do not restrict the parameters, we may end up with a situation that even some of the foreground pixels are represented with this model (imagine the case that we use a complete set of bases for background representation).
Therefore the number of nonzero components of $\alpha$ should be small (i.e. $\| \alpha \|_0$ should be small).
On the other hand we expect the majority of the pixels in each block to belong to the background component, therefore the number of nonzero components of $s$ should be small.
And the last but not the least one is that foreground pixels typically form connected components in an image, therefore we can add a regularization term which promotes the connectivity of foreground pixels. Here we used total variation of the foreground component to penalize isolated points in foreground.
Putting all of these priors together we will get the following optimization problem:
\begin{equation}
\begin{aligned}
& \underset{s, \alpha}{\text{minimize}}
& & \|\alpha \|_0+ \lambda_1 \| s \|_0+ \lambda_2 TV(s)   \\
& \text{subject to}
& &  f=P \alpha+s
\end{aligned} \label{eq:ch2_7}
\end{equation}
where $\lambda_1$ and $\lambda_2$ are some constants which need to be tuned. 
For the first two terms since $\ell_0$ is not convex, we use its approximated $\ell_1$ version to have a convex problem.
For the total variation we can use either the isotropic or the anisotropic version of 2D total variation \cite{tv}. To make our optimization problem simpler, we have used the anisotropic version in this algorithm, which is defined as:
\begin{equation}
\begin{aligned}
TV(s)=  \sum_{i,j} |S_{i+1,j}-S_{i,j}|+|S_{i,j+1}-S_{i,j}|
\end{aligned} \label{eq:ch2_8}
\end{equation}
After converting the 2D blocks into 1D vector, we can denote the total variation as below:
\begin{equation}
\begin{aligned}
TV(s)= \| D_xs \|_1+\| D_ys \|_1= \|Ds\|_1 \label{eq:ch2_9}
\end{aligned}
\end{equation}
where $D=[ D_x',D_y']'$. Then we will get the following problem:
\begin{equation}
\begin{aligned}
& \underset{s, \alpha}{\text{minimize}}
& & \|\alpha \|_1+ \lambda_1 \| s \|_1+ \lambda_2 \|Ds\|_1   \\
& \text{subject to}
& &  P \alpha+s=f
\end{aligned} \label{eq:ch2_10}
\end{equation}
From the constraint in the above problem, we get $s= f-P\alpha$ and then we derive the following unconstrained problem:
\begin{equation}
\begin{aligned}
& \underset{ \alpha}{\text{min}}
& & \|\alpha \|_1+ \lambda_1 \| f-P\alpha \|_1+ \lambda_2 \|Df-DP\alpha\|_1
\end{aligned} \label{eq:ch2_11}
\end{equation}

This problem can be solved with different approaches, such as alternating direction method of multipliers (ADMM) \cite{admm}, majorization minimization \cite{seles1}, proximal algorithm \cite{patrick1} and iteratively reweighted least squares minimization \cite{irwl}. 
Here we present the formulation using ADMM algorithm.

\subsection{ADMM for solving L1 optimization}
ADMM is a variant of the augmented Lagrangian method that uses the partial update for dual variable. 
It has been widely used in recent years, since it works for more general classes of problems than some other methods such as gradient descent (for example it works for cases where the objective function is not differentiable). 
To solve Eq. \eqref{eq:ch2_11} with ADMM, we introduce the auxiliary variable $y,z$ and $x$  and convert the original problem into the following form:
\begin{equation}
\begin{aligned}
& \underset{\alpha, y, z, x}{\text{minimize}}
& & \|y \|_1+ \lambda_1 \| z \|_1+ \lambda_2 \|x\|_1   \\
& \text{subject to} & &  y=\alpha \\
& & & z=f-P \alpha \\
& & & x=Df-DP\alpha
\end{aligned} \label{eq:ch2_12}
\end{equation}
Then the augmented Lagrangian for the above problem can be formed as:
\begin{equation}
\begin{aligned}
&L_{\rho_1,\rho_2,\rho_3}(\alpha,y,z,x)= \|y \|_1+ \lambda_1 \| z \|_1+ \lambda_2 \|x\|_1+ u_1^t(y- \alpha)+ u_2^t(z+P\alpha-f)+   \\
& u_3^t(x+DP\alpha-Df)+\frac{\rho_1}{2} \| y- \alpha \|_2^2+ \frac{\rho_2}{2} \| z+P\alpha-f \|_2^2+\frac{\rho_3}{2} \| x+DP\alpha-Df \|_2^2 
\end{aligned} \label{eq:ch2_13}
\end{equation}
where $u_1$, $u_2$ and $u_3$ denote the dual variables.
Then, we can find the update rule of each variable by setting the gradient of the objective function w.r.t. to the primal variables to zero and using dual descent for dual variables.
The detailed variable updates are shown in Algorithm 1.

\begin{algorithm}
\caption{pseudo-code for ADMM updates of problem in Eq. \eqref{eq:ch2_13}}
\label{euclid}
\begin{algorithmic}[1]
  \FOR{\texttt{$k$=1:$k_{max}$}}
   \STATE  $\alpha^{k+1}= \underset{\alpha}{\text{argmin}} \ L_{\rho_{1:3}}(\alpha,y^k,z^k,x^k, u_1^k, u_2^k, u_3^k) = A^{-1} \big[ u_1^{k}-P^tu_2^{k}-P^tD^tu_3^{k}+\rho_1y^k +\rho_2P^t(f-z^{k})+\rho_3P^tD^t(Df-x^{k}) \big] $ \vspace{0.2cm}  
   \STATE $y^{k+1}= \underset{y}{\text{argmin}} \ L_{\rho_{1:3}}(\alpha^{k+1},y,z^k,x^k, u_1^k, u_2^k, u_3^k)= \text{Soft}(\alpha^k- \frac{1}{\rho_1} u_1^{k},\frac{1}{\rho_1}) $ 
   \STATE $z^{k+1}= \underset{z}{\text{argmin}} \ L_{\rho_{1:3}}(\alpha^{k+1},y^{k+1},z,x^k, u_1^k, u_2^k, u_3^k) = \text{Soft}(f-P\alpha^{k+1}-\frac{1}{\rho_2} u_2^{k},\frac{\lambda_1}{\rho_2}) $ 
   \STATE $x^{k+1}= \underset{x}{\text{argmin}} \ L_{\rho_{1:3}}(\alpha^{k+1},y^{k+1},z^{k+1},x, u_1^k, u_2^k, u_3^k) = \text{Soft}(Df-DP\alpha^{k+1}-\frac{1}{\rho_3} u_3^{k},\frac{\lambda_2}{\rho_3}) $ \vspace{0.2cm}  
        \STATE $u_1^{k+1}= u_1^{k}+ \rho_1 (y^{k+1}-\alpha^{k+1})$     \vspace{0.2cm}    
        \STATE $u_2^{k+1}= u_2^{k}+ \rho_2 (z^{k+1}+P\alpha^{k+1}-f)$   \vspace{0.2cm} 
        \STATE $u_3^{k+1}= u_3^{k}+ \rho_3 (x^{k+1}+DP\alpha^{k+1}-Df)$   \vspace{0.25cm}        
\ENDFOR
\\ \vspace{0.4cm} Where $A=(\rho_3 P^tD^tDP+\rho_2P^tP+\rho_1 I)$          
\end{algorithmic}
\end{algorithm}
\vspace{3cm}Here $\text{Soft}(.,\lambda)$ denotes the soft-thresholding operator applied elementwise and is defined as:
\begin{gather}
\text{Soft}(x,\lambda)= \text{sign}(x) \ \text{max}(|x|-\lambda,0) \label{eq:soft}
\end{gather}
The setting for the parameters $\rho_{1:3}$ and the regularization weights $\lambda_{1:3}$ are explained in section IV.

After finding the values of $\alpha$, we can find the sparse component as $s=f-P\alpha$. 
Then those pixels with values less than an inlier threshold $\epsilon_{in}$ in $s$ will be considered as foreground.

\begin{figure}[h]
\begin{center}
    \includegraphics [scale=0.8] {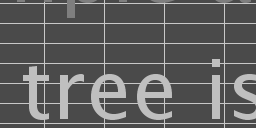} 
    \includegraphics [scale=0.8] {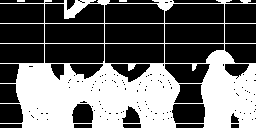} 
\end{center}
\vspace{0.02cm}      
\begin{center}
    \includegraphics [scale=0.8] {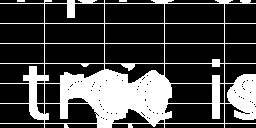}     
    \includegraphics [scale=0.8] {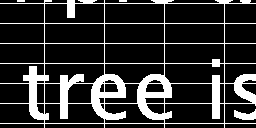}     
\end{center}
  \caption{ The original image (first row), the segmented foreground using least square fitting (second row), least absolute deviation (third row) and sparse decomposition (last row).} \label{fig:fig2_2}
\end{figure}

To show the advantage of minimizing $\ell_1$ over $\ell_2$, and also sparse decomposition over both $\ell_1$ and $\ell_2$ minimization approaches, we provide the segmentation result using least square fitting (LSF), least absolute deviation fitting (LAD) \cite{lad}, and also sparse decomposition (SD) framework for a sample image consists of foreground texts overlaid on a constant background. 
The original image and the segmentation results using LSF, LAD and SD are shown in Figure~\ref{fig:fig2_2}. 

\begin{figure}[ht]
\begin{center}
    \includegraphics [scale=0.7] {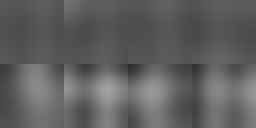} 
    \includegraphics [scale=0.7] {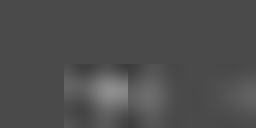}     
    \includegraphics [scale=0.7] {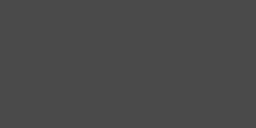}     
\end{center}
  \caption{ The reconstructed background layer using least square fitting (top image), least absolute deviation (middle image) and sparse decomposition (bottom image)} \label{fig:fig2_3}
\end{figure}

The reconstructed smooth model by these algorithms are shown in Figure~\ref{fig:fig2_3}. All methods used 10 DCT basis for representing the background and the same inlier threshold of 10 is used here.

As we can see, the smooth model derived by LSF is largely affected by the foreground pixels. 
The ideal smooth model should have the same color as actual background (here gray), but because of the existence of many text pixels with white color the LSF solution tries to find a trade-off between fitting the texts and fitting the actual background, which results in inaccurate representation of either background or foreground in the regions around text. Therefore the regions around texts will have error larger than the inlier threshold and be falsely considered as the foreground pixels. 
The smooth model produced by the LAD approach was less affected by the foreground pixels than the LSF solution, because it minimizes the $\ell_1$ norm of the fitting error $s$. However, in blocks where there is a larger percentage of foreground pixels (bottom middle  and right regions),  LAD solution is still adversely affected by the foreground pixels. The SD approach yielded accurate solution in this example, because it considers  the $\ell_1$ norm of the fitting coefficient, $\alpha$, and the TV norm of $s$, in addition to  the  $\ell_1$ norm of $s$. Although the LAD solution leads to smaller $\ell_1$ norm of the fitting error, it also  leads to a much larger $\ell_1$ norm of $\alpha$ as well.  By minimizing all three terms, the SD solution obtains  a background model that uses predominantly only the DC basis, which represented the background accurately.

To confirm that the SD solution indeed has a smaller $\ell_1$ norm of $\alpha$, we show below the derived $\alpha$ values using each scheme in \eqref{eq:ch2_14}. As we can see the derived $\alpha$ by SD has much smaller $\ell_0$ and $\ell_1$ norm than the other two.
\begin{equation}
\begin{aligned}
& \alpha^{LSF} \ = (7097,   -359,    19,   -882,    177,  -561,    863,  953,     113,   - 554)  \\
& \alpha^{LAD}= (5985,   -599,    201,  -859,  -13,   -96,    365,    39,    464,   -411)  \\
& \alpha^{SD}  \ \ = (4735,   -1,   0,   -4,   0,   -1,   0,    0,    0,    1) 
\end{aligned} \label{eq:ch2_14}
\end{equation}

Both of the proposed segmentation algorithms performs very well on majority of mixed content images, but for blocks that can be easily segmented with other methods, RANSAC/SD may be an overkill. 
Therefore, we propose a segmentation algorithm that has different modes in the next Section.

\section{Overall Segmentation Algorithms}
We propose a segmentation algorithm that mainly depends on RANSAC/SD but it first checks if a block can be segmented using some simpler approaches and it goes to RANSAC/SD only if the block cannot be segmented using those approaches. These simple cases belong to one of these groups: pure background block, smoothly varying background only and text/graphic overlaid on constant background.
 
Pure background blocks are those in which  all pixels have similar intensities. These kind of blocks are common in screen content images. These blocks can be detected by looking at the standard deviation or  maximum absolute deviation of pixels' intensities. If the standard deviation is less than some threshold we declare that block as pure background.

Smoothly varying background only is a block in which the intensity variation over all pixels can be modeled well by a smooth function. Therefore we try to fit $K$ DCT basis to all pixels using least square fitting. If all pixels of that block can be represented with an error less than a predefined threshold, $\epsilon_{in}$, we declare it as smooth background.

The last group of simple cases is text/graphic overlaid on constant background. 
The images of this category usually have zero variance (or very small variances) inside each connected component.  
These images usually have a limited number of different colors in each block (usually less than 10) and the intensities in different parts are very different. We calculate the percentage of each different color in that block and the one with the highest percentage will be chosen as background and the other ones as foreground. 

When a block does not satisfy any of the above conditions, RANSAC/SD will be applied to separate the background and the foreground. If the segmentation is correct, the ratio of background pixels over the total number of pixels should be fairly large (greater than at least half ). When the ratio is small, the background of the block may be too complex to be presented by the adopted smooth function model. This may also happen when the block sits at the intersection of two smooth backgrounds. To overcome these problems, we apply the proposed method recursively using a quadtree structure. When the inlier ratio of the current block is less than $\epsilon_2$, we divide it into 4 smaller blocks and apply the proposed algorithm on each smaller block, until the smallest block size is reached.\\
The overall segmentation algorithm is summarized as follows:
\begin{enumerate}
\item Starting with block size $N=64$, if the standard deviation of pixels' intensities is less than $\epsilon_1$ (i.e. pixels in the block have very similar color intensity), then declare the entire block as background.
If not, go to the next step;
\item Perform least square fitting using all pixels. If all pixels can be predicted with an error less than $\epsilon_{in}$, declare the entire block as background. If not, go to the next step;
\item If the number of different colors (in terms of the luminance value) is less than $T_1$ and the intensity range is above $R$, declare the block as text/graphics over a constant background and find the background as the color in that block with the highest percentage of pixels. If not, go to the next step;
\item Use RANSAC/SD to separate background and foreground using the luminance component only. Verify that the corresponding chrominance components of background pixels can also be fitted using $K$ basis functions with an error less than $\epsilon_{in}$. If some of them cannot be fitted with this error, remove them from inliers set.
If the percentage of inliers is more than a threshold $\epsilon_2$ or $N$ is equal to 8, the inlier pixels are selected as background. If not go to the next step;
\item Decompose the current block of size $N \times N$ into 4 smaller blocks of size $\frac{N}{2} \times \frac{N}{2}$ and run the segmentation algorithm for all of them. Repeat until $N=8$.
\end{enumerate}

\begin{figure}[h]
\begin{center}
    \includegraphics [scale=0.36] {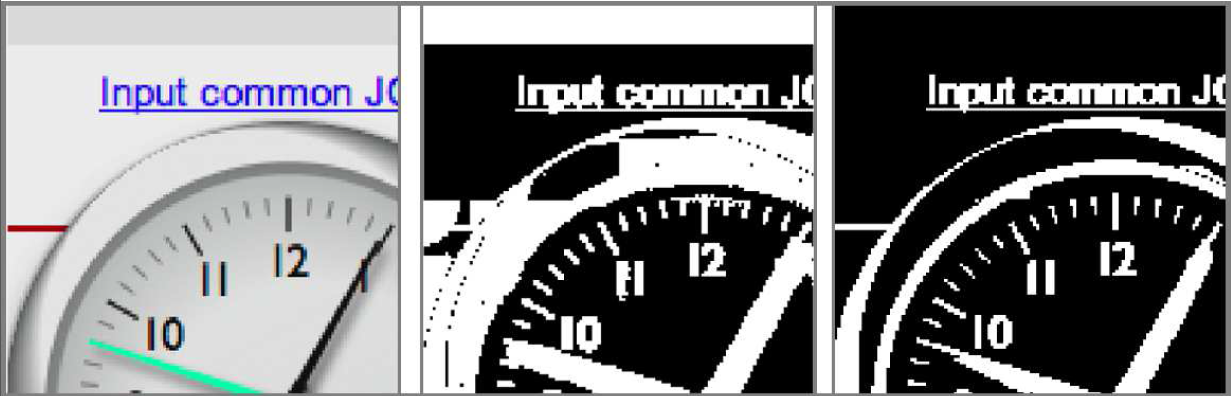} 
\end{center}
  \caption{Segmentation result for a sample image, middle and right images denote foreground map without and with quad-tree decomposition  using the RANSAC as the core algorithm. } \label{fig:fig2_4}
\end{figure}

To show the advantage of quad-tree decomposition, we provide an example of the segmentation map without and with quad-tree decomposition in Figure~\ref{fig:fig2_4}. 
As we can see, using quadtree decomposition we get much better result compared to the case with no decomposition. When we do not allow a $64 \times 64$ block to be further divided, only a small percentage of pixels can be represented well by a smooth function, leaving many pixels as foreground. 
It is worth mentioning that the gray region on the top of the image is considered as foreground in the segmentation result without using quadtree decomposition. This is because the first row of $64 \times 64$ blocks contain two smooth background regions with relatively equal size.

\section{Experimental Results}
To enable rigorous evaluation of different algorithms, we have generated an annotated dataset consisting of 328 image blocks of size $64\times 64$, extracted from sample frames from HEVC test sequences for screen content coding \cite{SCC_data}. 
The ground truth foregrounds for these images are extracted manually by the author and then refined independently by another expert. This dataset is publicly available at \cite{our_dataset}.

Table 2.1 summarizes the parameter choices in the proposed algorithms.
The largest block size is chosen to be $N$=64, which is the same as the largest CU size in HEVC standard.
The thresholds used for preprocessing (steps 1-3) should be chosen conservatively to avoid segmentation errors. In our simulations, we have chosen them as $\epsilon_1=3$, $T_1=10$ and $R=50$, which achieved a good trade off between computation speed and segmentation accuracy. 
For the RANSAC algorithm, the maximum number of iteration is chosen to be 200.
For the sparse decomposition algorithm, the weight parameters in the objective function are tuned by testing on a validation set and are set to be $\lambda_1=10$ and $\lambda_2=4$.
The ADMM algorithm described in Algorithm 1 is implemented in MATLAB, which the code available in \cite{our_dataset}.
The number of iteration for ADMM is chosen to be 50 and the parameter $\rho_1$, $\rho_2$ and $\rho_3$ are all set to 1 as suggested in \cite{boyd}. 

\begin{table}[h]
\centering
  \caption{Parameters of our implementation}
  \centering
\begin{tabular}{|m{9cm}|m{2cm}|m{2cm}|}
\hline
Parameter description &  \ Notation & \ \ \ Value\\
\hline
Maximum block size & \ \ \ \ \ $N$ & \ \ \ \ \  64 \\
\hline
Inlier distortion threshold & \ \ \ \ \ $\epsilon_{in}$ & \ \ \ \ \ 10 \\ 
\hline
Background standard deviation threshold & \ \ \ \ \ $\epsilon_{1}$ & \ \ \ \ \ 3 \\
\hline
Qaud-tree decomposition threshold & \ \ \ \ \ $\epsilon_{2}$ & \ \ \ \ \ 0.5 \\
 \hline
Max number of colors for text over constant background & \ \ \ \ \ $T_1$ & \ \ \ \ \ 10 \\ 
\hline
Min intensity range for text over constant background & \ \ \ \ \ $R$ & \ \ \ \ \ 50 \\
\hline
Sparsity weight in SD algorithm & \ \ \ \ \ $\lambda_1$  & \ \ \ \  \ 10 \\
\hline
Total variation weight in SD algorithm & \ \ \ \ \ $\lambda_2$  & \ \ \ \ \ \ 4 \\
\hline
\end{tabular}
\label{TblComp}
\end{table}

To find the number of DCT basis functions, $K$, and inlier threshold, $\epsilon_{in}$, for RANSAC and sparse decomposition, we did a grid search over pairs of these parameters, in the range of 6 to 10 for $K$ and 5 to 15 for $\epsilon_{in}$, on some training images, and then chose the one which achieved the best result in terms of average F1-score. 
The parameter values that resulted in the best F1-score on our training images are shown in Table 2.2.
\begin{table}[h]
\centering
  \caption{The chosen values for the inlier threshold and number of bases}
  \centering
  \vspace{0.2cm}
\begin{tabular}{|m{5cm}|m{1cm}|m{2cm}|m{1.5cm}|}
\hline
Segmentation Algorithm  &  \  LAD & \ RANSAC & \ \ \ \ SD \\
\hline
Inlier threshold & \ \ 10 & \ \ \ \ 10 & \ \ \ \  10 \\
\hline
Number of bases & \ \ \ 6 & \ \ \ \ 10 & \ \ \ \ 10 \\
\hline
\end{tabular}
\label{TblComp}
\end{table}

Before showing the segmentation result of the proposed algorithms on the test images, we illustrate how the segmentation result varies by changing different parameters in RANSAC algorithm. 
The sparse decomposition algorithm would also have the same behavior.

\begin{figure}[h]
\begin{center}
    \includegraphics [scale=0.22] {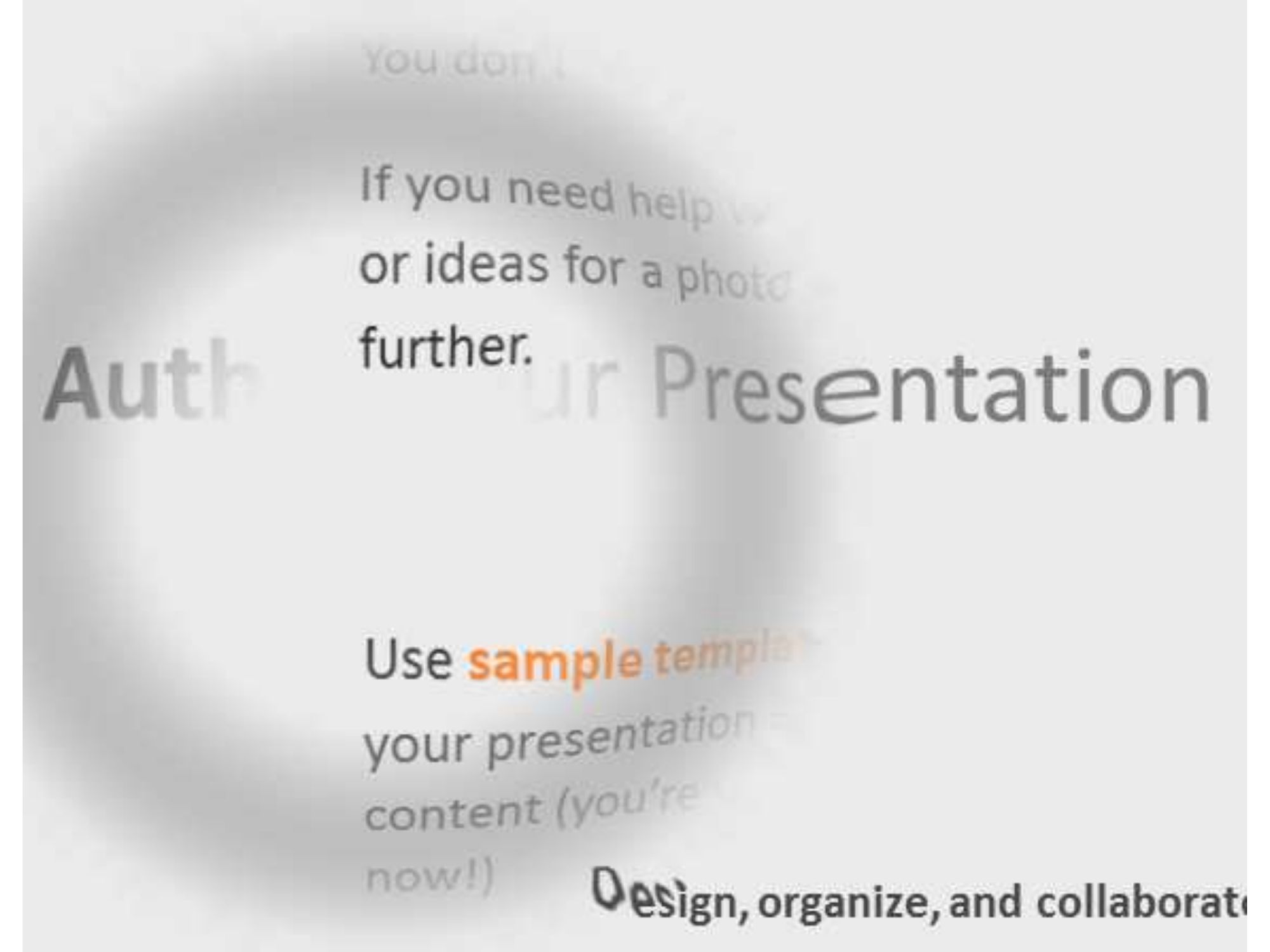} 
    \includegraphics [scale=0.22] {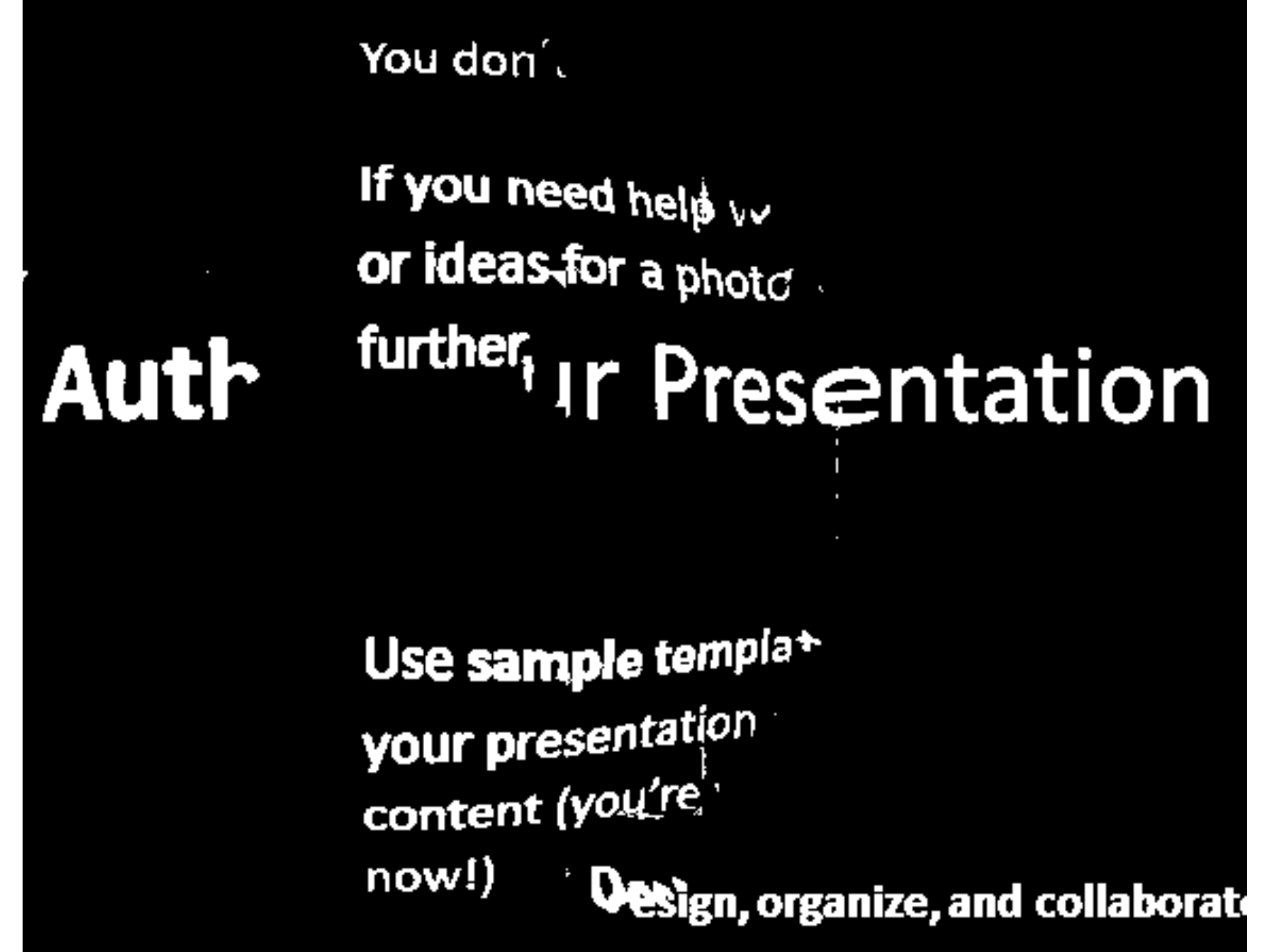} 
    \includegraphics [scale=0.22] {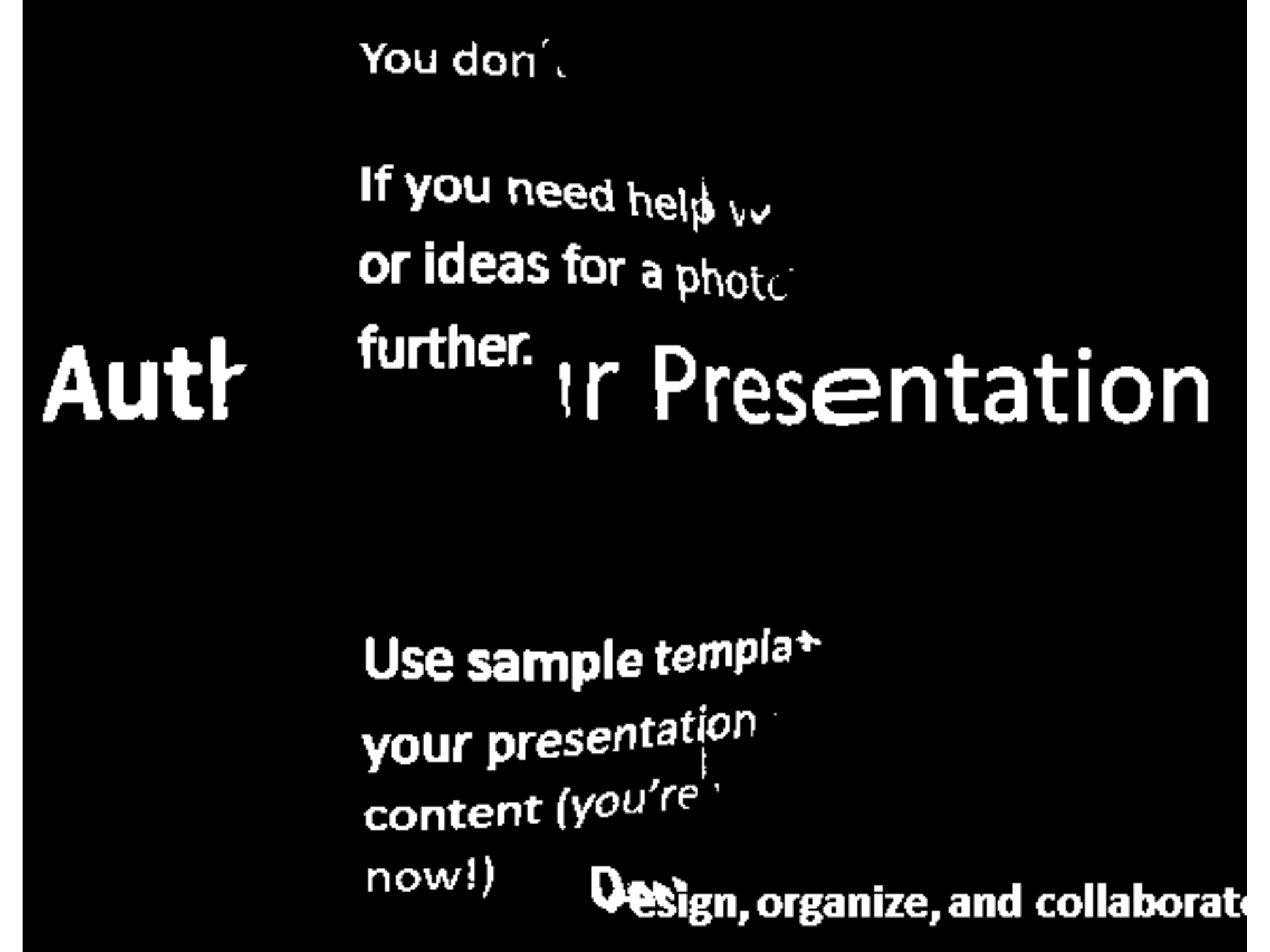}  
\end{center}
\begin{center}
    \includegraphics [scale=0.22] {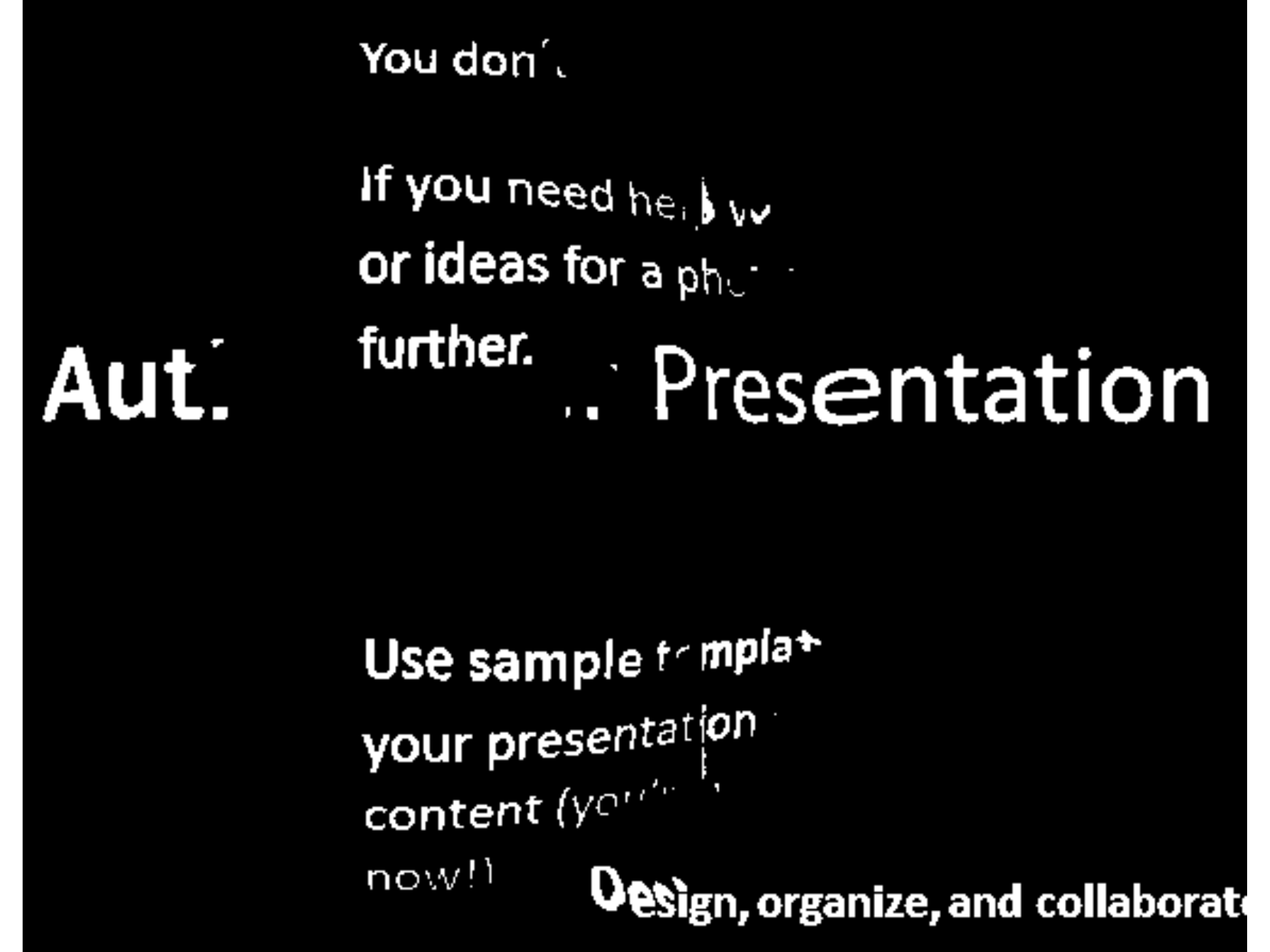}     
    \includegraphics [scale=0.22] {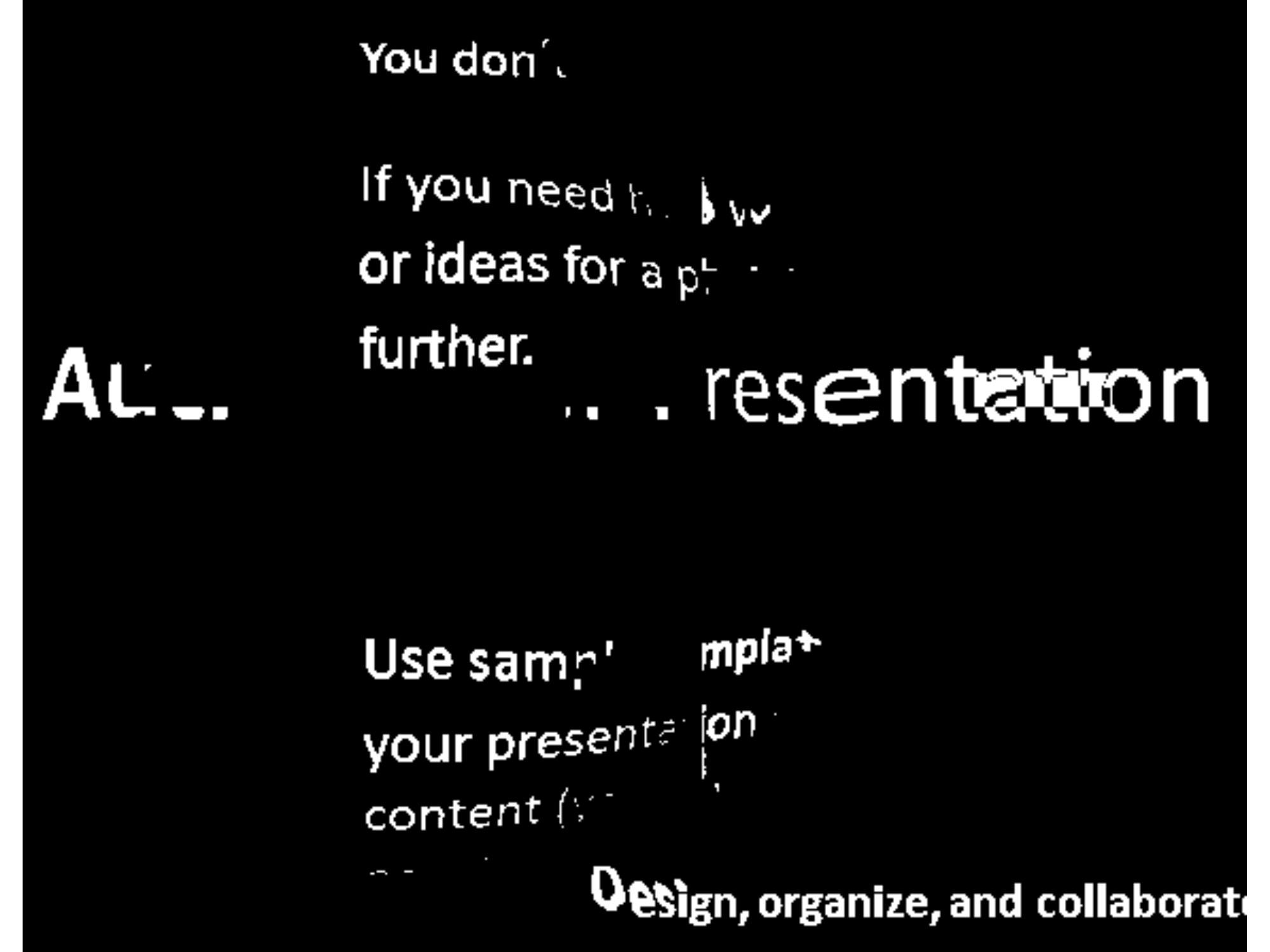}     
    \includegraphics [scale=0.22] {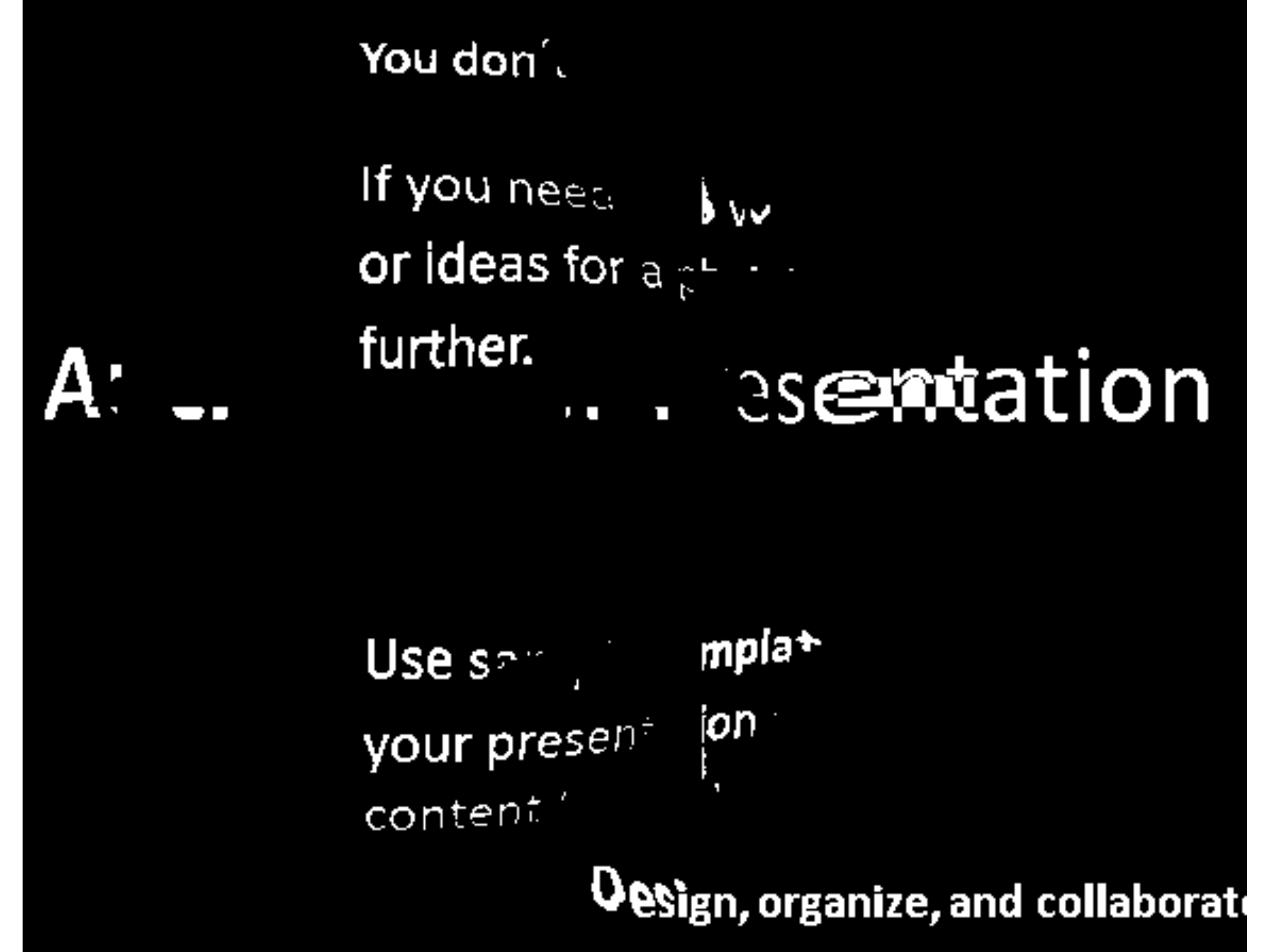}     
\end{center}
  \caption{ Segmentation results of the RANSAC method by varying the inlier threshold $\epsilon_{in}$. The foreground maps from left to right and top to bottom are obtained with $\epsilon_{in}$ setting to 5, 10, 25, 35, and 45, respectively.} \label{fig:fig2_5}
\end{figure}

To evaluate the effect of the distortion threshold, $\epsilon_{in}$, for inlier pixels in the final segmentation result, we show the foreground map derived by several different thresholds in Figure~\ref{fig:fig2_5}.  As we can see by increasing the threshold more and more pixels are considered as background.

\begin{figure}[h]
\begin{center}
    \includegraphics [scale=0.22] {2Original_Image-eps-converted-to.pdf} 
    \includegraphics [scale=0.22] {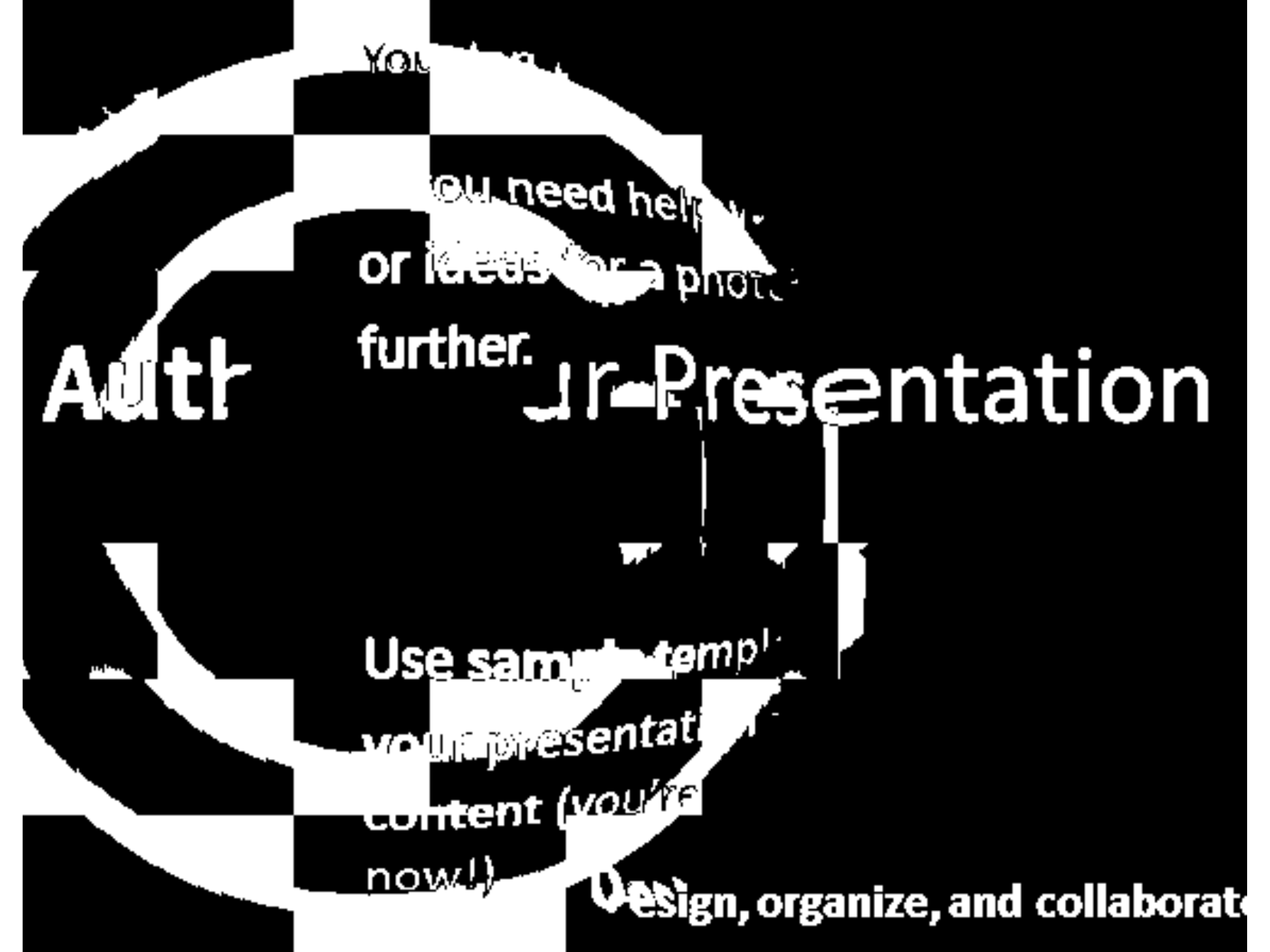} 
    \includegraphics [scale=0.22] {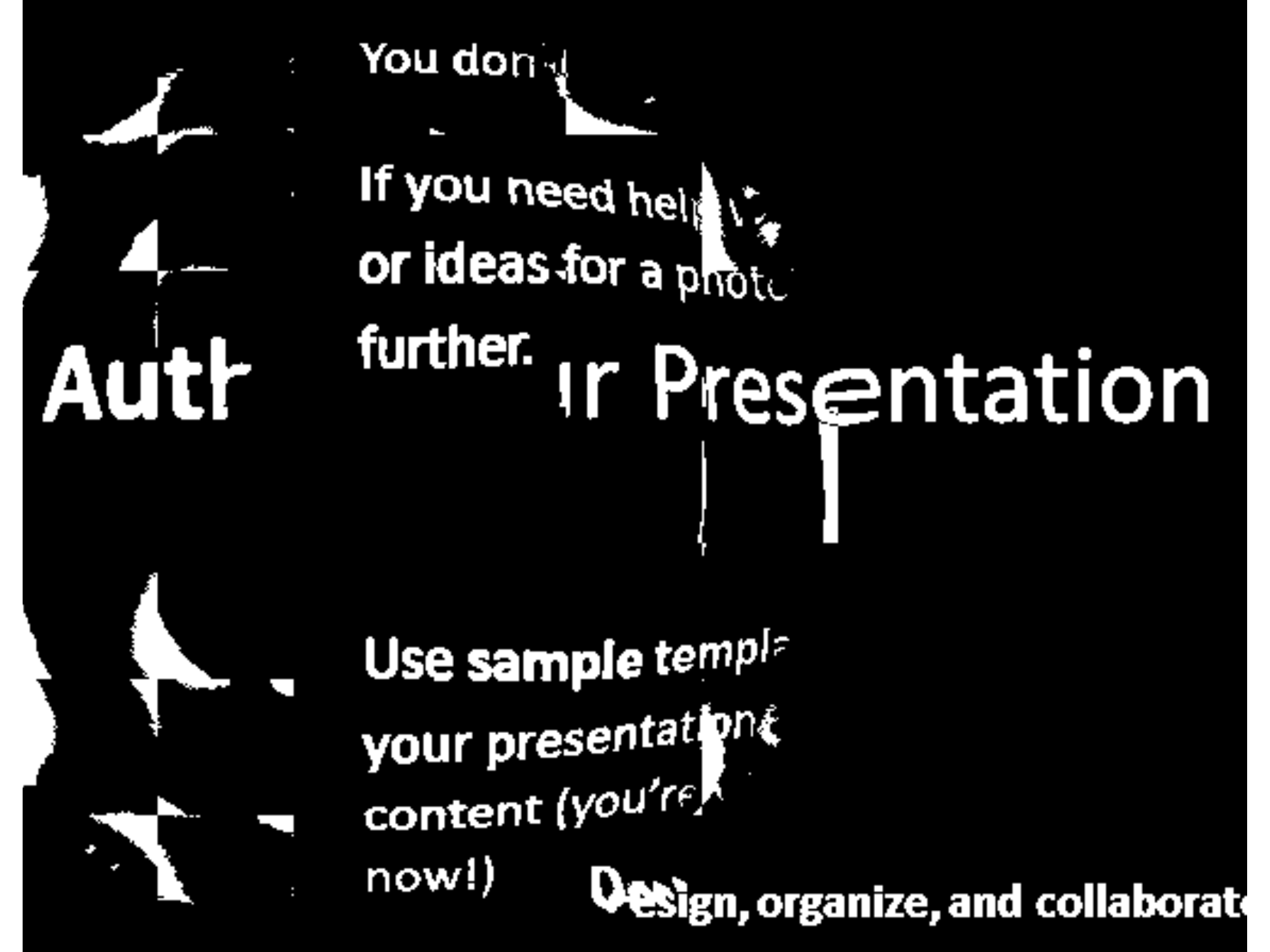}  
\end{center}
\begin{center}
    \includegraphics [scale=0.22] {2_map_basis10-eps-converted-to.pdf}     
    \includegraphics [scale=0.22] {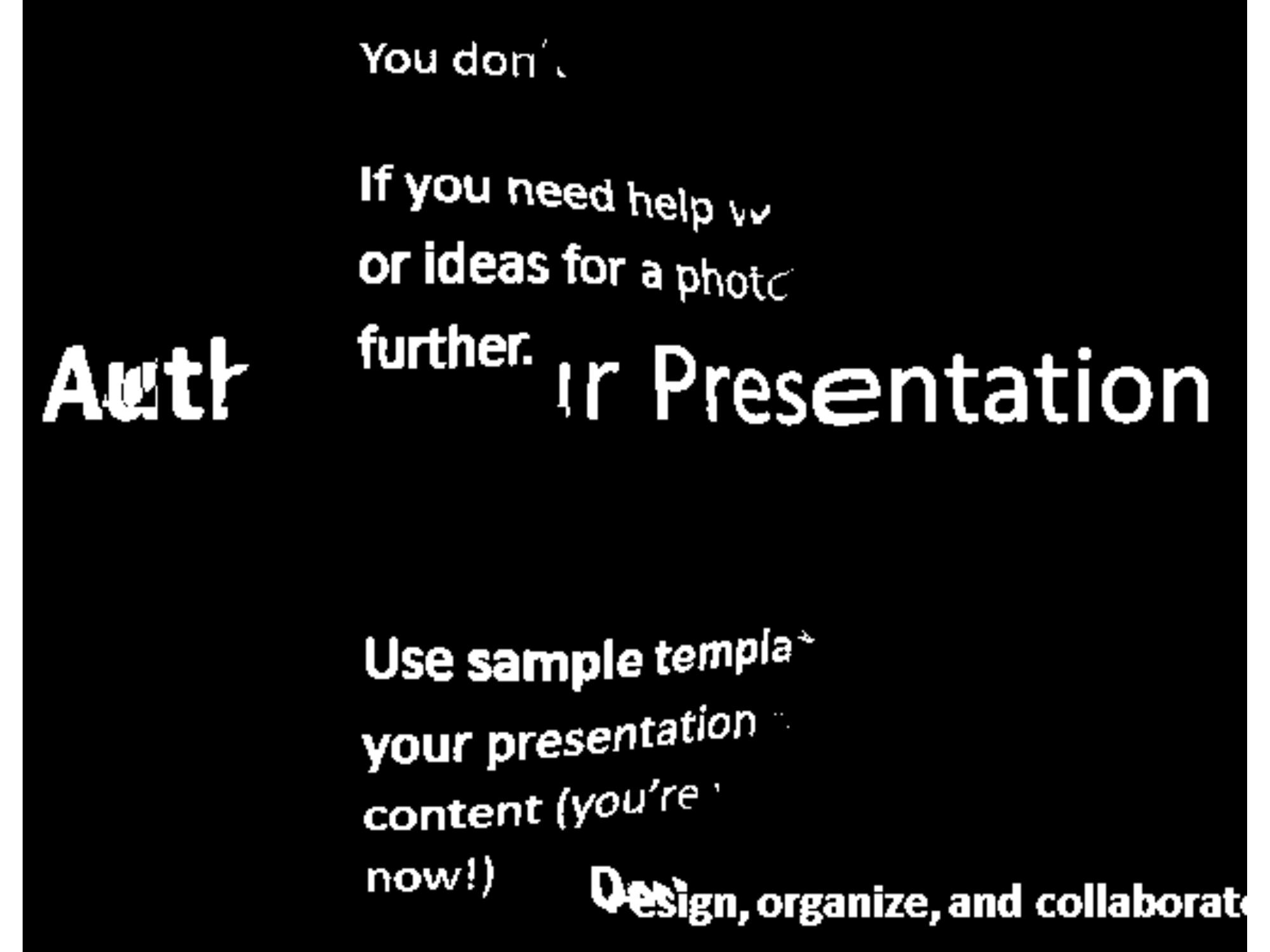}     
    \includegraphics [scale=0.22] {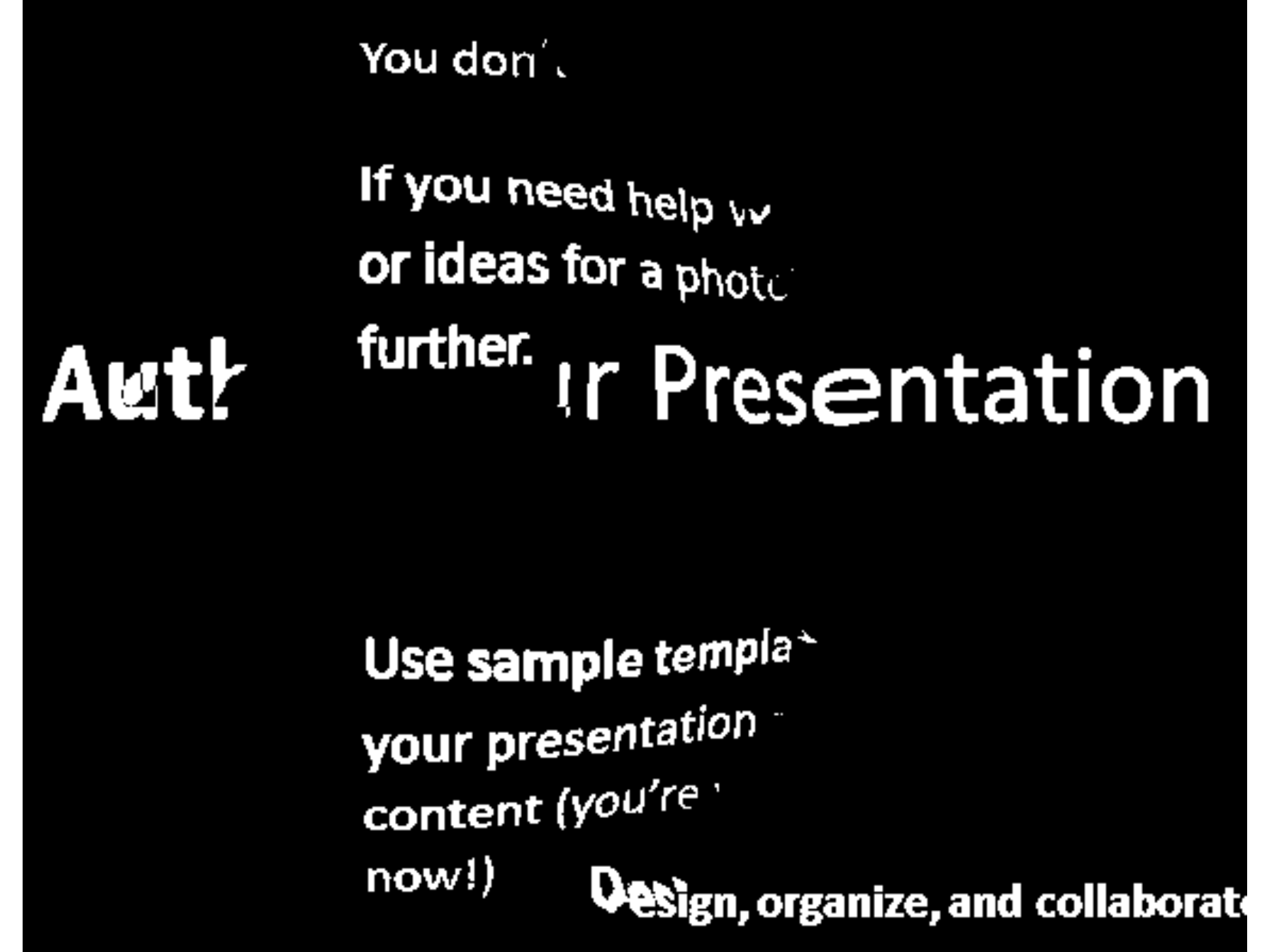}     
\end{center}
  \caption{Segmentation results of the RANSAC method using different number of basis functions. The foreground maps from left to right and top to bottom are obtained with 2, 5, 10, 15, and 20 basis functions, respectively.} \label{fig:fig2_6}
\end{figure}

To assess the effect of the number of basis, $K$, in the final segmentation result, we show the foreground map derived by several different number of basis functions using the RANSAC method in Figure~\ref{fig:fig2_6}.

\begin{figure}[h]
\begin{center}
    \includegraphics [scale=0.35] {2Original_Image-eps-converted-to.pdf}
\hspace{-0.18cm}	\includegraphics [scale=0.35] {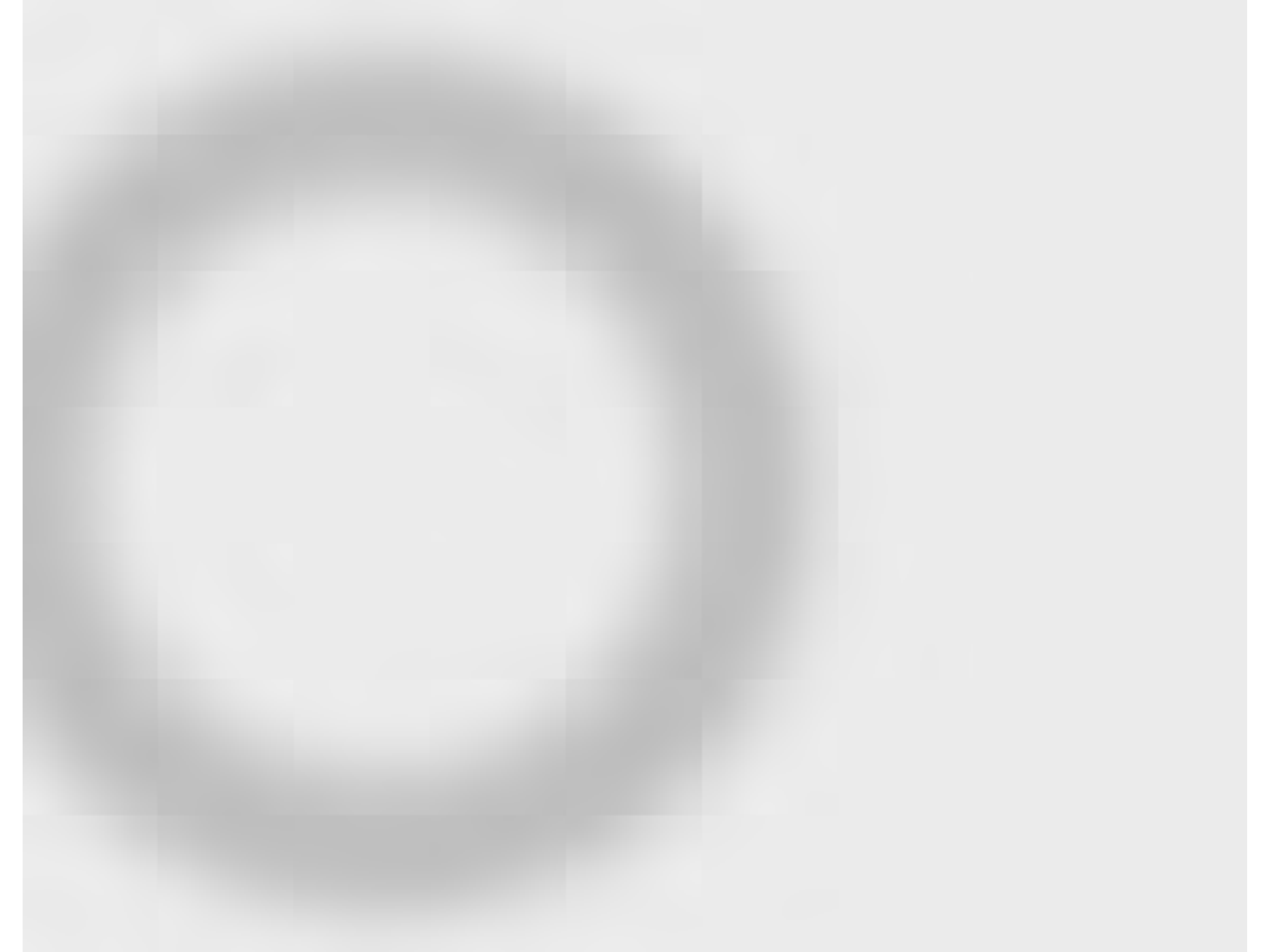}
\end{center}
  \caption{The reconstructed background of an image } \label{fig:fig2_7}
\end{figure}

To illustrate the smoothness of the background layer and its suitability for being coded with transform-based coding, the filled background layer of a sample image is presented in Figure~\ref{fig:fig2_7}. The background holes (those pixels that belong to foreground layers) are filled by the predicted value using the smooth model, which is obtained using the least squares fitting to the detected background pixels. As we can see the background layer is very smooth and does not have any sharp edges.

We compare the proposed algorithms with hierarchical k-means clustering used in DjVu, SPEC, least square fitting, and LAD algorithms.
For SPEC, we have adapted the color number threshold and the shape primitive size threshold from the default value given in \cite{spec}  when necessary to give more satisfactory result. Furthermore, for blocks classified as text/graphics based on the color number, we segment the most frequent color and any similar color to it (i.e. colors whose distance from most frequent color is less than 10 in luminance) in the current block as background and the rest as foreground.
We have also provided a comparison with least square fitting algorithm result, so that the reader can see the benefit of minimizing the $\ell_0$ and $\ell_1$ norm over minimizing the $\ell_2$ norm.

To provide a numerical comparison between the proposed scheme and previous approaches, we have calculated the average precision and recall and F1 score (also known as F-measure) \cite{metrics} achieved by different segmentation algorithms over this dataset. 
The precision and recall are defined as:
\begin{gather}
 Precision= \frac{\text{TP}}{\text{TP+FP}} \ , 
\ \ \ \ Recall= \frac{\text{TP}}{\text{TP+FN}} 	\ 	, 
\end{gather}
where $\text{TP}, \text{FP}$ and $\text{FN}$ denote true positive, false positive and false negative respectively. In our evaluation, we treat the foreground pixels as positive. A pixel that is correctly identified as foreground (compared to the manual segmentation) is considered true positive. The same holds for false negative and false positive. 
The balanced F1 score is defined as the harmonic mean of precision and recall, i.e. 
\begin{gather}
\text{F1}= 2 \ \frac{Precision \times Recall}{ Precision+Recall}  \label{eq:ch2_16}
\end{gather}
The average precision, recall and F1 scores by different algorithms are given in Table 2.3. 
As can be seen, the two proposed schemes achieve  much higher precision and recall than the DjVu and SPEC algorithms, and also provide noticeable gain over our prior LAD approach. Among the two proposed methods, sparse decomposition based algorithm achieved high precision, but lower recall than the RANSAC algorithm. 

\begin{table}[h]
\centering
  \caption{Segmentation accuracies of different algorithms}
  \centering
\begin{tabular}{|m{6.4cm}|m{2cm}|m{2cm}|m{2cm}|}
\hline
Segmentation Algorithm  &  \  Precision & \ \  Recall & \  F1 score\\
\hline
SPEC \cite{spec} & \ \ \ 50\% & \ \ \  64\% & \ \ \ 56\% \\
\hline
 DjVu \cite{djvu} & \ \ \ 64\% & \ \ \ 69\% & \ \ \ 66\% \\
\hline
 Least square fitting & \ \ \ 79\% & \ \ \ 60\% & \ \  \ 68\% \\ 
\hline
 Least Absolute Deviation \cite{lad} & \ \ \  90.5\% & \ \ \  87\% & \ \ \  88.7\% \\
\hline
 RANSAC based segmentation & \ \ \ 91\%  & \ \ \ 90\%  & \ \ \  90.4\%\\
 \hline
 Sparse Decomposition Algorithm & \ \ \ 94\%  & \ \ \ 87.2\%  & \ \ \  90.5\%\\
\hline
\end{tabular}
\label{TblComp}
\end{table}

\begin{figure}[h]
\begin{center}
    \includegraphics [scale=0.54] {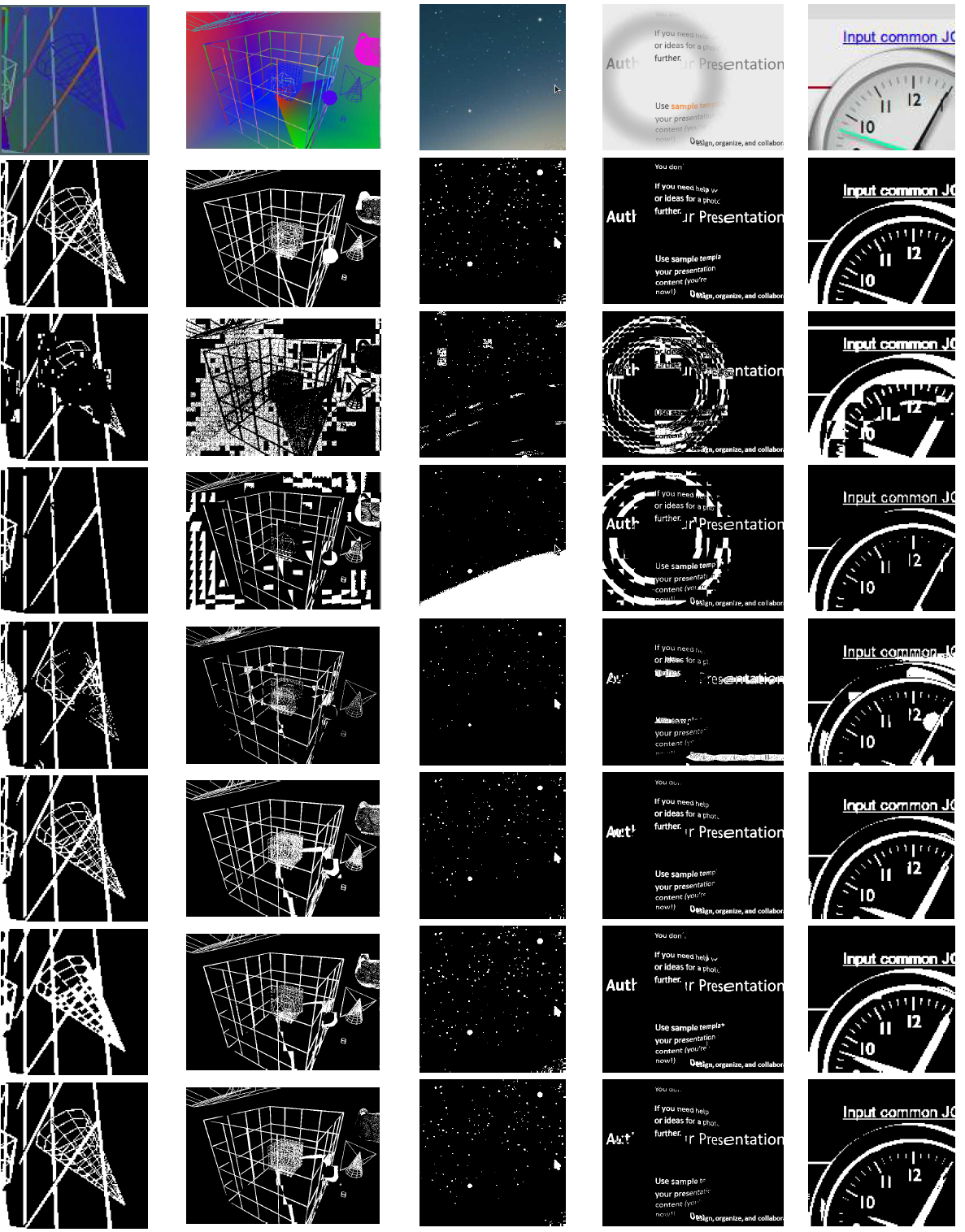} 
\end{center}
  \caption{Segmentation result for selected test images. The images in the first and second rows are the original and ground truth segmentation images. The images in the third, fourth, fifth and the sixth rows are the foreground maps obtained by shape primitive extraction and coding, hierarchical clustering in DjVu, least square fitting, and least absolute deviation fitting approaches. The images in the seventh and eighth rows include the results by the proposed RANSAC and sparse decomposition algorithms respectively.} \label{fig:fig2_8}
\end{figure}

The results for 5 test images (each consisting of multiple 64x64 blocks) are shown in Figure~\ref{fig:fig2_8}. Each test image is a small part of a frame from a HEVC SCC test sequence.

It can be seen that in all cases the proposed algorithms give superior performance over DjVu and SPEC, and slightly better than our prior LAD approach in some images. 
Note that our dataset mainly consists of challenging images where the background and foreground have overlapping color ranges. For simpler cases where the background has a narrow color range that is quite different from the foreground, both DjVu and the proposed methods will work well. On the other hand,  SPEC  does not work well when the background is fairly homogeneous within a block and the foreground text/lines have varying colors. 

It can be seen that in all cases the proposed algorithms give superior performance over DjVu and SPEC, and slightly better than our prior LAD approach in some images. 
Note that our dataset mainly consists of challenging images where the background and foreground have overlapping color ranges. For simpler cases where the background has a narrow color range that is quite different from the foreground, both DjVu and the proposed methods will work well. On the other hand,  SPEC  does not work well when the background is fairly homogeneous within a block and the foreground text/lines have varying colors. 

In terms of complexity, it took 20, 506 and 962 ms on average for a block of $64 \times 64$ to be segmented using RANSAC, LAD and sparse decomposition based segmentation algorithms (with the pre-processing steps) using MATLAB 2015 on a laptop with Windows 10 and Core i5 CPU running at 2.6GHz.

\section{Conclusion}
In this chapter, we proposed two novel segmentation algorithms for separating the foreground text and graphics from smooth background. 
The background is defined as the smooth component of the image that can be well modeled by a set of low frequency DCT basis functions and the foreground refers to those pixels that cannot be modeled with this smooth representation. 
One of the proposed algorithms uses robust regression technique to fit a smooth function to an image block and detect the outliers. The outliers are considered as the foreground pixels. Here RANSAC algorithm is used to solve this problem.
The second algorithm uses sparse decomposition techniques to separate the smooth background from the sparse foreground layer.
Total variation of the foreground component is also added to the cost function to enforce the foreground pixels to be connected.
Instead of applying the proposed algorithms to every block, which are computationally demanding, we first check whether the block satisfies several conditions and can be segmented using simple methods. We further propose to apply the algorithm recursively using quad-tree decomposition, starting with larger block sizes. A block is split only if RANSAC or sparse decomposition cannot find sufficient inliers in this block. 
These algorithms are tested on several test images and compared with three other well-known algorithms for background/foreground separation and the proposed algorithms show significantly better performance for blocks where the background and foreground pixels have overlapping intensities.

\chapter{Robust Subspace Learning}
Subspace learning is an important problem, which has many applications in image and video processing.
It can be used to find a low-dimensional representation of signals and images. 
But in many applications, the desired signal is heavily distorted by outliers and noise, which negatively affect the learned subspace.
In this work, we present a novel algorithm for learning a subspace for signal representation, in the presence of structured outliers and noise.
The proposed algorithm tries to jointly detect the outliers and learn the subspace for images.
We present an alternating optimization algorithm for solving this problem, which iterates between learning the subspace and finding the outliers.
We also show the applications of this algorithm in image foreground  segmentation. 
It is shown that by learning the subspace representation for background, better performance can be achieved compared to the case where a pre-designed subspace is used.

In Section 3.1. we talk about some background on subspace learning . 
We then present the proposed method in Section 3.2.
Section 3.3 provides the experimental results for the proposed algorithm, and its application for image segmentation.

\section{Background and Relevant Works}
Many of the signal and image processing problems can be posed as the problem of learning a low dimensional linear or multi-linear model.
Algorithms for learning linear models can be seen as a special case of subspace fitting. Many of these algorithms are based on least squares estimation techniques, such as principal component analysis (PCA) \cite{pca}, and linear discriminant analysis (LDA) \cite{lda}.
But in general, training data may contain undesirable artifacts due to occlusion, illumination changes, overlaying component (such as foreground texts and graphics on top of smooth background image). These artifacts can be seen as outliers for the desired signal.
As it is known from statistical analysis, algorithms based on least square fitting fail to find the underlying representation of the signal in the presence of outliers \cite{lsf}.
Different algorithms have been proposed for robust subspace learning to handle outliers in the past, such as the work by Torre \cite{rsl}, where he suggested an algorithm based on robust M-estimator for subspace learning.
Robust principal component analysis \cite{rpca} is another approach to handle the outliers.
In \cite{lerman}, Lerman et al proposed an approach for robust linear model fitting by parameterizing linear subspace using orthogonal projectors.
There have also been many works for online subspace learning/tracking for video background subtraction, such as GRASTA \cite{grasta}, which uses a robust $\ell_1$-norm cost function in order to estimate and track
non-stationary subspaces when the streaming data vectors are corrupted with outliers, and t-GRASTA \cite{tgrasta}, which simultaneously estimate a decomposition of a collection of images into a low-rank subspace, and sparse part, and a transformation such as rotation or translation of the image.

In this work, we present an algorithm for subspace learning from a set of images, in the presence of structured outliers and noise.
We assume sparsity and connectivity priors on outliers that suits many of the image processing applications.
As a simple example we can think of smooth images overlaid with texts and graphics foreground, or face images with occlusion (as outliers).
To promote the connectivity of the outlier component, the group-sparsity of outlier pixels is added to the cost function.
We also impose the smoothness prior on the learned subspace representation, by penalizing the gradient of the representation.
We then propose an algorithm based on the sparse decomposition framework for subspace learning. This algorithm jointly detects the outlier pixels and learn the low-dimensional subspace for underlying image representation. 

We then present its application for background-foreground segmentation in still images, and show that it achieves better performance than previous algorithms.
We compare our algorithm with some of the prior approaches, including sparse and low-rank decomposition, and group sparsity based segmentation using DCT bases.
The proposed algorithm has applications in text extraction, medical image analysis, and image decomposition \cite{subapp1}-\cite{subapp3}.

\begin{figure}[h]
\begin{center}
    \includegraphics [scale=0.8] {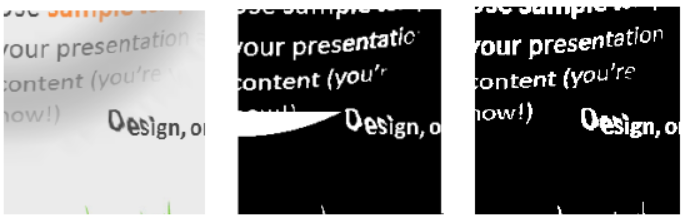}
\end{center}
  \caption{The left, middle and right images denote the original image, segmented foreground by hierarchical k-means and the proposed algorithm respectively.}
\label{fig:subspace1}
\end{figure}

Figure~\ref{fig:subspace1} shows a comparison between the foreground mask derived from the proposed segmentation algorithm and hierarchical clustering for a sample image.
\vspace{-0.1cm}

\section{Problem Formulation}
Despite the high-dimensionality of images (and other kind of signals), many of them have a low-dimensional representation.
For some category of images, this representation may be a very complex manifold which is not simple to find, but for many of the smooth images this low-dimensional representation can be assumed to be a subspace. 
Therefore each signal $x \in R^N$ can be efficiently represented:
\begin{equation}
\begin{aligned}
x\simeq P\alpha \label{eq:sub1}
\end{aligned}
\end{equation}
where  $P \in R^{N \times k}$ and $k \ll N$, and $\alpha$ denotes the representation coefficient in the subspace.
\\There have been many approaches in the past to learn $P$ efficiently, such as PCA and robust-PCA.
But in many scenarios, the desired signal can be heavily distorted with outliers and noise, and those distorted pixels should not be taken into account in subspace learning process, since they are assumed to not lie on the desired signal subspace.
Therefore a more realistic model for the distorted signals should be as:
\begin{equation}
\begin{aligned}
x= P \alpha+s+\epsilon  \label{eq:sub2}
\end{aligned}
\end{equation}
where $s$ and $\epsilon$ denote the outlier and noise components respectively.
Here we propose an algorithm to learn a subspace, $P$, from a training set of $N_d$ samples $x_i$, by minimizing the noise energy ($\|\epsilon_i  \|_2^2= \|x_i-P\alpha_i-s_i  \|_2^2$), and some regualrization term on each component, as shown in Eq. \eqref{eq:sub3}:
\begin{equation}
\begin{aligned}
& \underset{P, \alpha_i, s_i}{\text{min}}
 \sum_{i=1}^{N_d} \ \frac{1}{2} \| x_i-P\alpha_i-s_i  \|_2^2+ \lambda_1  \phi(P\alpha_i) + \lambda_2 \psi(s_i)  \\
& \ \text{s.t.}
\ \ \ \ \ \ \ \ P^tP= I, \ s_i \geq 0 
\end{aligned} \label{eq:sub3}
\end{equation}
where $\phi(.)$ and $\psi(.)$ denote suitable regularization terms on the first and second components, promoting our prior knowledge about them.
Here we assume the underlying image component is smooth, therefore it should have a small gradient. And for the outlier, we assume it is sparse and also connected.
Hence $\phi(P\alpha_i)= \| \nabla P \alpha_i \|_2^2$, and $\psi(s)= \|s\|_1+ \beta \sum_m \|s_{g_m}\|_2$, where $g_m$ shows the m-th group in the outlier (the pixels within each group are supposed to be connected).
\\Putting all these together, we will get the following optimization problem:
\begingroup\makeatletter\def\f@size{12}\check@mathfonts
\begin{equation}
\begin{aligned}
& \hspace{-1cm } \underset{P, \alpha_i, s_i}{\text{min}}
  \sum_{i=1}^{N_d} \frac{1}{2} \| x_i-P\alpha_i-s_i  \|_2^2+ \lambda_1  \| \nabla P \alpha_i \|_2^2 + \lambda_2 \|s_i\|_1+ \lambda_3 \sum_m \|s_{i,g_m}\|_2  \\
& \ \text{s.t.}
\ \ \ \ \ \ \ \ P^tP= I, \ s_i \geq 0 \label{eq:sub4}
\end{aligned}
\end{equation}
\endgroup
Here by $s_i \geq 0$ we mean all elements of the vector $s_i$ should be non-negative.
Note that $ \| \nabla P \alpha_i \|_2^2$ denotes the spatial gradient, which can be written as:
\begin{equation}
\begin{aligned}
\| \nabla P \alpha_i \|_2^2= \| D_x P \alpha_i \|_2^2+ \| D_y P \alpha_i \|_2^2= \| D P \alpha_i \|_2^2
\end{aligned} \label{eq:sub5}
\end{equation}
where $D_x$ and $D_y$ denote the horizontal and vertical derivative matrix operators, and $D=[D_x^t, D_y^t]^t$.

The optimization problem in Eq.  \eqref{eq:sub4} can be solved using alternating optimization over $\alpha_i$, $s_i$ and $P$.
In the following part, we present the update rule for each variable by setting the gradient of cost function w.r.t that variable to zero.
\\The update step for $\alpha_i$ would be:
\begin{equation}
\begin{aligned}
\alpha_i^*= & \underset{ \alpha_i}{\text{\ \ argmin}}
 \{ \frac{1}{2} \| x_i- P\alpha_i- s_i  \|_2^2+ \frac{\lambda_1}{2} \| D P \alpha_i \|_2^2 = F_{\alpha}(\alpha_i) \} \Rightarrow \\
& \hspace{-0.85cm}  \nabla_{\alpha_i}F_{\alpha}(\alpha_i^*)=0 \Rightarrow P^t(P\alpha_i^*+ s_i- x_i)+ \lambda_1 P^t D^t D P \alpha_i^*=0 \Rightarrow \\
& \hspace{-0.8cm} \alpha_i^*= (P^t P+ \lambda_1 P^t D^t D P )^{-1} P^t (x_i-s_i) \\
\end{aligned} \label{eq:sub6}
\end{equation}

The update step for the $m$-th group of the variable $s_i$ is as follows:
\begin{equation}
\begin{aligned}
& \  s_{i,g_m}= \underset{ s_i}{\text{\ \ argmin}}
 \{ \frac{1}{2} \| (x_i- P\alpha_i)_{g_m}- s_{i,g_m}  \|_2^2+  \lambda_2 \| s_{i,g_m} \|_1+ \\ 
& \ \lambda_3 \|s_{i,g_m}\|_2 = F_s(s_{i,g_m})\} \ \ \ \ \ \ \ \ \ \ \ \text{s.t.} \ \ \ \ \ \ s_{i,g_m} \geq 0 \\ 
&  \Rightarrow \nabla_{s_{i,g_m}}F_{s}(s_{i,g_m})=0 \Rightarrow s_{i,g_m}+(P\alpha_i-x_i)_{g_m}+ \lambda_2 \text{sign}(s_{i,g_m}) \\ 
& +\lambda_3 \frac{s_{i,g_m}}{\ \|s_{i,g_m}\|_2}= 0 \Rightarrow s_{i,g_m}+ \lambda_3 \frac{s_{i,g_m}}{\ \|s_{i,g_m}\|_2}= (x_i-P\alpha_i)_{g_m}-\lambda_2 1\\
& \ \hspace{-0.1cm} \Rightarrow s_{i,g_m}= \text{block-soft} ((x_i-P\alpha_i)_{g_m}-\lambda_2 1, \ \lambda_3)  \\
\end{aligned} \label{eq:sub7}
\end{equation}
Note that, because of the constraint $s_{i,g_m} \geq 0$, we can approximate $\text{sign}(s_{i,g_m})=1$, and then project the $s_{i,g_m}$ from soft-thresholding result onto $s_{i,g_m} \geq 0$, by setting its negative elements to 0.   
The block-soft(.) \cite{groupshervin} is defined as Eq. \eqref{eq:sub8}:
\begin{equation}
\begin{aligned}
\text{block-soft}(y,t)= \text{max}(1-\frac{t}{\ \|y\|_2},0) \ y
\end{aligned} \label{eq:sub8}
\end{equation}
\\For the subspace update, we first ignore the orthonormality constraint ($P^tP=I$), and update the subspace column by column, and then use Gram-Schmidt algorithm \cite{gram} to orthonormalize the columns. If we denote the j-th column of $P$ by $p_j$, its update can be derived as:
\begin{equation}
\begin{aligned}
& \ P= \underset{ P}{\text{\ \ argmin}}
 \{\sum_i \frac{1}{2} \| x_i- P\alpha_i- s_i  \|_2^2+  \lambda_1 \| D P \alpha_i \|_2^2 \} \Rightarrow \\ 
& \ p_j= \underset{ p_j}{\text{\ \ argmin}}
 \{\sum_i \frac{1}{2} \| (x_i- \sum_{k \ne j}p_k\alpha_i(k)- s_i) - p_j \alpha_i(j) \|_2^2+ \\
 & \ \lambda_1 \| D \sum_{k \neq j} p_k \alpha_i(k)+ D p_j \alpha_i(j) \|_2^2 = \sum_i  \frac{1}{2} \| \eta_{i,j} - p_j \alpha_i(j) \|_2^2+ \\
 &  \ \lambda_1 \| \gamma_{i,j}+ D p_j \alpha_i(j) \|_2^2 = F_{p}(p_j) \} \Rightarrow \nabla_{p_j}F_{p}(p_j^*)=0 \Rightarrow \\
 &  \ \sum_i \alpha_i(j) \big(\alpha_i(j)p_j-\eta_{i,j}\big)+ \lambda_1 \alpha_i(j) D^t \big( \alpha_i(j) Dp_j + \gamma_{i,j}  \big)=0 \Rightarrow \\
  & \  \big( \sum_i \alpha_i^2(j) \big) (I+ \lambda_1 D^tD) p_j= \sum_i \big( \alpha_i(j) \eta_{i,j}-\lambda_1 \alpha_i(j)D^t \gamma_{i,j} \big)= \beta_j \\
  & \ \Rightarrow p_j= (I+ \lambda_1 D^tD)^{-1} \beta_j /\big( \sum_i \alpha_i^2(j) \big) \\
\end{aligned} \label{eq:sub9}
\end{equation}
where $\eta_{i,j}=x_i-s_i- \sum_{k \ne j}p_k\alpha_i(k)$, and $\gamma_{i,j}=D \sum_{k \neq j} p_k \alpha_i(k)$.
After updating all columns of $P$, we apply Gram-Schmidt algorithm to project the learnt subspace onto $P^tP=I$. Note that orthonormalization should be done at each step of alternating optimization.
It is worth to mention that for some applications the non-negativity assumption for the structured outlier may not be valid, so in those cases we will not have the $s_i \geq 0$ constraint. In that case, the problem can be solved in a similar manner, but we need to introduce an auxiliary variable $s=z$, to be able to get a simple update for each variable.

looking at \eqref{eq:sub2} it seems very similar to our image segmentation framework based on sparse decomposition, which is discussed in previous chapter. 
Therefore we study the application of subspace learning for the foreground separation problem, and show that it can bring further gain in segmentation results by learning a suitable subspace for background modeling.
Here we consider foreground segmentation based on the  optimization problem in Eq.  \eqref{eq:subseg}.
Notice that we use group-sparsity to promote foreground connectivity in this formulation, to have a consistent optimization framework with our proposed subspace learning algorithm.
\begin{equation}
\begin{aligned}
& \underset{\alpha, s}{\text{min}}
  \frac{1}{2} \| x-P\alpha-s  \|_2^2+ \lambda_1  \| D P \alpha \|_2^2 + \lambda_2 \|s\|_1+ \lambda_3 \sum_m \|s_{g_m}\|_2    \\
& \ \text{s.t.}
\ \ \ \ \ \ \ \ s \geq 0 \label{eq:subseg}
\end{aligned}
\end{equation}

\section{Experimental Results}
To evaluate the performance of our algorithm, we trained the proposed framework on image patches extracted from some of the images of the screen content image segmentation dataset provided in \cite{lad}. 
Before showing the results, we report the weight parameters in our optimization.
We used $\lambda_1=0.5$, $\lambda_2=1$ and $\lambda_3=2$, which are tuned by testing on a validation set.
We provide the results for subspace learning and image segmentation in the following sections.

\subsection{The Learned Subspace}
\begin{figure}[H]
\begin{center}
    \includegraphics [scale=0.50] {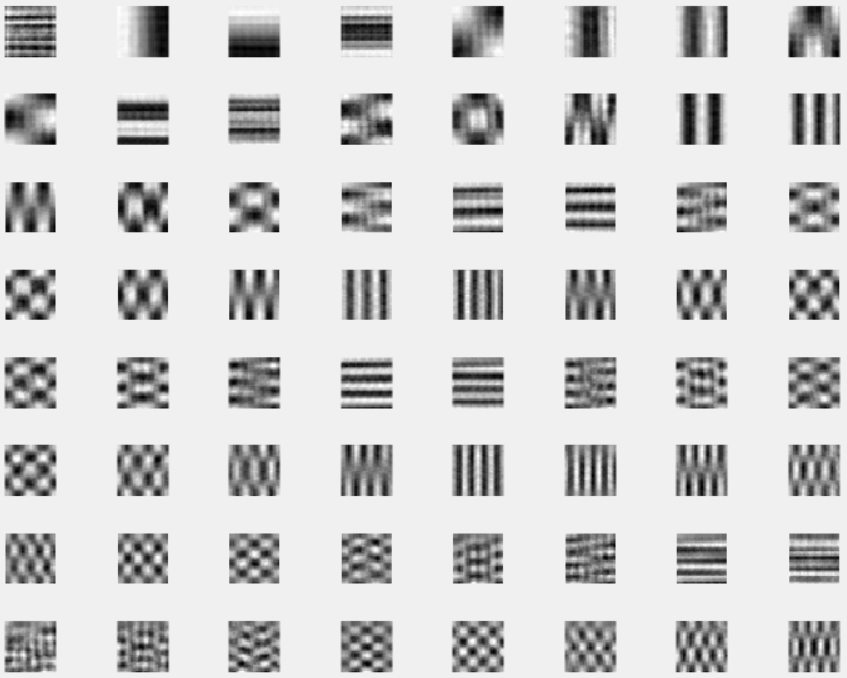}
\end{center}
\begin{center}
    \includegraphics [scale=0.60] {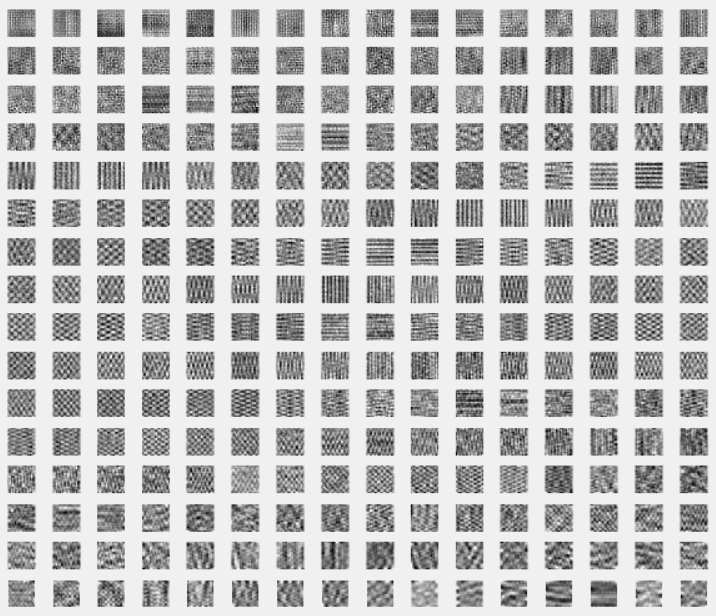}
  \caption{The learned basis images for the 64, and 256 dimensional subspace, for 32x32 image blocks} \label{fig:subspaces}
\end{center}
\end{figure}

We extracted around 8,000 overlapping patches of size 32x32, with stride of 5 from a subset of these images and used them for learning the subspace, and learned two subspaces, one 64 dimensional subspace (which means 64 basis images of size 32x32), and the other one 256 dimensional.
The learned atoms of each of these subspaces are shown in Figure~\ref{fig:subspaces}. 
As we can see the learned atoms contain different edge and texture patterns, which is reasonable for image representation. The right value of subspace dimension highly depends to the application. 
For image segmentation problem studied in this paper, we found that using only first 20 atoms performs well on image patches of 32x32.

\subsection{Applications in Image Segmentation}
After learning the subspace, we use this representation for foreground segmentation in still images, as explained in Section 3.2.
The segmentation results in this section are derived by using a 20 dimensional subspace for background modeling.
We use the same model as the one in Eq.  \eqref{eq:subseg} for decomposition of an image into background and foreground, and $\lambda_i$'s are set to the same value as mentioned before.
We then evaluate the performance of this algorithm on the remaining images from screen content image segmentation dataset \cite{our_dataset}, and some other images that includes text over textures, and compare the results  with two other algorithms;  sparse and low-rank decomposition \cite{lowrank}, and group-sparsity based segmentation using DCT basis \cite{groupshervin}.
For sparse and low rank decomposition, we apply the fast-RPCA algorithm \cite{lowrank} on the image blocks, and threshold the sparse component to find the foreground location.
For low-rank decomposition, we have used the MATLAB implementation provided by Stephen Becker at \cite{becker}.

To provide a numerical comparison, we report the average precision, recall and F1 score achieved by different algorithms over this dataset, in Table 3.1. 
\begin{table}[H]
\centering
  \caption{Comparison of accuracy of different algorithms}
  \centering
\begin{tabular}{|m{6cm}|m{2cm}|m{2cm}|m{2cm}|}
\hline
Segmentation Algorithm  &  Precision & \ \  Recall & \  F1 score\\
\hline 
 Low-rank Decomposition  \cite{lowrank} & \ \ \  78\% & \ \ \  86.5\% & \ \ \  82.1\% \\
\hline
 Group-sparsity with DCT Bases & \ \ \  92.2\% & \ \ \  86\% & \ \ \  89\% \\
\hline
 The proposed algorithm & \ \ \ 93\%  & \ \ \ 86\%  & \ \ \  89.3\%\\
\hline
\end{tabular}
\label{TblComp}
\end{table}
As it can be seen, the proposed scheme achieves better overall performance than low-rank decomposition. Compared to group-sparsity using DCT Bases, the proposed formulation has slightly better performance.

\begin{figure}[ht]
\begin{center}
    \includegraphics [scale=0.7] {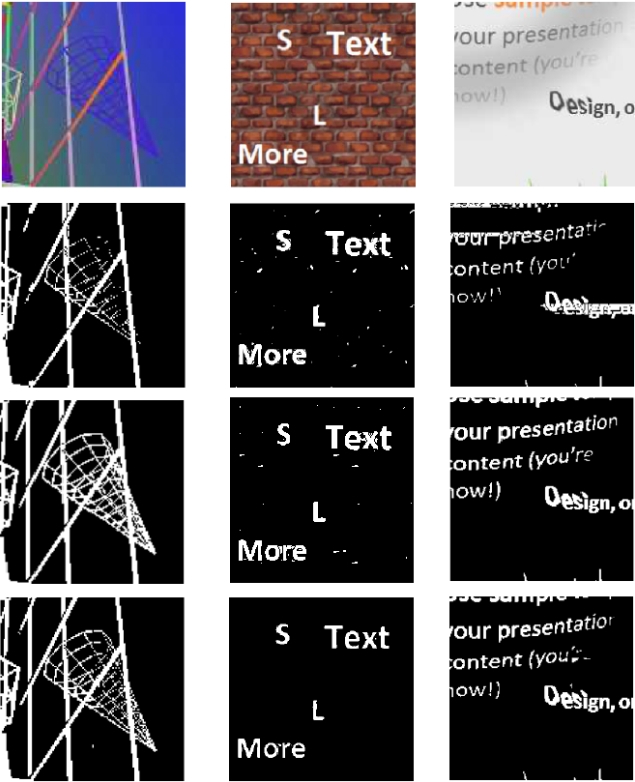}
\label{fig:subspace-seg}
\end{center}
\caption{Segmentation result for the selected test images. The images in the first to fourth rows denote the original image, and the foreground map by sparse and low-rank decomposition, group-sparisty with DCT bases, and the proposed algorithm respectively.} \label{fig:fig3_3}
\end{figure}

To see the visual quality of the segmentation, the results for 3 test images (each consisting of multiple 64$\times$64 blocks) are shown in Figure~\ref{fig:fig3_3}. 


It can be seen that, there are noticeable improvement for the segmentation results over low-rank decomposition.
For background with very smooth patterns, low frequency DCT bases are a effective subspace, and therefore, we do not see improvement with the learnt bases. 
But for more complex background such as that in the middle column, the learnt subspace can provide improvement.  
In general, the improvement depends on whether the background can be represented well by the chosen fixed bases or not. 
Also,  if we do the  subspace learning directly on the image you are trying to segment, (that is, we do subspace learning and segmentation jointly), we may be able to gain even more.

\section{Conclusion}
In this chapter, we proposed a subspace learning algorithm for a set of smooth signals in the presence of structured outliers and noise.
The outliers are assumed to be sparse and connected, and suitable regularization terms are added to the optimization framework to promote this properties.
We then solve the optimization problem by alternatively updating the model parameters, and the subspace.
We also show the application of this framework for background-foreground segmentation in still images, where the foreground can be thought as the outliers in our model, and achieve better results than the previous algorithms for background/foreground separation.

\chapter{Masked Signal Decomposition}
Signal decomposition is a classical problem in signal processing, which aims to separate an observed signal into two or more components each with its own property. 
Usually each component is described by its own subspace or dictionary.
Extensive research has been done for the case where the components are  additive, but in real world applications, the components are often non-additive. For example, an image may consist of a foreground object overlaid on a background, where each pixel either belongs to the foreground or the background. 
In such a situation, to separate signal components, we need to find a binary mask which shows the location of each component.
Therefore it requires to solve a binary optimization problem.
Since most of the binary optimization problems are intractable, we relax this problem to the approximated continuous problem, and solve it by alternating optimization technique.
We show the application of the proposed algorithm for three applications:  separation of text  from background in images, separation of moving objects from a background undergoing global camera motion in videos, separation of sinusoidal and spike components in one dimensional signals. We demonstrate in each case that considering the non-additive nature of the problem can lead to significant improvement \cite{mymask}.

In the remaining parts, we first go over some of the relevant works in Section 4.1.
We then presents the problem formulation in Section 4.2.
Section 4.3 shows the application of this approach for motion segmentation.
The experimental results, and the applications are provided in Section 4.4 and the chapter is concluded in Section 4.5.

\section{Background and Relevant Works}
Signal decomposition is an important problem in signal processing and has a wide range of applications.
Image segmentation, sparse and low-rank decomposition, and audio source separation \cite{audio1}-\cite{audio3} are some of the applications of signal decomposition.
Perhaps, Fourier transform \cite{fourier} is one of the earliest work on signal decomposition where the goal is to decompose a signal into different frequencies.
Wavelet and multi-resolution decomposition are also another big group of methods which are designed for signal decomposition in both time and frequency domain \cite{wave1}-\cite{wave3}. 
In the more recent works, there have been many works on sparsity based signal decomposition. 
In \cite{bofili}, the authors proposed a sparse representation based method for blind source sparation.
In \cite{starck}, Starck et al proposed an image decomposition approach using both sparsity and variational approach.
The same approach has been used for morphological component analysis by the Elad et al \cite{elad}.
In the more recent works, there have been many works on low-rank decomposition, where in the simplest case the goal is to decompose a signal into two components, one being low rank, another being sparse.
Usually the nuclear and $\ell_1$ norms \cite{fazel} are used to promote low-rankness and sparsity respectively. 
To name some of the promising works along this direction,
in \cite{rpca}, Candes et al proposed a low-rank decomposition for matrix completion.
In \cite{peng}, Peng et al proposed a sparse and low-rank decomposition approach with application for robust image alignment.
A similar approach has been proposed for transform invariant low-rank textures \cite{tilt}.
This approach has also been used for background subtraction in videos \cite{back_sub}.

In this work, we try to extend the signal decomposition problem for all the above applications, to the overlaid model, which is instead of multiple components contributing to a signal element at some point, one and only one of them are contributing at each element.

\section{The Proposed Framework}
Most of the prior approaches for signal decomposition consider additive model, i.e. the signal components are added in a mathematical sense to generate the overall signal. 
In the case of two components, this can be described by:
\begin{equation}
x= x_1+ x_2 \label{eq:ch4_1}
\end{equation}
Here $x$ denotes a vector in $R^N$. Assuming we have some prior knowledge about each component, we can form an optimization problem as in Eq. \eqref{eq:ch4_2} to solve the signal decomposition problem. Here $\phi_k(\cdot)$ is the regularization term that encodes the prior knowledge about the corresponding signal component.
\begin{equation}
\begin{aligned}
& \hspace{0.1cm} \underset{w_k, x_k}{\text{min}}
 \   \sum_{k=1}^2 \phi_k(x_k), \ \ \  \ \text{s.t.} \ \ \sum_{k=1}^2 x_k= x 
\end{aligned} \label{eq:ch4_2}
\end{equation}
In this work, we investigate a different class of signal decomposition, where the signal components are overlaid on top of each other, rather than simply added. 
In other word, at each signal element only one of the signal components contributes to the observed signal x.
We can formulate this as:
\begin{equation}
x= \sum_{k=1}^2 w_k \circ x_k   \ \ \ \ \text{s.t.}  \ \ w_k \in \{0,1\}^N, \  \sum_{k=1}^2 w_k= \textbf{1} \label{eq:ch4_3}
\end{equation}
where $\circ$ denotes the element-wise product \cite{hadamard}, and $w_k$'s are the binary masks, where at each element one and only one of the $w_k$'s is 1, and the rest are zero. 
The constraint $\sum_{k=1}^2 w_k= \textbf{1}$ results in $w_1= \textbf{1}- w_2$.
In our work, we assume that each component can be represented with a known subspace/dictionary.

If these components have different characteristics, it would be possible to separate them to some extent. 
One possible way is to form an optimization problem as below:
\begin{equation}
\begin{aligned}
& \hspace{2cm} \underset{w_k, x_k}{\text{min}}
 \ \phi_1(x_1, w_1)+\phi_2(x_2, w_2) \\
& \ \text{s.t.}
\ \ w_k \in \{0,1\}^n, \ w_2= \textbf{1}-w_1 ,\  \sum_{k=1}^2 w_k \circ x_k= x 
\end{aligned} \label{eq:ch4_4}
\end{equation}
where $\phi_k$ encodes our prior knowledge about each component and its corresponding mask.

One prior knowledge that we assume is that each component $x_k$ can be well represented using some proper dictionary/subsbase $P_k$,  and get:
\begin{equation}
x= (\textbf{1}-w) \circ {(P_1\alpha_1)}+ w \circ  (P_2\alpha_2) \label{eq:ch4_5}
\end{equation}
where $P_k$ is $N\times M_k$ matrix, where each column denotes one of the basis functions from the corresponding subspace/dictionary, and $M_k$ is the number of basis functions for component $k$.

Note that, in an alternative notation, Eq. \eqref{eq:ch4_5} can be written as:
\begin{equation}
x= (I-W) {P_1\alpha_1}+ W P_2\alpha_2 \label{eq:ch4_6}
\end{equation}
where $W= \text{diag}(w)$ is a diagonal matrix with the vector $w$ on its main diagonal.
If a diagonal element is 1, the corresponding element belongs to component 2, otherwise to component 1.

The decomposition problem in Eq. \eqref{eq:ch4_6} is a highly ill-posed problem. 
Therefore we need to impose some prior on each component, and also on $w$ to be able to perform this decomposition.
We assume that each component has a sparse representation with respect to its own subspace, but not with respect to the other one.
We also assume that the second component is sparse and connected. 
This would be the case, for example,  if the second component corresponds to text overlaid over a background image; or a moving object over a stationary background in a video.

To promote sparsity of the second component, we add the $\ell_0$ norm of $w$ to the cost function (note that $w$ corresponds to the support of the second component).
To promote connectivity, we can either add the group sparsity or total variation of $w$ to the cost function.
Here we use total variation.
The main reason is that for group sparsity, it is not very clear what is the best way to define groups, as the foreground pixels could be connected in any arbitrary direction, whereas total variation can deal with this arbitrary connectivity more easily.

We can incorporate all these priors in an optimization problem as shown below:
\begin{equation}
\begin{aligned}
& \hspace{-0.2cm}\underset{w, \alpha_1, \alpha_2}{\text{min}}
 \  \frac{1}{2} \| x- (\textbf{1}-w) \circ P_1\alpha_\textbf{1}-w \circ P_2\alpha_2  \|_2^2+ \lambda_1 \| w \|_0+ \lambda_2 \text{TV}(w)   \\
& \ \text{s.t.}
\ \ \ \ \ \ w \in \{0,1\}^N  , \ \| \alpha_1 \|_0 \leq K_1 , \ \| \alpha_2 \|_0 \leq K_2
\end{aligned} \label{eq:ch4_7}
\end{equation}

Total variation of an image can be defined as Eq. \eqref{eq:ch4_8}:
\begin{equation}
\begin{aligned}
TV(w)= \| D_xw \|_1+\| D_yw \|_1= \|Dw\|_1
\end{aligned} \label{eq:ch4_8}
\end{equation}
where $D_x$ and $D_y$ are the horizontal and vertical difference operator matrices, and $D=[ D_x',D_y']'$.

The problem in Eq. \eqref{eq:ch4_7} involves multiple variables, and is not tractable, both because of the  $\| w \|_0$ term in the cost function and also the binary nature of $w $. We relax these conditions to be able to solve this problem in an alternating optimization approach. 
We replace the $\| w \|_0$ in the cost function with $\| w \|_1$, and also relax the $w \in \{0,1\}^N$ condition to $w \in [0,1]^N$ (which is known as linear relaxation in the mixed integer programming).
Then we will get the following optimization problem:
\begin{equation}
\begin{aligned}
& \hspace{-0.2cm}\underset{w, \alpha_1, \alpha_2}{\text{min}}
 \  \ \frac{1}{2} \| x- (\textbf{1}-w)\circ  P_1\alpha_\textbf{1}-w \circ  P_2\alpha_2  \|_2^2+ \lambda_1 \| w \|_1+ \lambda_2 \| Dw \|_1   \\
& \ \text{s.t.}
\ \ \ \ \ \ w \in [0,1]^N  , \ \| \alpha_1 \|_0 \leq K_1 , \ \| \alpha_2 \|_0 \leq K_2
\end{aligned} \label{eq:ch4_9}
\end{equation}

This problem can be solved with different approaches, such as majorization minimization, alternating direction method, and random sampling approach.

To solve the optimization problem in Eq. \eqref{eq:ch4_9} with augmented Lagrangian algorithm, we first introduce two auxiliary random variables as shown in Eq. \eqref{eq:ch4_10}:
\begin{equation}
\begin{aligned}
& \underset{w, \alpha_1, \alpha_2}{\text{min}}
 \  \ \frac{1}{2} \| x- (\textbf{1}-w)\circ  P_1\alpha_\textbf{1}-w \circ  P_2\alpha_2  \|_2^2+ \lambda_1 \| y \|_1+ \lambda_2 \| z \|_1   \\
& \ \text{s.t.}
\ \ \ \ \ \ w \in [0,1]^N, \ y= w,\ z= Dw, \| \alpha_1 \|_0 \leq K_1 ,  \| \alpha_2 \|_0 \leq K_2
\end{aligned} \label{eq:ch4_10}
\end{equation}

We then form the augmented Lagrangian as below:
\begin{equation}
\begin{aligned}
L(\alpha_1, \alpha_2, w, y, z, u_1, u_2)=  \frac{1}{2} \| x- (\textbf{1}-w)\circ  P_1\alpha_\textbf{1}-w \circ  P_2\alpha_2  \|_2^2+ \lambda_1 \| y \|_1+ \lambda_2 \| z \|_1  \\ + u_1^t (w-y)+ u_2^t (Dw-z)+ \frac{\rho_1}{2} \|w-y\|_2^2+  \frac{\rho_2}{2} \|Dw-z\|_2^2 \hspace{4.3cm} \\
\text{s.t.}
\ \ \ w \in [0,1]^N,  \| \alpha_1 \|_0 \leq K_1 ,  \| \alpha_2 \|_0 \leq K_2 \hspace{6.85cm}
\end{aligned} \label{eq:ch4_11}
\end{equation}
where $u_1$ and $u_2$ denote the dual variables.
Now we can solve this problem by minimizing the Augmented Lagrangian w.r.t. to primal variables ($\alpha_1$, $\alpha_2$, $w$, $y$ and $z$) and using dual ascent for dual variables ($u_1$, $u_2$). 
For updating the variables $\alpha_1$ and $\alpha_2$, we first ignore the constraints and take the derivative of $L$ w.r.t. them and set it to zero. Then we project the solution on the constraint $\| \alpha_i \|_0 \leq K_i $ by keeping the $K_i$ largest components.
Since the cost function is symmetric in $\alpha_1$ and $\alpha_2$, we only show the solution for $\alpha_2$ here.
The solution for $\alpha_1$ is very similar.

\begin{small}
\begin{equation}
\begin{aligned}
  \alpha_2=  \underset{ \alpha_2}{\text{\ \ argmin}}
  L(\alpha_1, \alpha_2, w, y, z, u_1, u_2) = \hspace{2.8cm}
  \\ \underset{ \alpha_2}{\text{\ \ argmin}} 
 \  \| x- (\textbf{1}-w)\circ  P_1\alpha_\textbf{1}-w \circ  P_2\alpha_2  \|_2^2= \hspace{2.1cm} \\
  \underset{ \alpha_2}{\text{\ \ argmin}} 
 \  \| x- (I-W) P_1\alpha_\textbf{1}-W  P_2\alpha_2  \|_2^2 \Rightarrow \hspace{2.5cm}\\
  \nabla_{\alpha_2}L=0 \Rightarrow P_2^tW^t \big( WP_2\alpha_2+(I-W)P_1\alpha_1-x \big)=0 \hspace{0.54cm} \\
 \Rightarrow \alpha_2= (P_2^tW^tWP_2)^{-1} P_2^tW^t (x-(I-W)P_1\alpha_1) \hspace{1.27cm}
\end{aligned} \label{eq:ch4_12}
\end{equation}
\end{small}
\hspace{-0.2cm}We then keep the $K_2$ largest components of the above $\alpha_2$, which is denoted by: $\alpha_2^*= \Pi_{top-K_2}(\alpha_2)$.

We now show the optimization with respect to $w$.
We solve this optimization by first ignoring the constraint, and then projecting the optimal solution of the cost function onto the feasible set ($w \in [0,1]^n$).
It basically follows the same methodology, we just need to notice that $\text{diag}(w) P_2\alpha_2$ is the same as
$ \text{diag}(P_2\alpha_2) w $.
Therefore we will get the following optimization for $w$:
\begin{small}
\begin{equation}
\begin{aligned}
w=  \underset{ w}{\text{\ argmin}}
\ \frac{1}{2} \| x- \text{diag}(P_1\alpha_1)(\textbf{1}-w) - \text{diag}(P_2\alpha_2) w  \|_2^2 \\
+u_1^t (w-y)+ u_2^t (Dw-z)+ \frac{\rho_1}{2} \|w-y\|_2^2+ \frac{\rho_2}{2} \|Dw-z\|_2^2  \hspace{-0.1cm}  
\end{aligned} \label{eq:ch4_13}
\end{equation}
\end{small}

We can rewrite this problem as:
\begin{small}
\begin{equation}
\begin{aligned}
w= \underset{w}{\text{argmin}}
 \  \ \frac{1}{2} \| h - C w  \|_2^2+ \frac{\rho_1}{2} \|w-y\|_2^2+ \frac{\rho_2}{2} \|Dw-z\|_2^2   + u_1^t (w-y)+ u_2^t (Dw-z) 
\end{aligned} \label{eq:ch4_14}
\end{equation}
\end{small}
\hspace{-0.27cm} where $C= \text{diag}(P_2\alpha_2)-\text{diag}(P_1\alpha_1)= \text{diag}(P_2\alpha_2-P_1\alpha_1)$, and $h= x- P_1\alpha_1$.
If we take the derivative w.r.t. $w$ and set it  to zero we will get:

\begin{small}
\begin{equation}
\begin{aligned}
C^t(Cw-h)+\rho_1 (w-y)+ \rho_2 D^t(Dw-z)+u_1+D^t u_2=0 \Rightarrow \\
 (C^tC+ \rho_2 D^t D+ \rho_1 I)w= C^t h+ \rho_1 y+ \rho_2 D^t z-u_1-D^t u_2 \Rightarrow \\
 w= M_w^{-1} (C^t h+ \rho_1 y+ \rho_2 D^t z-u_1-D^t u_2) \hspace{3.01cm}
\end{aligned} \label{eq:ch4_15}
\end{equation}
\end{small}
\hspace{-.24cm} where $M_w=(C^tC+ \rho_2 D^t D+ \rho_1 I) $.
After finding $w$ using the above equation, we need to project them on the set $w \in [0,1]^n$, which basically maps any negative number to 0, and any number larger than 1 to 1. Denoting the projection operator by $\Pi_{[0,1]}$, the optimization solution of the $w$ step would be:
\begin{small}
\begin{equation}
\begin{aligned}
 \ w= \Pi_{[0,1]}\big( M_w^{-1} (C^t h+ \rho_1 y+ \rho_2 D^t z-u_1-D^t u_2) ) 
\end{aligned}   \label{eq:ch4_16}
\end{equation}
\end{small}

The optimization w.r.t. $y$ and $z$ are quite simple, as they result in a soft-thresholding solution \cite{soft}.
The overall algorithm is summarized in Algorithm 2.

\begin{algorithm}
  \caption{pseudo-code for variable updates of problem in Eq. \eqref{eq:ch4_11}}
  \label{euclid}
  \begin{algorithmic}[1]
   \STATE Given a block of size NxN, represented by a vector x, and the subspace matrices $P_1$ and $P_2$, and preset values for parameters $\rho_1$, $\rho_2$, $T_{max}$, initialize the loss value with $L^{(0)}= 1$, and: \vspace{0.15cm}
      \FOR{\texttt{$j$=1:$T_{max}$}} \vspace{0.2cm} 
        \STATE $\alpha_1^{j+1}= (P_1^tW_*^tW_*P_1)^{-1} P_1^tW_*^t (x-WP_2\alpha_2^{j})$ \vspace{0.2cm}
        \STATE $\alpha_2^{j+1}= (P_2^tW^tWP_2)^{-1} P_2^tW^t (x-(I-W)P_1\alpha_1^{j+1})$ \vspace{0.2cm}
        \STATE $ w^{j+1}= \Pi_{[0,1]}\big( M_w^{-1} (C^t h^{j+1}+ \rho_1 y^{j}+ \rho_2 D^t z^{j}-u_1^{j}- D^t u_2^{j})  $   \vspace{0.2cm} 
        \STATE $ y^{j+1}=  \text{soft}( w^{j+1}+ \frac{u_1^{j}}{\rho_1}, \lambda_1/\rho_1)     $    \vspace{0.2cm}  
        \STATE $ z^{j+1}= \text{soft}( Dw^{j+1}+\frac{u_2^{j}}{\rho_2}, \lambda_2/\rho_2)    $    \vspace{0.2cm}   
         \STATE $ u_1^{j+1}= u_1^{j}+ \rho_1 (w^{j+1}-y^{j+1})   $    \vspace{0.2cm}  
         \STATE $ u_2^{j+1}= u_2^{j}+ \rho_2 (Dw^{j+1}-z^{j+1})  $    \vspace{0.2cm} 
          
         \STATE $ \textbf{if} \ \ \frac{|L^{(j)}-L^{(j-1)}|}{L^{(j-1)}} \leq 10^{-6}  $    \vspace{0.2cm} 
		\STATE    \ \ \ \ \ \text{skip the for loop}  \vspace{0.2cm} 
		\STATE \textbf{end if}   \vspace{0.2cm}         
         
      \ENDFOR \vspace{0.4cm}  
      \\ Where  $C= \text{diag}(P_2\alpha_2-P_1\alpha_1)$, $h= x- P_1\alpha_1, W= \text{diag}(w)$, $W_*= I-W$
    \\ and $M_w=(C^tC+ \rho_2 D^t D+ \rho_1 I)$
  \end{algorithmic}
\end{algorithm}
Where $\text{soft}(x,\lambda)$ denotes the soft-thresholding operator applied element-wise and defined as in Eq. \eqref{eq:soft}.

\section{Extension to Masked-RPCA}
Sparse and low-rank decomposition is a popular problem, with many applications in signal and image processing, such as matrix completion, motion segmentation, moving object detection, system identification, and optics \cite{ms1}-\cite{4opt}.
In the simplest case, this problem can be formulated as Eq. \eqref{eq:rpca1}:
\begin{equation}
\begin{aligned}
&  \ \underset{L, S}{\text{min}}
  \ \ \ \text{rank}(L)+ \lambda_1 \|S\|_0 \\
& \ \ \text{s.t.}
\ \  \ X= L+  S
\end{aligned} \label{eq:rpca1}
\end{equation}
where $L$ and $S$ denote the low-rank and sparse components of the signal $X$.
This problem is clearly ill-posed.
There have been many studies to find under what conditions this decomposition is possible, such as the work in \cite{rpca}
Usually the nuclear and $\ell_1$ norms are used to promote low-rankness and sparsity respectively \cite{fazel}.

Extensive research has been done for the case where the components are added together, but in many of the real world applications, the components are not added, but overlaid on top of each other.
In this work, we consider slightly different approach toward sparse and low-rank decomposition, where we assume the two components are super-imposed on top of each other (instead of simply being added). In this way, at each the $(i,j)$-th element of $X$, comes only from one of the components. Therefore besides deriving the sparse and low-rank component we need to find their supports. Assuming $W \in \{0,1\}^\text{NxM}$ denotes the support of $S$, we can write this overlaid signal summation as:
\begin{equation}
\begin{aligned}
X= (1-W) \circ L+ W \circ S
\end{aligned} \label{eq:rpca2}
\end{equation}
where $\circ$ denotes the element-wise product.

By assuming some prior on each component, we will be able to jointly find each component and estimate the binary mask $W$ using an optimization problem as:
\begin{equation}
\begin{aligned}
&  \ \ \ \ \ \ \ \ \underset{L, S, W}{\text{min}}
  \ \ \phi(L)+ \lambda_1 \psi(S)+ \lambda_2 \gamma(W) \ \\
& \ \text{s.t.}
\ \ W \in \{0,1\}^{\text{NxM}}, \ \ X= (1-W) \circ L+ W \circ S
\end{aligned} \label{eq:rpca3}
\end{equation}
where $\phi(.)$, $\psi(.)$, $\gamma(.)$ encodes our prior knowledge about sparse and low-rank components and the binary mask respectively.
The exact choice of these functions depends on the application. 
Here we study this problem for the case where $L$ and $S$ are low-rank and sparse, and  the binary mask is connected along each row (which in video foreground segmentation, promotes the same position to belong to the foreground in multiple frames).
Using these priors we will get the following optimization problem:
\begin{equation}
\begin{aligned}
&  \ \ \ \ \ \ \ \ \underset{L, S, W}{\text{min}}
  \ \ \text{rank}(L)+ \lambda_1 \|S\|_0+ \lambda_2 \|W\|_{2,1} \ \\
& \ \text{s.t.}
\ \ W \in \{0,1\}^{\text{NxM}}, \ \ X= (1-W) \circ L+ W \circ S
\end{aligned} \label{eq:rpca4}
\end{equation}
where $\|.\|_{2,1}$ is the sum of $\ell_2$ norm of each row, and is defined as:
\begin{equation}
\begin{aligned}
\|W\|_{2,1}= \sum_i \ \sqrt{\sum_j w_{i,j}^2}   
\end{aligned} \label{eq:rpca5}
\end{equation}

The problem in Eq (5) is not tractable because of multiple issues. First, since $W$ is binary matrix, to estimate its elements we need to solve a combinatorial problem.
We relax this issue to get a more tractable problem by approximating $W$'s element as a continuous variable $W \in [0,1]^{\text{NxM}}$. 
Second issue is that the $\text{rank}(.)$ is not a convex function.
We address this issue by approximating rank with the nuclear norm \cite{nuc_norm} which is defined as $\|L\|_*= \sum_i \sigma_i(L)$.
The last issue originates from having $\ell_0$ term in the cost function. We approximate  $\ell_0$ norm with its convex approximation, $\ell_1$ norm, to overcome this issue. Then we will get the following optimization problem:
\begin{equation}
\begin{aligned}
&  \ \ \ \ \ \ \ \ \underset{L, S, W}{\text{min}}
  \ \ \|L\|_*+ \lambda_1 \|S\|_1+ \lambda_2 \|W\|_{2,1} \ \\
& \ \text{s.t.}
\ \ W \in [0,1]^{\text{NxM}}, \ \ X= (1-W) \circ L+ W \circ S
\end{aligned} \label{eq:rpca6}
\end{equation}

Note that since the variables $W$ and $L$ (also $S$) are coupled in  (7), we cannot use methods such as ADMM \cite{admm} to solve this problem.
Here we propose an algorithm based on linearized-ADMM to solved this optimization \cite{16ladmm}, which works by alternating over variables and each time approximating the cost function with the first order Taylor series expansions.
Before diving into the details of the proposed optimization framework, let us first briefly introduce the linearized ADMM approach.

\subsection{Linearized ADMM}
Many machine learning and signal processing problems can be formulated as linearly constrained convex programs, which could be efficiently solved by the alternating direction method (ADM) \cite{admm}. 
However, usually the subproblems in ADM are easily solvable only when the linear mappings in the constraints are identities, or the variable are decoupled. 
To address this issue, Lin\cite{16ladmm} proposed  a technique, which linearizes the quadratic penalty term and adds a proximal term when solving the sub-
problems. To have better idea, consider the following problem:
\begin{equation}
\begin{aligned}
& \  \hspace{-0.01cm}\underset{x,y}{\text{min}}
 \  f(x)+g(y)   \\
& \text{s.t.}
\ \ Ax+By=C
\end{aligned} \label{eq:rpca7}
\end{equation}
where $x$ and $y$ could be vectors, or matrices.
In alternating direction method (ADM), this problem can be solved by forming the augmented Lagrangian multiplier as below:
\begin{equation}
 \mathcal{L}(x,y,u)=
 \ f(x)+g(y)+ <Ax+By-C, u>+  
 \ \ \ \rho/2 \| Ax+By-C \|^2 \label{eq:rpca8}
\end{equation}
where $u$ is the Lagrange multiplier, <.,.> is the inner product operator, and $\rho$ is the penalty term.
Then ADM optimizes this problem by alternating minimizing $\mathcal{L}(x,y,u)$ w.r.t. $x$ and $y$, as below:
\begin{equation}
\begin{aligned}
& \hspace{-0.35cm} x_{k+1}= \underset{x}{\text{argmin}} \ \mathcal{L}(x, y_k, u_k)= \ f(x)+ \frac{\rho}{2} \| Ax+By_k-C+ u_k/\rho \|^2 \\
& \hspace{-0.35cm} y_{k+1}= \underset{y}{\text{argmin}} \ \mathcal{L}(x_{k+1}, y, u_k)=  \ g(y)+ \frac{\rho}{2} \| Ax_{k+1}+By-C+ u_k/\rho \|^2 \hspace{-0.25cm} \\
& \hspace{-0.35cm} u_{k+1}= u_k+ \rho (Ax_{k+1}+By_{k+1}-C)
\end{aligned}  \label{eq:rpca9}
\end{equation}

For many cases where $f$ and $g$ are vector (or matrix) norms, and $A$ and $B$ are identity (or diagonal), the $x$ and $y$ sub-problems have closed-form solution.
But for some cases, such as for general $A$ and $B$ there is no closed-form solution for primal sub-problems, and each one of them should be solved iteratively which is not desired. 

One could introduce auxiliary variables, and introduce more constraint, which increases the memory requirements. 
A more efficient approach is to linearize the quadratic terms in the $x$ and $y$ iterations, and add a proximal term at $x_k$ and $y_k$ as below:
\begin{footnotesize}
\begin{equation}
\begin{aligned}
x_{k+1}= \ \ \underset{x}{\text{argmin}} \ f(x)+  \rho < A^T(Ax_k+By_k-C+ u_k/\rho),\ x-x_k >+ \frac{\rho \rho_A}{2} \|x-x_k\|^2  \hspace{1.2cm} \\
  =\underset{x}{\text{argmin}} \ f(x)+ \frac{\rho \rho_A}{2} \|x-x_k+ A^T(u_k/\rho+Ax_k+By_k-C)/ \rho_A \|^2 \hspace{4cm} \\
y_{k+1}= \ \ \underset{y}{\text{argmin}} \ g(y)+ 
\rho < B^T(Ax_{k+1}+By_k-C+ u_k/\rho),\ y-y_k >+ \frac{\rho \rho_B}{2} \|y-y_k\|^2  \hspace{1cm}  \\
 =\underset{y}{\text{argmin}} \ g(y)+ \frac{\rho \rho_B}{2} \|y-y_k+ B^T(u_k/\rho+Ax_{k+1}+By_k-C)/ \rho_B \|^2 \hspace{3.6cm }\\
& \hspace{-14.1cm} u_{k+1}= u_k+ \rho (Ax_{k+1}+By_{k+1}-C) 
\end{aligned} \label{eq:rpca10}
\end{equation}
\end{footnotesize}
We can use the same idea to solve the masked-RPCA problem in Eq. \eqref{eq:rpca6}.

\subsection{The Proposed Optimization Framework}
To solve the optimization problem in Eq. \eqref{eq:rpca6} we first form the Augmented Lagrangian function as:
\begin{equation}
\begin{aligned}
& \hspace{-14.1cm} \mathcal{L}(L,S,W,U)=
  \| L \|_*+ \lambda_1 \|S\|_1+    \\ \lambda_2 \|W\|_{2,1}+ <U, X- (1-W) \circ L+ W \circ S>+ \frac{\rho}{2} \| X- (1-W) \circ L+ W \circ S  \|_2^2 \hspace{-0.1cm} \\
& \hspace{-14cm} \text{s.t.}
\ \ \ \ \ \ W \in [0,1]^{\text{NxM}}
\end{aligned} \label{eq:rpca11}
\end{equation}
where $U$ is a matrix of the same size as $X$, and $\rho$ is the penalty term.
The augmented Lagrangian function can be written in a more compact way as below:
\small
\begin{equation}
\begin{aligned}
&  \mathcal{L}(L,S,W,U)=
 \  \| L \|_*+ \lambda_1 \|S\|_1+ \lambda_2 \|W\|_{2,1}   +  \frac{\rho}{2} \| X- (1-W) \circ L- W \circ S+ U/\rho  \|_2^2 \\
& \text{s.t.}
\ \ \ \ \ \ W \in [0,1]^{\text{NxM}}
\end{aligned} \label{eq:rpca12}
\end{equation}

\normalsize
We then solve this problem by linearized-ADM approach explained above.
This would lead to the following sub-problems:
\begin{footnotesize}
\begin{equation}
\begin{aligned}
& \hspace{-0.02cm} L_{k+1}= \ \ \underset{L}{\text{argmin}} \ \|L\|_*+  < \rho \big( (1-W_k) \circ L_k+ W_k \circ S_k- X -U_k/\rho \big) \circ (1-W_k), \ L-L_k>  \\
& \hspace{-0.03cm} +\frac{\rho \rho_L}{2} \|L-L_k\|_F^2 =  \underset{L}{\text{argmin}} \ \big( \|L\|_*+ \frac{\rho \rho_L}{2} \|L-A_k\|_F^2 \big) \\
& \hspace{-0.02cm} S_{k+1}= \ \ \underset{S}{\text{argmin}} \ \|S\|_1+ < \rho \big( (1-W_k) \circ L_{k+1}+ W_k \circ S_k- X -U_k/\rho \big) \circ W_k, \ S-S_k>  \\
& \hspace{-0.03cm} +\frac{\rho \rho_S}{2} \|S-S_k\|_F^2 =  \underset{S}{\text{argmin}} \ \big( \|S\|_1+ \frac{\rho \rho_S}{2} \|S-B_k\|_F^2 \big) \\
& \hspace{-0.02cm} W_{k+1}= \ \underset{W}{\text{argmin}} \ \|W\|_{2,1}+
< \rho \big( (L-S) \circ W_k - L+ X+ U/\rho \big) \circ (L-S), \ W-W_k> \\
& \hspace{-0.03cm} +\frac{\rho \rho_W}{2} \|W-W_k\|_F^2 =  \underset{W}{\text{argmin}} \ \big( \|W\|_{2,1}+ \frac{\rho \rho_W}{2} \|W-C_k\|_F^2 \big) \\
& \hspace{-0.011cm} U_{k+1}= U_k+ \rho \ (X- (1-W_{K+1}) \circ L_{K+1}- W_{K+1} \circ S_{K+1} )
\end{aligned} \label{eq:rpca13}
\end{equation}
\end{footnotesize}
\normalsize
where:
\small
\begin{equation}
\begin{aligned}
A_k= L_K + \frac{1}{\rho_L} \big( (1-W_k) \circ L_k+ W_k \circ S_k- X -U_k/\rho \big) \circ  (1-W_k)  \\ 
B_k= S_k+ \frac{1}{\rho_S} \big( (1-W_k) \circ L_{k+1}+ W_k \circ S_k- X -U_k/\rho \big) \circ W_k \ \ \ \ \  \\
C_k= W_k+ \frac{1}{\rho_W} \big( (L-S) \circ W_k - L+ X+ U/\rho \big) \circ (L-S) \ \ \ \ \ \ \ \ \ \ \
\end{aligned} \label{eq:rpca14}
\end{equation}

\normalsize
The solution for the first step, $L$, can be found using singular value thresholding \cite{svt} as below:
\begin{equation}
\begin{aligned}
L=  \underset{ L}{\text{\ argmin}}
\ \| L \|_*+ + \frac{\rho \rho_L}{2} \|L-A_k\|_F^2 
\Rightarrow L= D_{1/\rho \rho_L} \big(A_k \big)
\end{aligned} \label{eq:rpca15}
\end{equation}
where $D_{\tau}(Y)$ refers to singular value thresholding of $Y= U \Sigma V^T$, and is defined as:
\begin{equation}
\begin{aligned}
D_t(Y)= U {S_{\tau}}(\Sigma) V^T \hspace{0.86cm}\\
{S_{\tau}}(\Sigma)_{ii}= max( \Sigma_{ii}- \tau, 0)
\end{aligned} \label{eq:rpca16}
\end{equation}

The solution for the second step, $S$, is straightforward, and it leads to the following soft-thresholding operator \cite{soft}:
\begin{equation}
\begin{aligned}
S=  \underset{ S}{\text{\ argmin}}
\ \| S \|_1+ + \frac{\rho \rho_S}{2} \|S-B_k\|_F^2 
\Rightarrow S= \text{Soft}_{1/\rho \rho_S} \big(B_k \big)
\end{aligned}
\end{equation}

We now show the optimization with respect to $W$.
We solve this optimization by first ignoring the constraint, and then projecting the optimal solution of the cost function onto the feasible set ($w \in [0,1]^n$).
Let us denote the $i$-th row of $W$ as $w_i$, then the update rule of $w_i$ can be found as:
\begin{equation}
\begin{aligned}
w_i=  \underset{ w_i}{\text{\ argmin}}
\ \| w_i \|_2+ \frac{\rho \rho_W}{2} \|w_i-c_{k,i}\|_F^2
\Rightarrow
w_i=  \text{block-soft}( \ c_{k,i}, \ \frac{1}{\rho \rho_W} ) 
\end{aligned}
\end{equation}
where block-soft(.) \cite{groupshervin} is defined as:
\begin{equation*}
\begin{aligned}
\text{block-soft}(x; \lambda)= max(0, 1- \lambda/ \|x\|_2) \ x
\end{aligned}
\end{equation*}

After deriving $w_i$ using the above equation, we need to project it onto the set $w \in [0,1]^n$, which basically maps any negative number to 0, and any number larger than 1 to 1. If we show the projection operator by $\Pi_{[0,1]}$, then the optimization solution of the $w$ step would be:
\begin{equation}
\begin{aligned}
 \ w_i= \Pi_{[0,1]}\big( \text{block-soft}( \ c_{k,i}, \ \frac{1}{\rho \rho_W} )  \big) 
\end{aligned} 
\end{equation}

Note that at the end, we need to have binary values for $W_{i,j}$s, as they show the support of the second component.
After the algorithm is converged, we threshold the $W_{i,j}$ to get a binary value. The threshold value can be derived by evaluating this scheme on a set of validation data.
One can also think of them as some membership probability for both components, which in that case we can skip thresholding.
The overall algorithm is summarized in Algorithm 3.

\begin{algorithm}
  \caption{Update Rule for Problem in Eq. \eqref{eq:rpca12}}
  \label{euclid}
  \begin{algorithmic}[1]
      \FOR{\texttt{$k$=1:$K_{max}$}}  \vspace{0.2cm}  
        \STATE $L^{k+1}= D_{1/\rho \rho_L} \big(A_k \big)$ \vspace{0.2cm}  
        \STATE $S^{k+1}= \text{Soft}_{1/\rho \rho_S} \big(B_k \big)$ \vspace{0.2cm}  
        \FOR{\texttt{$i$=1:$M$}} \vspace{0.2cm}  
        \STATE $w_i^{k+1}= \Pi_{[0,1]}\big( \text{block-soft}( \ c_{k,i}, \ \frac{1}{\rho \rho_W} )  \big) $  \vspace{0.2cm}   
        \ENDFOR \vspace{0.2cm}  
        \STATE $ U_{k+1}= U_k+ \rho \ (X- (1-W_{K+1}) \circ L_{K+1}- W_{K+1} \circ S_{K+1} ) $ \vspace{0.2cm}    
      \ENDFOR \vspace{0.2cm}  
      \\ Output: The components $L$, $S$, and binary mask $W$. Where: \vspace{0.4cm}  
     \\ $A_k= L_K + \frac{1}{\rho_L} \big( (1-W_k) \circ L_k+ W_k \circ S_k- X -U_k/\rho \big) \circ  (1-W_k)$      \\  $B_k= S_k+ \frac{1}{\rho_S} \big( (1-W_k) \circ L_{k+1}+ W_k \circ S_k- X -U_k/\rho \big) \circ W_k $.
     \\ $C_k= W_k+ \frac{1}{\rho_W} \big( (L-S) \circ W_k - L+ X+ U/\rho \big) \circ (L-S)$
  \end{algorithmic}
\end{algorithm}

The main goal of this section is to show that the extension of many of the traditional signal decomposition problems to the overlaid model (such as RPCA), can be solved with a relatively similar algorithm, without too much overhead. 
Therefore, we skip presenting the experimental results for the Masked-RPCA in this Chapter, as RPCA is not the main focus of this thesis.

\section{Application for Robust Motion Segmentation}
One potential application of the proposed formulation is for moving object detection under global camera motion in a video, or essentially segmentation of a motion field into regions with global motion and object motions, respectively.

Suppose we use the homography mapping (also known as perspective mapping) to model the camera motion, where each pixel in the new frame is related to its position in the previous frame as shown in Eq. \eqref{eq:motion1}:
\begin{equation}
\begin{aligned}
x_{new}= \frac{a_1+a_2x+a_3y}{1+a_7x+a_8y}  \\
y_{new}= \frac{a_4+a_5x+a_6y}{1+a_7x+a_8y}
\end{aligned}  \label{eq:motion1}
\end{equation}
Let $u=x_{new}-x$ and $v=x_{new}-x$, we can rewrite the above equation as:
\begin{equation}
\begin{aligned}
(x+u)(1+a_7x+a_8y)= (a_1+a_2x+a_3y)  \\
(y+v)(1+a_7x+a_8y)= (a_4+a_5x+a_6y)
\end{aligned}  \label{eq:motion2}
\end{equation}
For each pixel $(x,y)$ and its motion vector $(u,v)$, we will get two equations for the homograph parameters $a= [a_1,...,a_8]^T$, that can be written as below:
\begin{small}
\begin{equation}
\begin{bmatrix} 
    1&x&y&0&0&0&-x(x+u)&-y(x+u) \\ 0&0&0&1&x&y&-x(y+v)&-y(y+v)
\end{bmatrix} a= 
\begin{bmatrix} 
    x+u \\ y+v
\end{bmatrix} \label{eq:motion3}
\end{equation}
\end{small}
\hspace{-0.28cm} Using  the equations at all pixels, we will get a matrix equation as:
\begin{equation}
\begin{aligned}
Pa=b
\end{aligned} \label{eq:motion4}
\end{equation}

Suppose $u(x,y)$ and $v(x,y)$ are derived using a chosen optical flow estimation algorithm. 
Then the goal is to find the global motion parameters and the set of pixels which do not follow the global motion. 
Note that here, $P$ will not be the same for different video frames, as it depends on the optical flow (which could be different for different frames).
Assuming there are some outliers (corresponding to moving objects) in the video, we can use the model $b=(\textbf{1}-w) \circ Pa+w \circ s$, where $w$ denotes the outlier pixels, and $s$ denotes the new location for the outlier pixels.
If the outlier pixels belong to a single object with a consistent local motion, we can model $s$ as $s=P_2 a_2$.
However, in general, the outlier pixels may correspond  to multiple foreground objects with different motions or different subregions of a single object (e.g. different parts of a human body) with different motions.  Therefore, we do not want to model s with a single parameterized motion. Rather we will directly solve for s with a sparsity  constraint.  These considerations lead to the following optimization problem:
\begin{small}
\begin{equation}
\begin{aligned}
& \underset{w, a, s}{\text{min}}
 \  \ \frac{1}{2} \| b- (\textbf{1}-w)\circ  Pa- w \circ  s  \|_2^2+ \lambda_1 \| s \|_1   + \lambda_2 \| w \|_1+\lambda_3 \|Dw \|_1 \\
&  \text{s.t}
\ \ \   w \in [0,1]^n
\end{aligned} \label{eq:motion5}
\end{equation}
\end{small}
Note that $b$, $P$ and $s$ in Eq. \eqref{eq:motion5} have two parts, one corresponding to the horizontal direction, and another part corresponding to vertical direction (denoted with subscripts $x$ and $y$ respectively). Therefore we can re-write this problem as:
\begin{equation}
\begin{aligned}
& \underset{w, a, s}{\text{min}}
 \  \ \frac{1}{2} \| b_x- (\textbf{1}-w)\circ  P_xa- w \circ  s_x  \|_2^2+ \lambda_1 \| s_x \|_1+ \lambda_1 \| s_y \|_1     \\ 
 & \ \ \ \ \ \ \ \frac{1}{2} \| b_y- (\textbf{1}-w)\circ  P_ya- w \circ  s_y  \|_2^2+ \lambda_2 \| w \|_1+ \lambda_3 \| Dw \|_1 \\
& \ \ \text{s.t}
\ \ \ \  w \in [0,1]^n
\end{aligned}  \label{eq:motion6}
\end{equation}

This problem can be solved with ADMM. After solving this problem we will get the mask for the moving objects.
Note that this approach works for other global motion models such as the affine mapping.
In the extended version of this algorithm, we can directly work on a volume of $\tau$ frames to use the temporal information for mask extraction. In that case, the mask $w$ would be a 3D tensor.
In the experimental result section, we provide the result of motion segmentation using the proposed algorithm.

\section{Experimental Results}
In this section we provide the experimental study on the application of the proposed algorithm for 1D signals decomposition, image segmentation, and also motion segmentation. 
For each application, different sets of parameters are used, which are tuned on a validation data from the same task.

\subsection{1D signal decomposition}
\begin{figure}[h]
\begin{center}
 \hspace{-0.3cm} \includegraphics [scale=0.8] {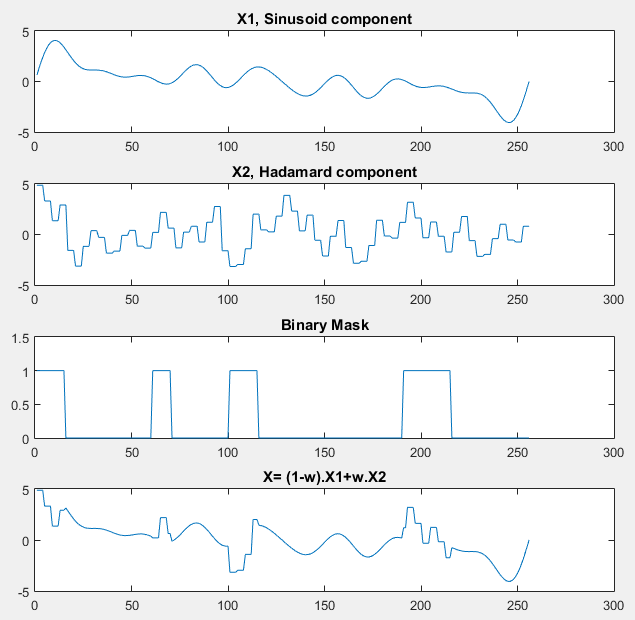} 
\end{center}
 \vspace{-0.1cm} \caption{The binary mask, and signal components} \label{fig:fig4_1}
\end{figure}

To illustrate the power of the proposed algorithm for non-additive decomposition, we first experiment with a toy example using 1D signals. We generate two 1D signals, each 256 dimensional, using different subspaces. The first signal is generated from a 10-dimensional sinusoid subspace, and the second component is generated from a 10-dimensional Hadamard subspace.
We then generate a random binary mask with the same size as the signal, and added these two components using the mask as: $x= (\textbf{1}-w)\circ x_1+w\circ x_2$.
The goal is to separate these two components, and estimate the binary mask.
The signal components and binary mask for one example are shown in Figure~\ref{fig:fig4_1}.

\begin{figure}[H]
\begin{center}
 \hspace{-0.3cm} \includegraphics [scale=0.45] {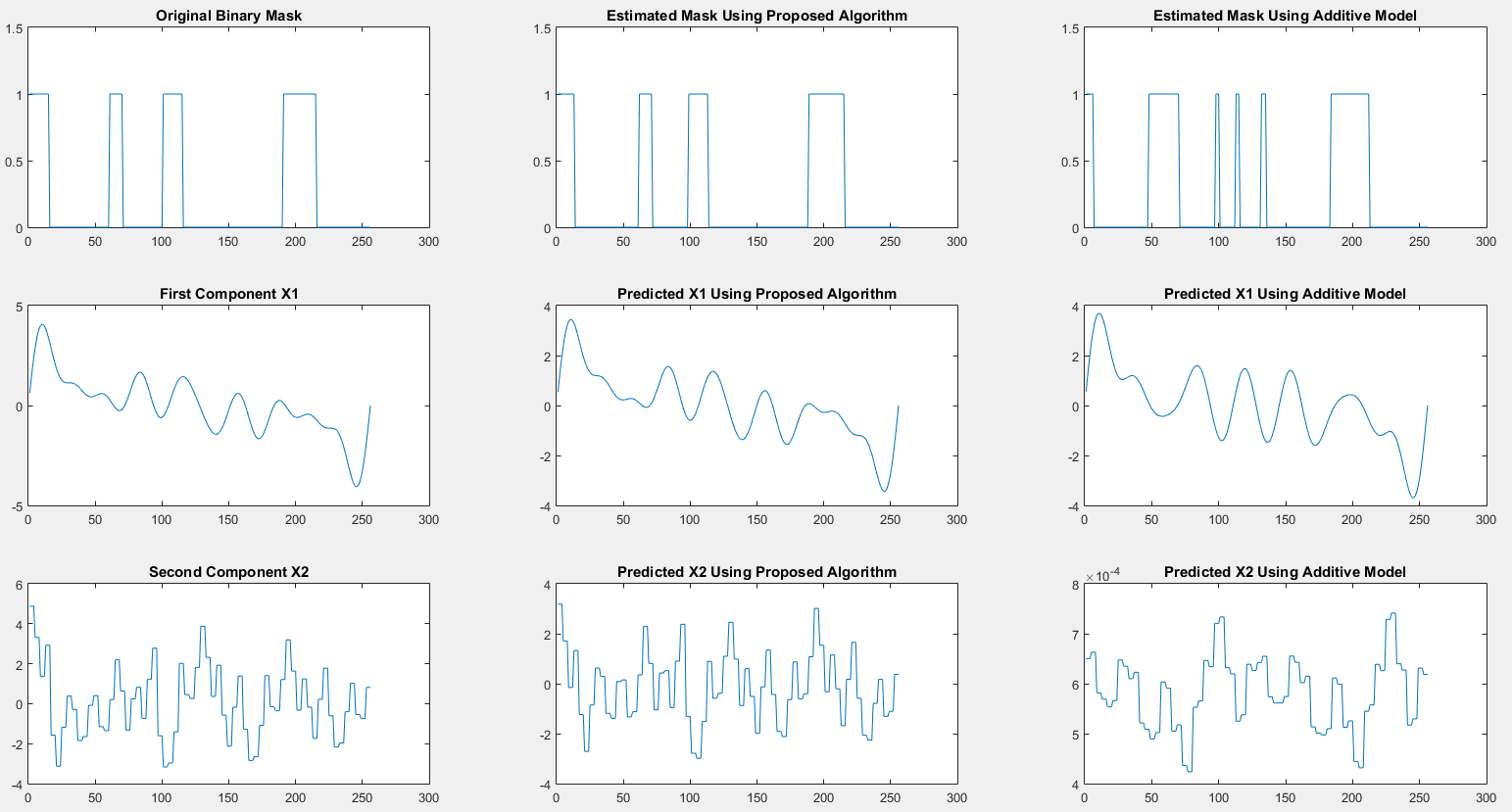} 
\end{center}
\begin{center}
 \hspace{-0.3cm} \includegraphics [scale=0.45] {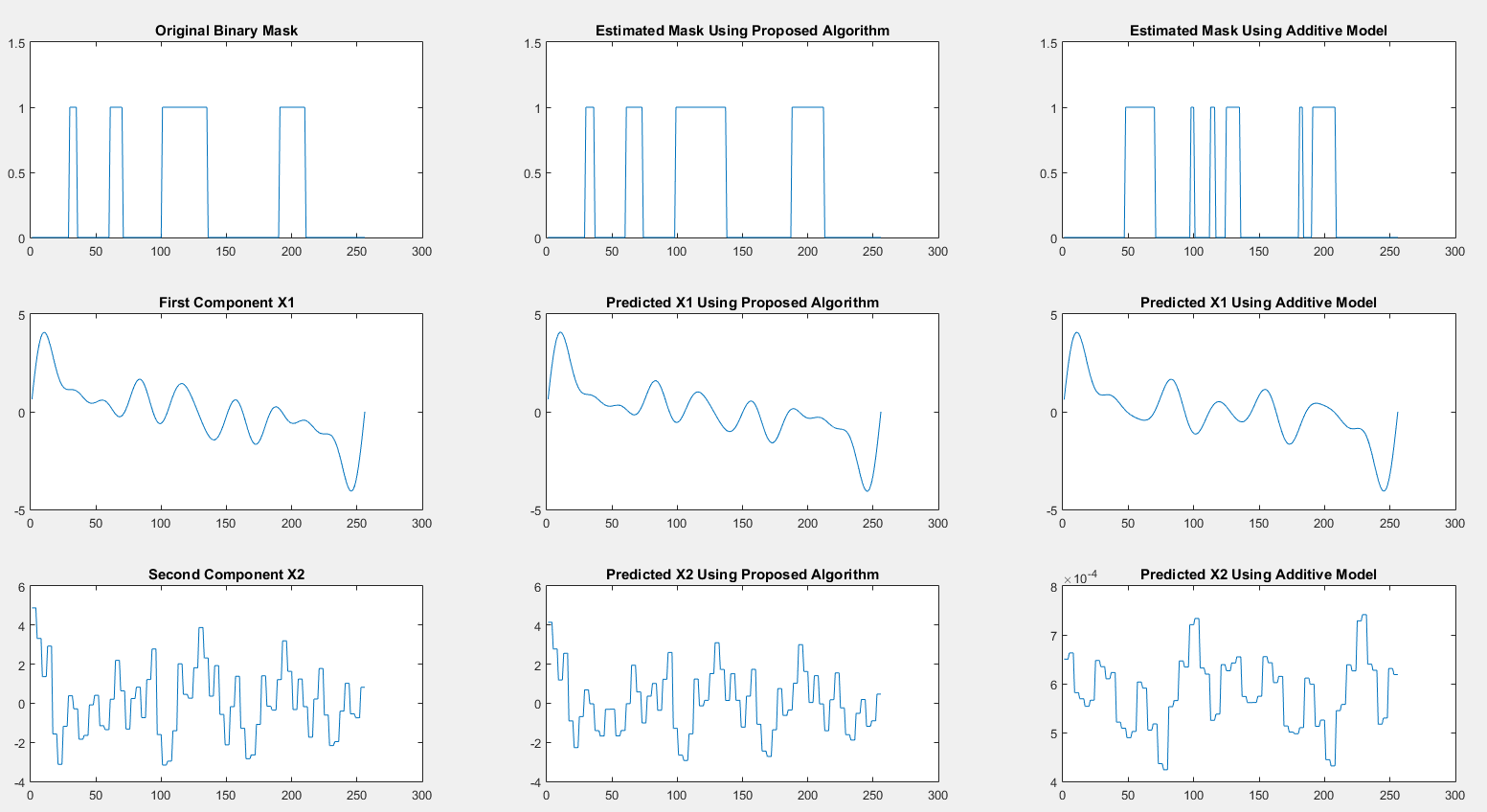} 
\end{center}
\caption{The 1D signal decomposition results for two examples. The figures in the first column denotes the original binary mask, first and second signal components respectively. The second and third columns denote the estimated binary mask and signal components using  the proposed algorithm and the additive model signal decomposition, respectively.} \label{fig:fig4_2}
\end{figure}

We then use the proposed model to estimate each signal component and extract the binary mask, and compare it with the signal decomposition under additive model. By additive model we mean the following optimization problem:
\begin{equation}
\begin{aligned}
& \underset{\alpha_1, \alpha_2}{\text{min}}
 \  \frac{1}{2} \| x-  P_1\alpha_1-  P_2\alpha_2  \|_2^2+ \lambda_1 \| P_2\alpha_2 \|_1+ \lambda_2 TV(P_2\alpha_2)    \\
& \ \text{s.t.}
\ \ \ \ \ \  \| \alpha_1 \|_0 \leq k_1 , \ \| \alpha_2 \|_0 \leq k_2
\end{aligned} \label{eq:additive}
\end{equation}
We need to mention that for the above additive model, the binary mask is derived by thresholding the values of the second component (we adaptively chose the threshold values such that it yields the best visual results).
The estimated signal components and binary mask by each algorithm, for two examples are shown in Figure~\ref{fig:fig4_2}. 
In our experiment, the weight parameters for the regularization terms in Eq. \eqref{eq:additive} are chosen to be $\lambda_1= 0.3$ and $\lambda_2= 10$.
The number of iterations for alternating optimization algorithm is chosen to be 20. 
As it can be seen the proposed algorithm achieves much better result than the additive signal model.
This is as expected, because the additive model could try to model some parts of the second component with the first subspace and vice versa.

\subsection{Application in Text/Graphic Segmentation From Images}
Next we test the potential of the proposed algorithm for the text and graphic segmentation from images.
We perform segmentation on two different sets of images. 
The first one is on a dataset of screen content images, which consist of 332 image blocks of size 64x64, extracted from sample frames of HEVC test sequences for screen content coding \cite{SCC_data}, \cite{SCC_tran}.
The second set of images are generated manually by adding text on top of other images.

We apply our algorithm on blocks of 64x64 pixels. We first convert each block into a vector of  dimension 4096, and then apply the proposed algorithm.
For the smooth background we use low-frequency DCT basis with $k_1=40$, and for the second component we use Hadamard basis with $k_2=8$.
The weight parameters for the regularization terms are chosen to be $\lambda_1= 10$ and $\lambda_2= 0.2$, which are tuned by testing on a separate validation set of more than 50 patches.
The number of iterations for alternating optimization algorithm is chosen to be 10.
We compare the proposed algorithm with four previous algorithms: hierarchical k-means clustering in DjVu \cite{djvu}, SPEC \cite{spec}, least absolute deviation fitting (LAD) \cite{lad}, and sparsity based signal decomposition \cite{mytv}-\cite{mytv2}.

\begin{figure}[H]
\begin{center}
\includegraphics [scale=0.75] {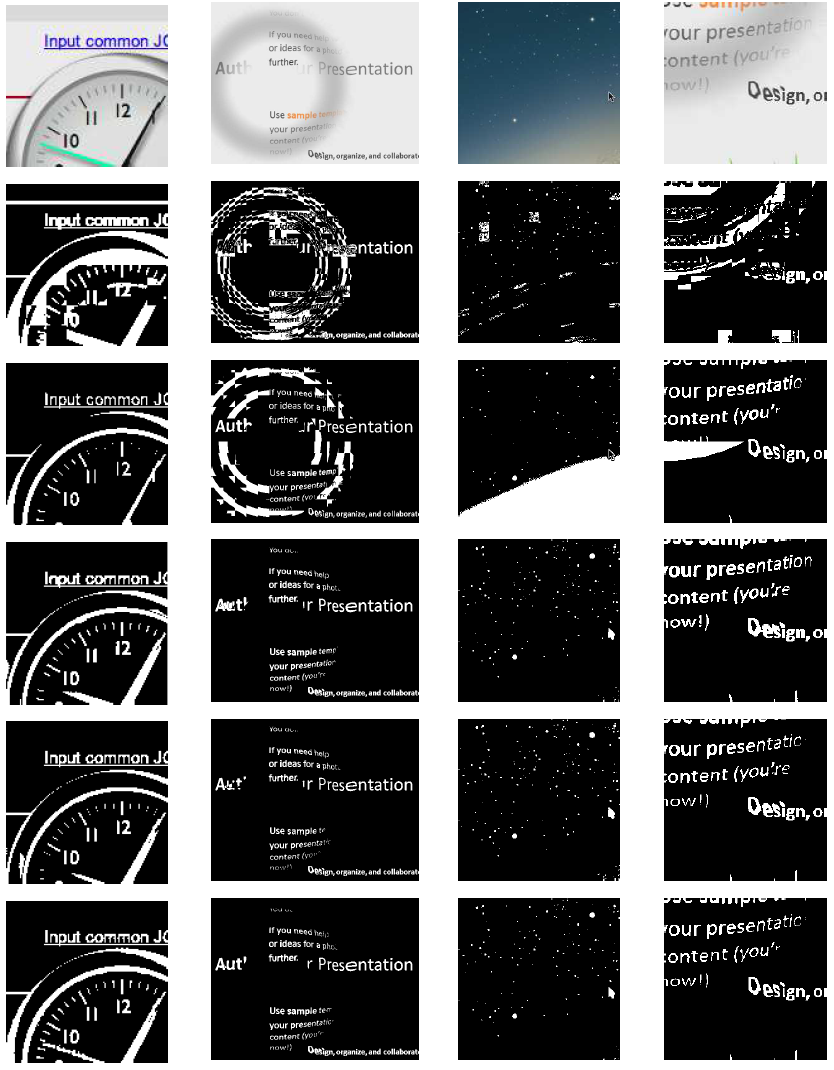}  
\end{center}
\caption{Segmentation result for selected test images for screen content image compression. The images in the first row denotes the original images. And the images in the second, third, fourth, fifth and the sixth rows denote the extracted foreground maps by shape primitive extraction and coding, hierarchical k-means clustering, least absolute deviation fitting, sparse decomposition, and the proposed algorithm respectively.} \label{fig:fig4_3}
\end{figure}

The results for 4 test images (each consisting of multiple 64x64 blocks) are shown in Figure~\ref{fig:fig4_3}. 

It can be seen that the proposed algorithm gives superior performance over DjVu and SPEC in all cases.
There are also noticeable improvement over our prior works on LAD  and sparse decomposition based image segmentation. For example, in the left part of the second image (around the letters AUT), and in the left part of the first image next to the image border, where the LAD algorithm detects some part of background as foreground.
We would like to note that, this dataset mainly consists of challenging images where the background and foreground have overlapping color ranges. For simpler cases where the background has a narrow color range that is quite different from the foreground, both DjVu and least absolute deviation fitting will work well. 

\begin{figure}[H]
\begin{center}
\includegraphics [scale=0.72] {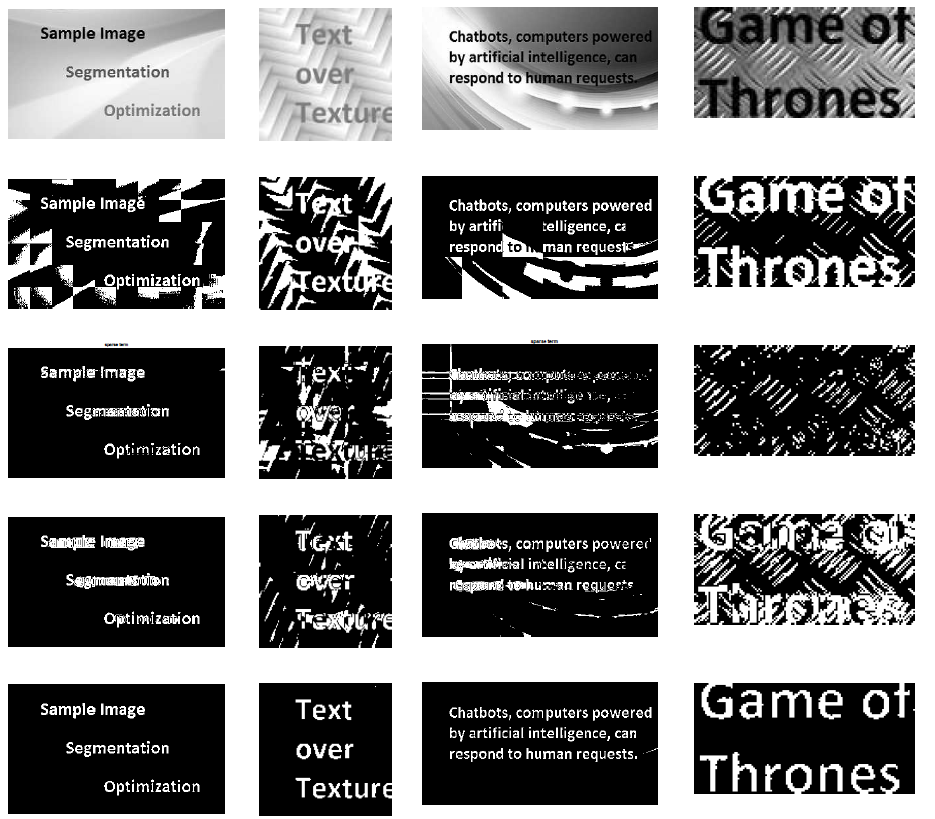} 
\end{center}
\caption{Segmentation result for the text over texture images. The images in the first row denotes the original images. And the images in the second, third, fourth and the fifth rows denote the foreground map by hierarchical k-means clustering \cite{djvu}, sparse and low-rank decomposition \cite{lowrank}, sparse decomposition \cite{mytv}, and the proposed algorithm respectively.} \label{fig:fig4_4}
\end{figure}

In another experiment, we manually added text on top of an image, and tried to extract them using the proposed algorithm.
Figure~\ref{fig:fig4_4} shows the comparison between the proposed algorithm and the previous approaches.
For this part we also provide the results derived by the method of sparse and low-rank decomposition \cite{lowrank}, using the MATLAB implementation provided in \cite{becker}.
Essentially this method assumes the background image block is low rank and the text part is sparse.
To derive the foreground map using this approach, we threshold the absolute value of the sparse component after decomposition.
For all images, we see that the proposed method yields significantly better text segmentation.

We also provide the average precision, recall and F1 score achieved by different algorithms for the above sample images. 
The average precision, recall and F1 score by different algorithms are given in Table 4.1.
As it can be seen, the proposed scheme achieves much higher precision and recall than hierarchical k-means clustering and sparse decomposition approach. 
We did not provide the results by SPEC \cite{spec} algorithm for these images, since the derived segmentation masks for these test images using SPEC was not satisfactory.

\begin{table}[ht]
\centering
  \caption{Comparison of accuracy of different algorithms for text image segmentation for images in Figure~\ref{fig:fig4_4}} \vspace{0.4cm}
  \centering
\begin{tabular}{|m{4.9cm}|m{2cm}|m{2cm}|m{2cm}|}
\hline
Segmentation Algorithm  &  \  Precision & \ \ \ Recall & \ \ F1 score\\
\hline
 Hierarchical Clustering \cite{djvu} & \ \ \ 66.5\% & \ \ \ 92\% & \ \ \ 77.2\% \\
\hline
 Sparse and Low-rank \cite{lowrank} & \ \ \ 54\% & \ \ \ 62.8\% & \ \ \ 57.7\% \\
\hline 
 Sparse Dec. with TV \cite{mytv} & \ \ \  71\% & \ \ \  91.7\% & \ \ \  80\% \\
\hline
 The proposed algorithm & \ \ \ 95\%  & \ \ \ 92.5\%  & \ \ \  93.7\%\\
\hline
\end{tabular}
\label{TblComp}
\end{table}

\subsection{Application for Motion Segmentation}
In this section, we demonstrate the application of the proposed algorithm for motion based object segmentation in video. We assume the video undergoes a global camera motion (modeled by a homography mapping) as well as localized object motion (modeled by a sparse component). The optical flow field between two frames can thus being modeled by a masked decomposition of the global motion and object motion.

To extract the optical flow, we use the optical flow implementation in \cite{of1}, \cite{of2}.
We then use the formulation in Eq.~\eqref{eq:motion5}, with $\lambda_1= 1$, $\lambda_2= 0.8$ and $\lambda_3= 0.5$, to find both the global motion parameters and the object mask $w$. Note that the estimated $w$ from Eq.~\eqref{eq:motion5} is a continuous mask where each element is in [0,1], and we threshold these values to derive the binary mask for foreground.
We compare our work with the simple least squares fitting method  where we fit the homography model to the whole optical flow by solving the optimization problem in Eq. \eqref{eq:mot}, and detect foreground pixels by thresholding the fitting error image.
\begin{equation}
\begin{aligned}
& \underset{a}{\text{min}}
 \  \ \frac{1}{2} \| b_x- P_xa  \|_2^2+\frac{1}{2} \| b_y- P_ya \|_2^2
\end{aligned} \label{eq:mot}
\end{equation}

\begin{figure}[h]
\begin{center}
\includegraphics [scale=1.1] {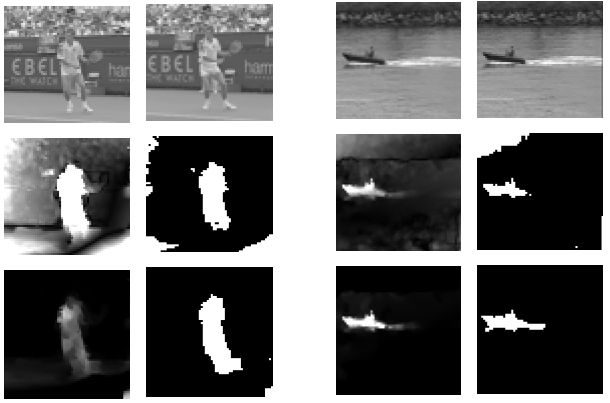} 
\end{center}
\caption{Motion segmentation result for Stefan (on the left) and Coastguard (on the right) videos. The images in the first row denote two consecutive frames from Stefan and Coastguard test video. The images in the second row denote the global motion estimation error and its corresponding binary mask. The images in the last row denote the continuous and the binary motion masks using the proposed algorithm.} \label{fig:fig4_5}
\end{figure}

The motion segmentation results using the proposed algorithm, and the comparison with the least squares fitting for two videos are provided in Figure~\ref{fig:fig4_5}.
As we can see the proposed algorithm achieves better segmentation compared to the baseline.
We would like to note that, this is a preliminary study to show the motion segmentation as one of the potential applications of this work, and the result could be much improved by using more accurate optical flow extraction scheme.

\subsection{Binarization at each step vs. at the end}
\begin{figure}[h]
\begin{center}
\includegraphics [scale=0.8] {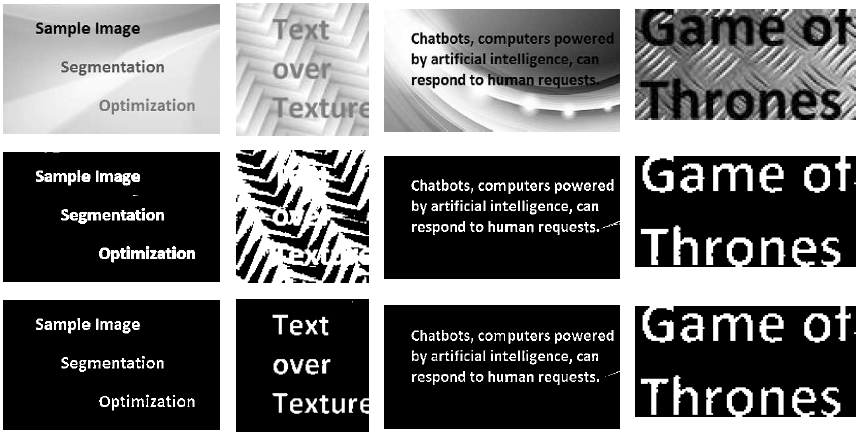} 
\end{center}
\caption{Segmentation result of the proposed method with different binarization methods. The images in the first row denotes the original images. And the images in the second and third rows show the foreground maps by binarization at the end of each iteration, and at the the very end respectively.} \label{fig:fig4_6}
\end{figure}

As mentioned earlier, the original masked decomposition problem requires the solution of a binary optimization problem.
To make it a tractable problem, we approximate the binary variables with continuous variables in $[0,1]$ (called linear relaxation), and binarize them after solving the relaxed optimization problem.
There are two ways to do this binarization: 
The first approach solves the optimization problem in Eq. \eqref{eq:ch4_9}, and binarizes the variables $w$ at the very end;
The second approach binarizes the variables $w$ after each update of $w$ in algorithm 2. 
We have tested both these approaches for some of the test images, and provided the results in Figure~\ref{fig:fig4_6}.
As we can see, doing the binarization at the very end works better for all images.
Results presented previously in Secs. A-C are all obtained with the first approach.

\subsection{Robustness to Initialization}
\begin{figure}[h]
\begin{center}
\includegraphics [scale=0.8] {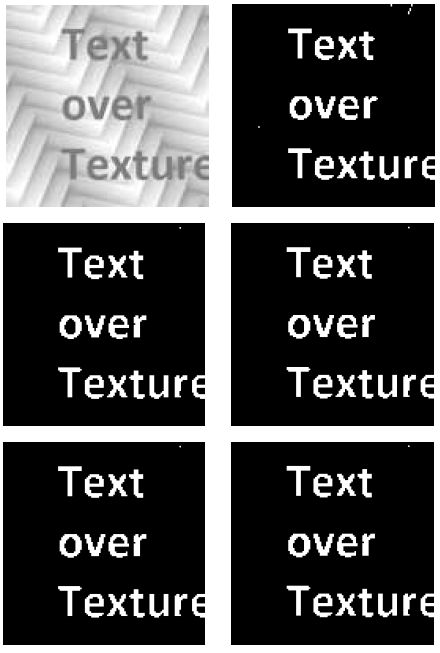} 
\end{center}
\caption{Segmentation result for different initialization schemes. The second, third, fourth, fifth and sixth images denote the segmentation results by all-zeros initialization, constant value of 0.5, zero-mean unit variance Gaussian, uniform distribution in [0,1], and error based initialization respectively.} \label{fig:fig4_7}
\end{figure}

In this section we present the stability of the algorithm with respect to the initialization of $w$.
One way to evaluate the stability of the optimization algorithm and its convergence, is to evaluate the effect of initialization in the final results.
If the final result does not depend much on the initialized values, it shows the robustness of the algorithm. To make sure the proposed algorithm is robust to the initialization, we provide the segmentation results for a test images, with 5 different initializations in Figure~\ref{fig:fig4_7}. 
The first one is to initialize the $w$ values with all zeros. The second one is to initialize them with the constant value of 0.5. The third one is to initialize them with Gaussian random variable with mean and variance equal to 0.5 and 0.1 respectively (and clipping the values to between 0 and 1).
The fourth one is to initialize them with uniform distribution in [0,1]. 
And the last scheme is to perform least squares fitting using $P_1$ only as the basis, and consider the pixels with large fitting error as foreground.
It is worth mentioning that the number of iterations in our optimization is set to 10, which is not very large to make the effect of initialization disappear.
As we can see the segmentation results with different initialization schemes are roughly similar, showing the robustness of this algorithm.

\subsection{Convergence Analysis}
\begin{figure}[h]
\begin{center}
    \includegraphics [scale=0.51] {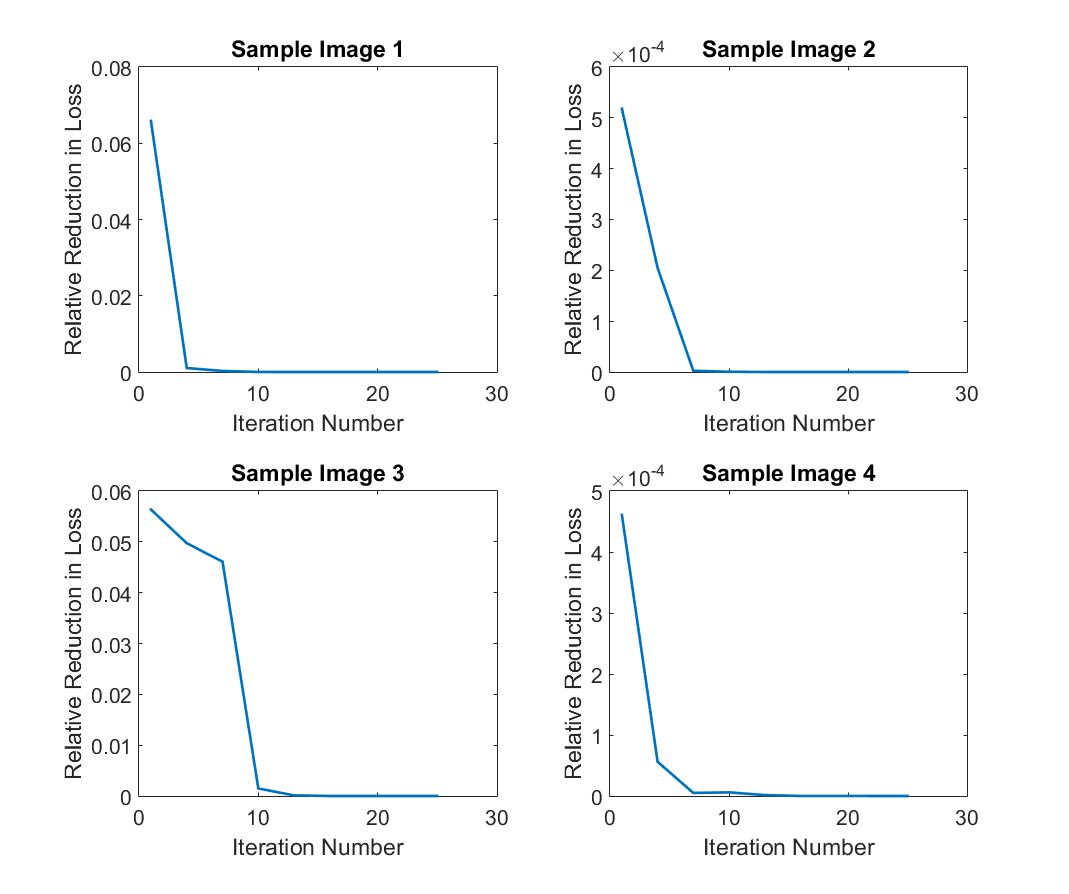}
  \caption{The relative loss reduction for four images.} \label{fig:fig4_8}
\end{center}
\end{figure}

The optimization problem in Eq. \eqref{eq:ch4_7} is a mixed integer programming problem, and is very difficult to solve directly.
In this work, we solve a relaxed constrained optimization problem defined in Eq. \eqref{eq:ch4_9}, and then binarize the resulting mask image. 
The relaxed problem is still a bi-convex problem as it involves bi-linear terms of the unknown variables (product of $w$ and $\alpha$).
We solve this problem iteratively using the ADMM method. 
We provide experimental convergence analysis by looking at the reduction in the loss (Eq. \eqref{eq:ch4_9}) at successive iterations. 
Specifically we look at the absolute relative loss reduction, calculated as $\frac{| L^{(k+1)}-L^{(k)} |}{L^{(k)}}$ over different iterations, where $L^{(k)}$ denotes the loss function value at $k$-th iteration.
The experimental convergence analysis for 4 sample images are shown in  Figure~\ref{fig:fig4_8}.

As we can see from this figure, the loss reduction keeps decreasing until it converges to zero typically under 10 iterations. This is why we set the maximum iteration number to 10 for the experimental results shown earlier.
In terms of computational time, it takes around 2 seconds to solve this optimization for an image block of size 64x64, using MATLAB 2015 on a Laptop with core i-5 CPU running at 2.2 GHz.
This can be order of magnitudes faster by running it on a more powerful machine and possibly on GPU.

\subsection{Choice of Subspaces}
\begin{figure}[h]
\begin{center}
    \includegraphics [scale=0.9] {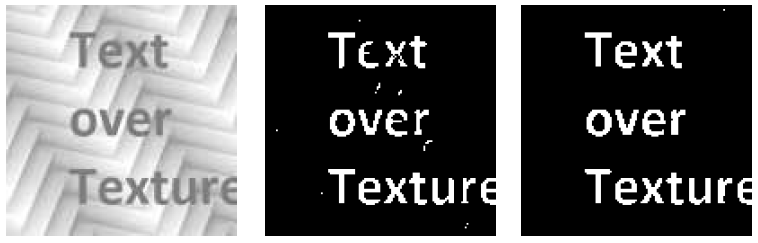}
  \caption{The left, middle and right images denote the original image, the foreground maps by using DCT bases for both background and foreground, and using DCT bases for background and Hadamard for foreground respectively.} \label{fig:fig4_9}
\end{center}
\end{figure}

As we can see from the optimization problem in Eq. \eqref{eq:ch4_9}, we assume that the subspaces/dictionaries are known beforehand.
It is obvious that the choice of $P_1$ and $P_2$ significantly affects the overall performance of the proposed signal decomposition framework.
Choice of $P_1$ and $P_2$ largely depends on the applications.
One could choose these subspaces by the prior knowledge in the underlying applications. For example as shown in the experimental result section, for separation of smooth background from foreground text and graphics, DCT \cite{dct} and Hadamard subspaces \cite{had_tran} are suitable for background and foreground components respectively.
Figure~\ref{fig:fig4_9} shows a comparison between segmentation results using low-frequency DCT subspace for both background and foreground, and DCT subspace for background and Hadamard subspace for foreground.
As we can see using Hadamard bases for foreground yields better results.

\section{Conclusion}
In this chapter we looked  at  signal decomposition problem under overlaid addition, where the signal values at each point comes from one and only one of the components (in contrast with the traditional signal decomposition case, which assumes a given signal is the sum of all signal components).
This problem is formulated in an optimization framework, and an algorithm based on the augmented Lagrangian method is proposed to solve it.
Suitable regularization terms are added to the cost function to promote desired structure of each component.
We evaluate the performance of this scheme for different applications, including 1D signal decomposition, text extraction from images, and moving object detection in video.
We also provide a comparison of this algorithm with some of the previous signal decomposition techniques on image segmentation task.
As the future work, we want to use subspace/dictionary learning algorithms to learn the subspaces for our application. 
There are many algorithms available for subspace/dictionary learning \cite{dic1}-\cite{dic5}.
The ideal subspace for each component should be such that, this component can be efficiently (with small error and sparse representation) represented in that subspace, but the other component cannot be.
In case where it is possible to access training data which only consist of individual components, one could use transform learning methods (such as the KLT \cite{klt} or K-SVD algorithm \cite{dic3}) on a large training set, to derive $P_1$ and $P_2$ separately.
The more challenging problem is when we do not have access to the training data of one component only.
In that case, one could use a training set of super-imposed signals, and use an optimization framework that simultaneously performs masked decomposition and subspace learning.

\chapter{Conclusions and Future Work}
In this thesis, we sought means to improve the quality of foreground segmentation algorithms for screen content as well as mixed-content images.

\section{Summary of Main Contribution}
In this thesis, we looked at different aspects of foreground segmentation problem, and proposed several solutions to solve them.

In Chapter 2, we proposed two novel foreground segmentation algorithms, that are developed for screen content images, one based on sparse decomposition and the other one using robust regression. 
We modeled the background and foreground parts of the image as a smooth and sparse components respectively.
We proposed to use a subspace representation to model the background.
We then formulated this foreground segmentation problem as a sparse decomposition problem, and proposed an optimization algorithm based on ADMM to solve this problem.
To show the performance of these algorithms, we applied them on a dataset of screen content image segmentation, and achieved significantly better performance than previous works.

In Chapter 3, we studied the problem of robust subspace learning for modeling the background layer in the presence of structured outliers and noise.
We proposed a new subspace learning algorithm which can learn the subspace representation of the underlying signal in case it is heavily corrupted with outliers and noise. 
We showed the application of this algorithm for text extraction in images with complicated backgrounds and achieved promising results.

In Chapter 4, we presented a very novel perspective for a new class of signal decomposition problems, specifically the case where the signal components are overlaid on top of each other, rather than simple addition. 
In this case, beside estimating each signal component, we also need to estimate their supports. 
We proposed an optimization framework which can jointly estimate signal components and their supports.
We showed that this scheme could significantly improve the segmentation results for text over textures.
We also discussed about the extension of "Robust Principal Component Analysis (RPCA)" \cite{rpca}, to masked-RPCA, for doing sparse and low-rank decomposition under overlaid model.
Through experimental studies, we showed the application of this algorithm for motion segmentation in videos, and also one-dimensional signal decomposition.

In Chapter 5, we briefly discussed an end-to-end deep learning framework, which can directly segment the foreground through a convolutional auto-encoder.
We show that after training, this model is able to perform the segmentation much faster than previous methods, because it no longer needs to solve an optimization to segment the foreground.

\section{Future Research}

\subsection{Simultaneous Learning of Multiple Subspaces}
In Chapter 4, we discussed  overlaid signal decomposition, where the signal components are assumed to be well-represented with suitable subspaces as $x= (I-W) {P_1\alpha_1}+ W P_2\alpha_2$.
Here we assumed the subspaces for different components are known in advance. 
However, there could be many cases where suitable subspaces are not known, and it would be interesting to simultaneously learn these subspaces from a set of mixed signals.
There are many works that jointly learn subspace/dictionary representation for signals with different class labels \cite{multi1}-\cite{multi2}, but they assume that each signal only belongs to one class, whereas here we deal with  the case where each signal contains multiple components (different classes).
With suitable priors on each subspace and component, it is possible to jointly learn a suitable subspace representation of different components, and separate them.

\subsection{Using an End-to-End Framework for Foreground \\Segmentation Using Deep Learning}
We can also use a deep learning approach to directly segment an image using a convolutional encoder-decoder architecture. 
Similar architectures have been used for various image segmentation tasks \cite{deepseg1}-\cite{deepseg3}, but not specifically for mixed-content image segmentation.
We can add suitable regularization terms to the loss function that encourages the foreground mask to be sparse and connected (as Eq. \eqref{eq:ch4_9}).
For example, the $\ell_1$-norm and total variation of the output mask can be added to promote the sparsity and connectivity respectively.
Instead of convolutional auto-encoder on patch level, we can also use a recurrent neural network, which at each time predicts the foreground mask of a given patch, and uses that as the hidden state (memory) for segmentation mask of future patches.





\newpage
\subsection*{List of Publications}
\vspace{1cm}

\begin{enumerate}
\item{S Minaee and Y Wang, \emph{``Screen Content Image Segmentation Using Robust Regression and Sparse Decomposition''}, IEEE Journal on Emerging and Selected Topics in Circuits and Systems, no.99, pp.1-12, 2016.}
\item{S Minaee and Y Wang, \emph{``Screen Content Image Segmentation Using Least Absolute Deviation Fitting''}, IEEE International Conference on Image Processing, 2015.}
\item{S Minaee and A Abdolrashidi, Y Wang, \emph{``Screen Content Image Segmentation Using Sparse-Smooth Decomposition''}, Asilomar Conference on Signals,Systems, and Computers, IEEE, 2015 (Selected as best paper award finalist).}
\item{S Minaee and Y Wang, \emph{``Subspace Learning in The Presence of Sparse Structured Outliers and Noise''}, International Symposium on Circuits and Systems, IEEE, 2017.}
\item{S Minaee, Y. Wang, \emph{``Screen Content Image Segmentation Using Sparse Decomposition and Total Variation Minimization''},  International Conference on Image Processing, IEEE, 2016.}
\item{S Minaee and Y Wang, \emph{``Masked Signal Decomposition Using Subspace Representation and Its Applications''}, arXiv preprint arXiv: 1704.07711, 2017 (submitted to IEEE Transactions on Image Processing)}
\item{S Minaee and Y Wang, \emph{``Text Extraction From Texture Images Using Masked Signal Decomposition''}, Global Conference on Signal and Information Processing, IEEE, 2017.}

\item{S Minaee and A Abdolrashidi, Y Wang, \emph{``An Experimental Study of Deep Convolutional Features For Iris Recognition''},  IEEE Signal Processing in Medicine and Biology Symposium, 2016.}
\item{S Minaee and Y Wang, \emph{``Fingerprint Recognition Using Translation Invariant Scattering Network''},  IEEE Signal Processing in Medicine and Biology Symposium, 2015.}
\item{S Minaee and A Abdolrashidi, Y Wang, \emph{``Iris Recognition Using Scattering Transform and Textural Features''}, IEEE Signal Processing Workshop, 2015.}
\item{S Minaee and Y Wang, \emph{``Palmprint Recognition Using Deep Scattering Convolutional Network''}, International Symposium on Circuits and Systems, \\IEEE,  2017.}
\item{S Minaee and A Abdolrashidi, \emph{``Highly Accurate Multispectral Palmprint Recognition Using Statistical and Wavelet Features''}, IEEE Signal Processing Workshop, 2015.}

\item{S Minaee, Y Wang and YW Lui, \emph{``Prediction of Longterm Outcome of Neuropsychological Tests of MTBI Patients Using Imaging Features''},  IEEE Signal Processing in Medicine and Biology Symposium, 2013.}
\item{S Minaee, Y Wang, S Chung, X Wang, E Fieremans, S Flanagan, J Rath, YW Lui, \emph{``A Machine Learning Approach For Identifying Patients with Mild Traumatic Brain Injury Using Diffusion MRI Modeling''},  ASFNR 11th Annual Meeting, 2017.}
\item{S Minaee, S Wang, Y Wang, S Chung, X Wang, E Fieremans, S Flanagan, J Rath, YW Lui, \emph{``Identifying Mild Traumatic Brain Injury Patients From MR Images Using Bag of Visual Words''},  IEEE Signal Processing in Medicine and Biology Symposium, 2017.}
\item{S Minaee, Y Wang, A Choromanska, S Chung, X Wang, E Fieremans, S Flanagan, J Rath, YW Lui, \emph{``A Deep Unsupervised Learning Approach Toward MTBI Identification Using Diffusion MRI''},  International Engineering in Medicine and Biology Conference, IEEE, 2018 (submitted).}

\item{S Minaee, Z Liu, \emph{``Automatic Question-Answering Using A Deep Similarity Neural Network''},  Global Conference on Signal and Information Processing, IEEE, 2017.}
\end{enumerate}

\vspace{0.5cm}
\subsection*{List of Patents}
\vspace{0.5cm}
\begin{enumerate}
\item{S Minaee and Haoping Yu, \emph{``Apparatus and method for compressing color index map''}, U.S. Patent 9,729,885, issued August 8, 2017.}
\item{H Yu, M Xu, W Wang, F Duanmu, and S Minaee, \emph{``Advanced Coding Techniques For High Efficiency Video Coding (HEVC) Screen Content Coding (SCC) Extensions''}, U.S. Patent Application 15/178,316, filed Dec 22, 2016.}
\end{enumerate}

\end{document}